\documentclass[single-column]{fairmeta}

\usepackage{placeins}
\usepackage{multirow}
\usepackage{pifont}
\usepackage{booktabs}
\usepackage{makecell}
\usepackage[table]{xcolor}
\usepackage{amssymb}

\hypersetup{
  colorlinks=true,
  linkcolor=blue,
  citecolor=blue,
  urlcolor=blue
}

\definecolor{headmethod}{RGB}{226, 235, 247}
\definecolor{headmodal}{RGB}{226, 239, 228}
\definecolor{headfusion}{RGB}{246, 235, 220}
\definecolor{headrep}{RGB}{238, 229, 245}
\definecolor{tablesubhead}{RGB}{218, 218, 218}
\definecolor{tablecolhead}{RGB}{255, 255, 255}
\definecolor{tablerowA}{RGB}{235, 235, 235}
\definecolor{tablerowB}{RGB}{255, 255, 255}
\definecolor{cA}{RGB}{210, 228, 255}
\definecolor{cB}{RGB}{255, 238, 204}
\definecolor{cC}{RGB}{204, 245, 214}
\definecolor{cD}{RGB}{250, 212, 230}
\definecolor{cS}{RGB}{225, 225, 225}
\newcommand{\yes}{\ding{51}}
\newcommand{\y}{\ding{51}}
\newcommand{\n}{}
\newcommand{\rot}[1]{\makebox[\linewidth][c]{\rotatebox[origin=lB]{90}{\tiny\strut #1\hspace{2pt}}}}
\newcolumntype{T}{>{\centering\arraybackslash}p{0.27cm}}
\newcommand{\orcidauthor}[2]{%
  \href{https://orcid.org/#2}{\textcolor{black}{#1}}%
}

\providecommand{\Description}[1]{}
\providecommand{\authorsaddresses}[1]{}
\bibpunct{[}{]}{,}{n}{}{,}
\setcitestyle{numbers,square,comma,sort}
\makeatletter
\def\NAT@sort{1}
\makeatother

\pretocmd{\section}{\FloatBarrier}{}{}
\pretocmd{\subsection}{\FloatBarrier}{}{}

\AtBeginDocument{%
  }

\begin{document}

\nocite{zhi2025learningunifiedpolicyposition, portela2024learning, noseworthy2025forge, zhang2025ta, sferrazza2024power, hou2025adaptive, zhou2025admittance, luo2025precise, buamanee2024bi, chen2025dexforce, he2025foar, liu2025forcemimic, aburub2026learning, kamijo2024learning, lin2025learning, xue2025reactive, kang2025robotic, wu2025tacdiffusion, helmut2025tactile, heng2025vitacformer, huang20243d, xu2025unit, jones2025beyond, yu2025forcevla, huang2025tactile, hao2025tla, ge2025filic, lee2025manipforce, shirai2025sim, zhao2025touch, murooka2025tact, ye2026visual, huang2026tactile, fang2026force, liu2025factr, ablett2024multimodal, collins2024forcesight, zhang2026craft, seo2025equicontact, tang2026towards, li2026forcevla2, ruan2026retac, zhang2026tacvla, zhang2025kinedex, zheng2026omnivta, cheng2025omnivtla, tian2026vitas, bi2025vla, wu2025canonical, sun2025vtao, zhang2025vtla, sun2024arch, zhang2026touchguide}

\title{Learning Physical Interaction: A Survey of \\ 
Tactile- and Force-aware Robot Learning}


\author[1,*,\ddag]{\orcidauthor{Shilin~Shan}{0000-0002-2248-0215}}
\author[1,*]{\orcidauthor{Chuhao~Zhou}{0000-0002-4363-8931}}
\author[1,*]{\orcidauthor{Ruize~Wang}{0000-0001-7252-6237}}
\author[1,*]{\orcidauthor{Xinyan~Chen}{0000-0002-9174-6558}}
\author[1,*]{\orcidauthor{Xiangyu~Chen}{0000-0001-8848-879X}}
\author[1,*]{\orcidauthor{Xinyu~Zhou}{0009-0004-5734-1305}}
\author[1,*]{\orcidauthor{Boyu~Ma}{0000-0002-9628-9886}}
\author[1]{\orcidauthor{Iris~Yuxuan~Hu}{0000-0003-3799-4072}}
\author[1]{\orcidauthor{Jingliang~Li}{0009-0003-1896-991X}}
\author[1]{\orcidauthor{Celeste~Yuxuan~Hu}{0009-0001-0592-6430}}
\author[1]{\orcidauthor{Geng~Li}{0009-0002-1989-7922}}
\author[1]{\orcidauthor{Guohao~Chen}{0009-0007-9736-4642}}
\author[1]{\orcidauthor{Tianrui~Zhu}{0009-0003-7502-4875}}
\author[1]{\orcidauthor{Zhe~Li}{0009-0008-3896-7694}}
\author[2]{Yanjie~Ze}
\author[3]{Haoran~Geng}
\author[4]{Zhiyang~Dou}
\author[5]{Jianxin~Bi}
\author[2]{Yuejiang~Liu}
\author[5]{Jianshu~Zhou}
\author[6]{Jiachen~Li}
\author[4]{Paul~Liang}
\author[7]{Tatsuya~Harada}
\author[8]{Robert~Katzschmann}
\author[5]{Harold~Soh}
\author[9]{Na~Li}
\author[10]{Edward~Johns}
\author[11]{Danica~Kragic}
\author[12]{Jan~Peters}
\author[4]{Wojciech~Matusik}
\author[3]{Masayoshi~Tomizuka}
\author[3]{Jitendra~Malik}
\author[1,\dag]{Jianfei~Yang}

\affiliation[1]{{\footnotesize Nanyang~Technological~University}}
\affiliation[2]{{\footnotesize Stanford~University}}
\affiliation[3]{{\footnotesize University~of~California,~Berkeley}}
\affiliation[4]{{\footnotesize Massachusetts~Institute~of~Technology}}
\affiliation[5]{{\footnotesize National~University~of~Singapore}}
\affiliation[6]{{\footnotesize Georgia~Institute~of~Technology}}
\affiliation[7]{{\footnotesize The~University~of~Tokyo}}
\affiliation[8]{{\footnotesize ETH~Zurich}}
\affiliation[9]{{\footnotesize Harvard~University}}
\affiliation[10]{{\footnotesize Imperial~College~London}}
\affiliation[11]{{\footnotesize KTH~Royal~Institute~of~Technology}}
\affiliation[12]{{\footnotesize Technical~University~of~Darmstadt}}

\contribution[*]{{\footnotesize Equal Contribution}}
\contribution[\ddag]{{\footnotesize Project Lead}}
\contribution[\dag]{{\footnotesize Corresponding Author}}

\abstract{
  Physically grounded robot intelligence requires robots to perceive, reason about, and regulate their interactions with the physical world. This capability is particularly critical in contact-sensitive manipulation, where successful task execution depends not only on visual perception and motion generation, but also on force regulation and adaptive control. In this context, recent robot learning methods have made substantial progress by integrating force, tactile, vision, language, and proprioceptive sensing into learned manipulation policies. In parallel, many systems adopt multi-phase architectures that combine high-level policies, action-refinement modules, and low-level controllers to bridge semantic task understanding with reactive physical execution. Despite these advances, existing surveys have not explicitly reviewed force- and tactile-aware robot learning from a unified perspective that jointly captures multimodal sensing and multi-phase system design. This survey addresses this gap by proposing \textbf{TF-ART}, a \textbf{T}actile/\textbf{F}orce-\textbf{A}ware \textbf{R}obot learning \textbf{T}axonomy for multimodal and multi-phase frameworks, which maps individual methods into a unified hierarchical structure. The framework characterizes how recent works organize observation modalities, encode and fuse heterogeneous sensory inputs, generate and refine actions across multiple phases, and connect learned policies to reactive robot-end control. Building on this methodological view, we further examine the task settings and infrastructure requirements of physical interaction, thereby integrating both algorithmic and practical perspectives on force- and tactile-aware robot learning.
}

\metadata[Keywords]{Force-aware Manipulation, Multimodal Learning, Compliant Control, Embodied AI}

\metadata[GitHub]{\url{https://github.com/NTUMARS/Awesome-Tactile-Force-aware-Robot-Learning}}

\metadata[Webpage]{\url{https://lorenzo-0-0.github.io/tactile-force-survey/}}

\correspondence{Jianfei Yang at \email{jianfei.yang@ntu.edu.sg};}

\maketitle

\section{Introduction}

\subsection{Background}

\begin{figure}[!t]
  \centering
  \includegraphics[width=\textwidth,height=0.98\textheight,keepaspectratio,trim=0.5cm 5.5cm 0.5cm 5.0cm,clip]{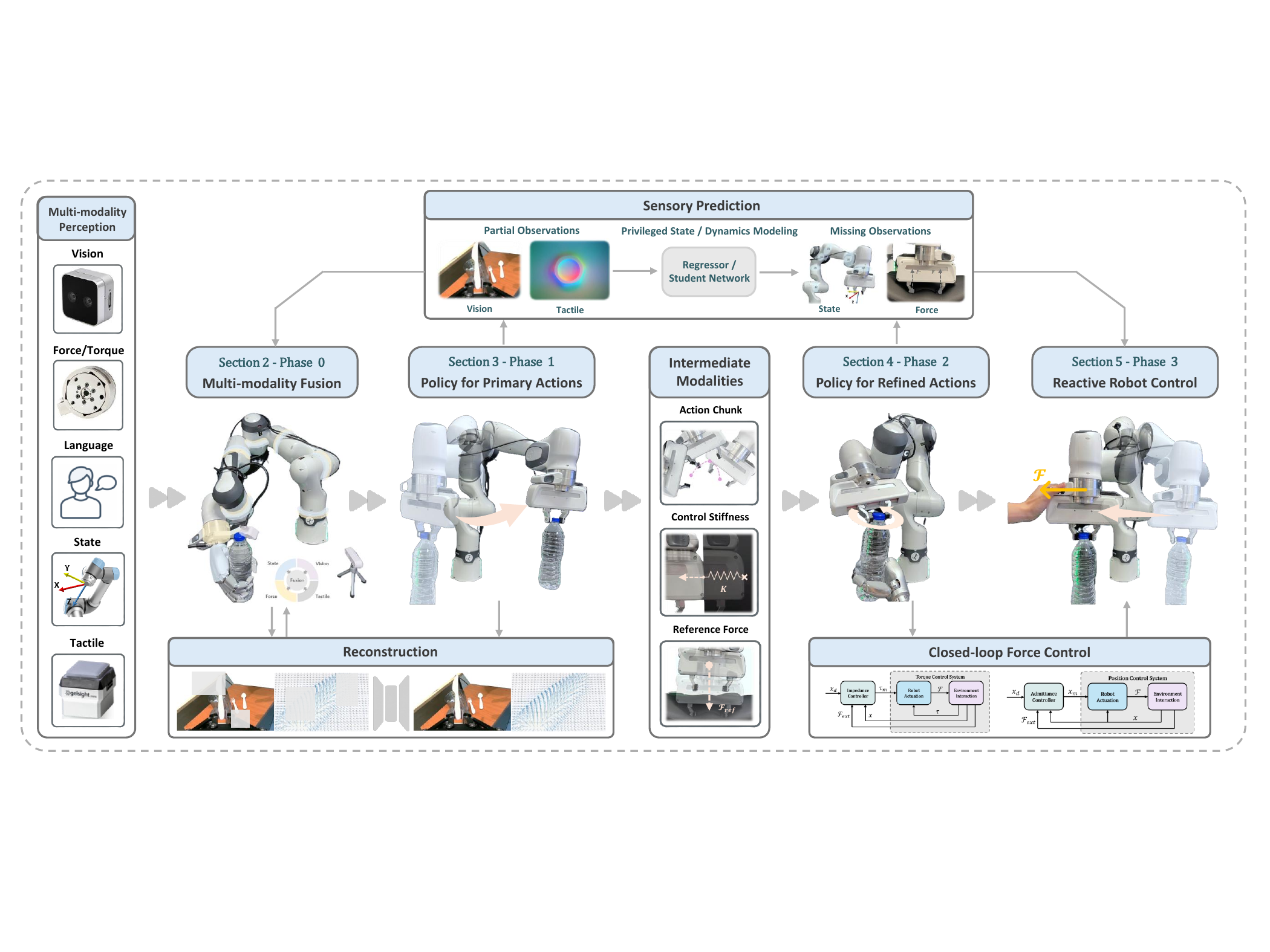}
  \caption{Overview of TF-ART: a unified force/tactile-aware multi-modal multi-phase framework. The main hierarchical architecture includes multimodal perception, encoding and fusion, primary action generation, action refinement, and low-level robot-end control. The branching modules include input data reconstruction, missing-modality prediction, and explicit force-control algorithms.}
  \Description{}
  \label{Fig:teaser}
\end{figure}

Embodied intelligence has experienced rapid development in recent years, driven by advances in multimodal sensing, large-scale reasoning models, and generative motion planning~\cite{brohan2022rt, driess2023palm, chi2025diffusion, black2024pi, intelligence2025pi}. Existing approaches often combine visual perception with language instruction for object recognition and scene understanding. These observations provide rich semantic and geometric cues and are effective for general manipulation tasks. However, when robots move from free-space motion to physical interaction, the most task-critical information often shifts from visual observations to interaction feedback~\cite{zhao2025touch, he2025foar, wu2025tacdiffusion, ablett2024multimodal}. Under contact, the target object may be occluded, its state may change due to deformation, and successful execution may depend on where contact occurs, whether it is stable, and how large the interaction force becomes. Force and tactile sensing therefore provide essential feedback that complements vision and proprioception~\cite{sferrazza2024power, huang20243d, chen2025dexforce, song2025opentouch}. Specifically, force sensing captures global interaction information, such as end-effector wrench and joint torque, supporting force regulation and compliant control. Tactile sensing captures local contact information, such as pressure distribution, deformation, and surface texture. Together, force and tactile feedback allow robots to sense, interpret, and regulate physical interaction beyond what can be inferred from vision-centered observations.

The use of force and tactile information has been studied from both learning-based and control-oriented perspectives. Learning-based methods enable task-relevant representations and action-generation policies to be developed from collected data, including expert demonstrations and autonomous interaction data gathered through exploration, rather than requiring their behavior to be fully specified analytically~\cite{sferrazza2024power, murooka2025tact, ye2026visual, ablett2024multimodal}. Classical control-oriented methods use explicitly designed feedback mechanisms to provide stable physical interaction and force regulation under contact~\cite{raibert1981hybrid, Hogan1985, Khatib1987, keemink2018admittance}. These approaches offer different but compatible ways of developing contact-aware robot behavior, and recent literature increasingly combines learning-based and control-oriented components across different parts of the manipulation pipeline. This development is reflected in two closely related directions: multimodal integration for contact-aware perception and action generation, and multi-phase policy-to-control architectures for contact-sensitive execution.

From the perspective of multimodal integration, recent works have explored the complementarity between contact-free and contact-rich perception by jointly integrating force, tactile, vision, language, and proprioceptive state information for action generation~\cite{jones2025beyond, yu2025forcevla, huang2026tactile, cheng2025omnivtla, song2025opentouch}. Since the same object or interaction event can be observed through multiple sensory modalities, multimodal representation learning provides a natural way for different modalities to exchange information. These works therefore raise key questions about how force and tactile signals complement other modalities, how their representations can leverage existing pre-trained models, and how such integration guides contact-aware action generation.

From the perspective of multi-phase design, force and tactile feedback are increasingly incorporated across different stages of the action-generation pipeline rather than treated as passive inputs to a single end-to-end policy. In recent studies, a high-level policy may first generate a coarse trajectory from vision, language, and state observations, while subsequent phases use force or tactile feedback for motion refinement or direct force control~\cite{he2025foar, xue2025reactive, zheng2026omnivta}. This pattern highlights a central question in contact-rich manipulation: semantic reasoning and global trajectory generation can benefit from large vision-language-based policies, yet such policies may not satisfy the low-latency and high-precision requirements of local, force-sensitive interaction.

\begin{table*}[t]
\centering
\caption{Comparison of prior surveys by their primary organizing dimensions.}
\label{tab:survey_comparison}
\setlength{\tabcolsep}{3.5pt}
\renewcommand{\arraystretch}{1.12}
\resizebox{\textwidth}{!}{%
\begin{tabular}{@{}llcccccc@{}}
\toprule
\textbf{Survey family} & \textbf{Representative surveys} & \textbf{Obs.} & \textbf{Rep./Fusion} & \textbf{Policy} & \textbf{Refine.} & \textbf{Control} & \textbf{Pipeline} \\
\midrule
Tactile sensing, skins, and hardware & \cite{nicholls1989survey,lee1999tactile,tegin2005tactile,dahiya2010tactile,yousef2011tactile,girao2013tactile,dahiya2013directions,kappassov2015tactile,roberts2021soft,cheng2019comprehensive} & $\bullet$ & $\circ$ & -- & -- & $\circ$ & -- \\
Tactile and visuo-haptic perception & \cite{luo2017robotic,li2020review,navarro2023visuohaptic,li2024vision,albini2025representing}  & $\bullet$ & $\bullet$ & $\circ$ & -- & -- & -- \\
Active-haptic and sensorimotor interaction & \cite{seminara2019active,donato2025sensorimotor}  & $\bullet$ & $\circ$ & $\circ$ & $\circ$ & $\bullet$ & $\circ$ \\
General robot learning and learning from demonstration & \cite{kroemer2021review,ravichandar2020recent,osa2018algorithmic,zare2024survey,urain2025survey} & $\circ$ & $\circ$ & $\bullet$ & -- & $\circ$ & -- \\
Contact manipulation and contact-rich imitation learning & \cite{suomalainen2022survey,tsuji2025survey} & $\circ$ & $\circ$ & $\bullet$ & $\circ$ & $\bullet$ & $\circ$ \\
Dexterous and interactive robot learning & \cite{11317793,an2025dexterous,welte2025interactive} & $\circ$ & $\circ$ & $\bullet$ & $\circ$ & $\circ$ & -- \\
Robot foundation-model perspectives & \cite{firoozi2025foundation,hu2023toward,xie2025towards} & $\circ$ & $\bullet$ & $\bullet$ & $\circ$ & $\circ$ & $\circ$ \\
Broad tactile-robotics outlook & \cite{luo2025tactile} & $\bullet$ & $\circ$ & $\circ$ & $\circ$ & $\circ$ & -- \\
\textbf{TF-ART (ours)} & \textbf{This survey} & $\bullet$ & $\bullet$ & $\bullet$ & $\bullet$ & $\bullet$ & $\bullet$ \\
\bottomrule
\end{tabular}%
}
\vspace{2pt}

\begin{minipage}{\textwidth}
\footnotesize
\textit{Obs.}: contact-aware multimodal observations; \textit{Rep./Fusion}: multimodal representation and fusion; \textit{Policy}: learned primary action generation; \textit{Refine.}: contact-aware action refinement; \textit{Control}: robot-end force/compliance control; \textit{Pipeline}: unified policy--control pipeline mapping. $\bullet$ denotes a primary organizing dimension, $\circ$ denotes partial coverage, and -- denotes no systematic coverage. 
\end{minipage}
\end{table*}

Taken together, the above developments suggest that force- and tactile-aware robot learning is no longer defined merely by the inclusion of additional sensory inputs, but by how these inputs are integrated with other modalities and organized across multiple stages of the policy pipeline. Despite this progress, existing research tends to approach multimodal learning and multi-phase design from only partial perspectives, usually covering only a subset of components such as representation learning, multimodal fusion, policy architecture, action refinement, and robot-end control. Although prior surveys have covered related topics~\cite{kroemer2021review, ravichandar2020recent, suomalainen2022survey, tsuji2025survey}, as summarized in the next subsection, these discussions remain scattered across different areas and do not yet provide a unified view of contact-rich manipulation. This gap motivates a dedicated survey on force- and tactile-aware robot learning.

\subsection{Related Surveys}

Earlier tactile surveys primarily organized the field around sensing technologies and embodiment. Foundational reviews examined tactile transduction, sensor structures, signal processing, and robotic applications~\cite{nicholls1989survey, lee1999tactile, tegin2005tactile, dahiya2010tactile}. Subsequent surveys extended this perspective to dexterous hands, tactile skins, application-specific sensing requirements, and soft robotic skins~\cite{yousef2011tactile, girao2013tactile, dahiya2013directions, kappassov2015tactile, roberts2021soft}, while Cheng et al.~\cite{cheng2019comprehensive} considered the integration of robot skin with sensing, control, and applications. Other reviews shifted the focus toward tactile perception and representation~\cite{luo2017robotic, li2020review, albini2025representing}, visuo-haptic fusion~\cite{navarro2023visuohaptic, li2024vision}, and active sensorimotor interaction~\cite{seminara2019active, donato2025sensorimotor}. Together, these surveys establish the sensing and perception foundations of tactile robotics, but are generally organized around sensors, representations, or specific interaction capabilities rather than learned policy pipelines.

A complementary body of surveys focuses on robot learning and contact-rich manipulation. General reviews cover manipulation learning and learning from demonstration~\cite{kroemer2021review, ravichandar2020recent, osa2018algorithmic, zare2024survey}, generative robot learning~\cite{urain2025survey}, and foundation models~\cite{firoozi2025foundation, hu2023toward}. Surveys of contact manipulation and contact-rich imitation learning discuss task formulations, demonstrations, learning methods, and force/compliance control~\cite{suomalainen2022survey, tsuji2025survey}, while dexterous and interactive-learning surveys emphasize hand hardware, teleoperation, and human involvement~\cite{11317793, an2025dexterous, welte2025interactive}. Particularly close to our scope, the tactile-robotics outlook~\cite{luo2025tactile}, the sensorimotor reviews~\cite{seminara2019active, donato2025sensorimotor}, and the force-aware foundation-model perspective~\cite{xie2025towards} cover several components of force- and tactile-aware learning. However, none uses the complete multimodal policy-control pipeline as its primary organizing axis.

Table~\ref{tab:survey_comparison} compares the evaluation dimensions of related surveys. TF-ART differs from prior surveys mainly in its unit of analysis. Rather than organizing methods solely by sensor type, task, or learning paradigm, TF-ART maps each method onto a common information and control flow: multimodal observation, representation and fusion, primary action generation, contact-aware refinement, and robot-end control. This organization identifies not only whether force or tactile feedback is used, but also where and for what purpose it enters the system. TF-ART thereby connects multimodal integration with multi-phase policy organization and provides a unified view of how learned policies interface with high-frequency force and compliance control.

\subsection{Contributions}

Motivated by the need for a unified organization of force- and tactile-aware robot learning, this survey proposes \textbf{TF-ART}, a \textbf{T}actile/\textbf{F}orce-\textbf{A}ware \textbf{R}obot learning \textbf{T}axonomy for multimodal and multi-phase frameworks, as illustrated in Figure~\ref{Fig:big_framework_all}. The main contributions are summarized as follows:

\begin{itemize}
  \item TF-ART summarizes recent methodologies through a unified hierarchical architecture. By mapping diverse methodological pipelines to corresponding framework modules, TF-ART provides a comprehensive view of recent progress in force- and tactile-aware robot learning.

  \item We examine how force and tactile signals are incorporated into multimodal robot policies. Beyond listing input modalities, we discuss how contact-related observations are represented, aligned, and fused with vision, language, and state information, as well as how different primary policy architectures use these fused representations for action generation. The discussion emphasizes the basic mechanisms, benefits, limitations, and task-dependent suitability of these multimodal designs.

  \item We analyze how force and tactile feedback reshape robot learning pipelines beyond single-stage action prediction. In particular, we discuss phase-specific sensing, discriminative modality injection, action refinement, and low-level control, showing how recent systems use contact feedback to correct high-level policy outputs, improve real-time responsiveness, and support stable physical interaction.

  \item We discuss practical and implementation-level factors in force- and tactile-aware robot learning. These include sensor types and sensor models, robot hardware capabilities, and representative contact-rich manipulation tasks considered in recent studies.
\end{itemize}

\subsection{Framework Overview and Survey Organization}

\begin{figure}[!h]
  \centering
  \includegraphics[width=\textwidth,height=0.88\textheight,keepaspectratio,trim=1.2cm 0.0cm 0.3cm 0.0cm, clip]{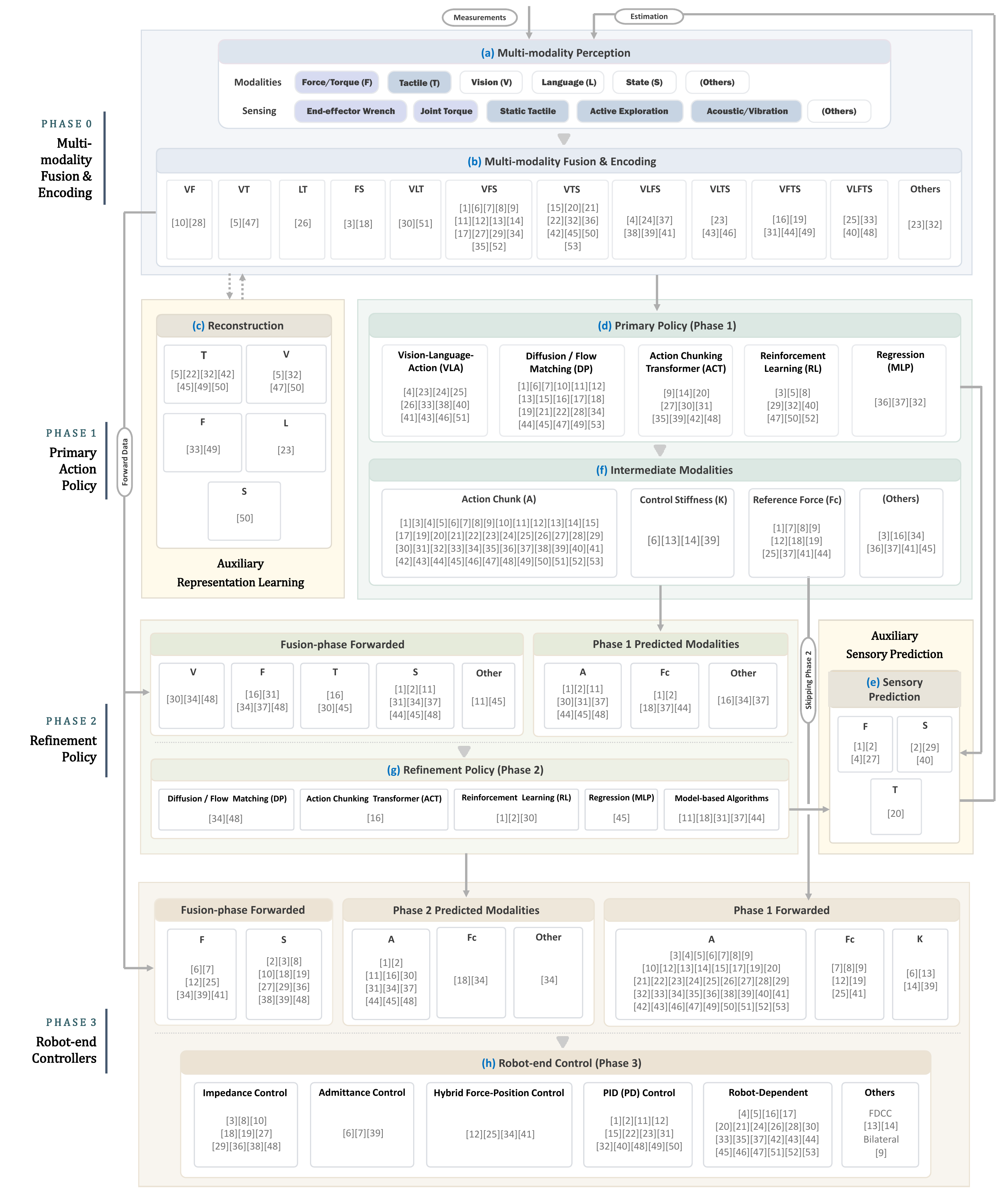}
  \vspace{-5pt}
  \caption{Detailed TF-ART framework covering all investigated research papers.\textsuperscript{*} The framework components span the following: (a) multimodal observation; (b) multimodal fusion; (c) input reconstruction; (d) Phase 1: primary action-generation policy; (e) sensory prediction; (f) intermediate modalities; (g) Phase 2: action-refinement policy; and (h) Phase 3: robot-end control.}
  \vspace{10pt}
  {\raggedright\footnotesize\textsuperscript{*}Readers may use the PDF viewer's search function to locate a paper's reference number, as listed in the reference section, within the framework and inspect the corresponding components and methodological architecture categorized in this survey.\par}
  \Description{}
  \label{Fig:big_framework_all}
\end{figure}

Figure~\ref{Fig:big_framework_all} presents the complete TF-ART framework and illustrates how the reviewed force- and tactile-aware studies are mapped onto it. Each paper is analyzed and categorized according to its methodological components, allowing its pipeline to be aligned with the corresponding modules of the proposed framework. Specifically, in a top-down sequence, TF-ART organizes its components as follows: (a) the observation space, that is, the sensory measurements used as model inputs; (b) input encoding and modality fusion methods, grouped according to modality-usage criteria; (c) input reconstruction and representation methods for encoder training; (d) primary action-generation architectures, which constitute the phase-one policy and are shared by most reviewed works; (e) sensory prediction methods for missing-modality replacement; (f) intermediate modalities that cannot be directly measured and are better suited for control or action refinement; (g) the action-refinement layer, or phase-two policy, which leverages forwarded encoded inputs and phase-one predictions to generate refined actions for fine-grained objectives; and (h) the control layer, or phase three, which connects learned models to robot hardware and supports safe execution through explicit force-control mechanisms.

For literature search and selection, we consider a method tactile-aware when it uses contact measurements from camera-based tactile sensors, tactile pads or arrays, robotic skins, or related contact-sensing devices. A method is considered force-aware when it uses or explicitly estimates force-related information obtained from wrist or end-effector force/torque sensors, joint-torque sensors, motor currents, or other proprioceptive signals. To qualify, tactile or force information must be used as a policy input or training signal, or explicitly predicted as an intermediate representation or output within the learned action-generation process. Robot learning is defined here as a data-driven method that generates robot actions or action-related control targets. Accordingly, we exclude studies limited to sensor design, calibration, contact or force estimation, and representation learning without evaluation in a downstream action-generation policy, as well as classical controllers without a learned action-generation component and generic manipulation policies that neither use nor predict tactile or force information. Multi-phase design is an analysis category rather than an inclusion requirement; it refers to an explicit separation between primary action generation and one or more subsequent contact-aware refinement or force/compliance-control stages. Candidate papers were identified through searches of Google Scholar, arXiv, and IEEE Xplore, supplemented by backward and forward citation tracing. The search window for the core corpus mapped to TF-ART covered work published from 2024 through April 2026, while the survey’s broader reference list was last updated on July 21, 2026. This process yielded a final TF-ART corpus of 53 papers.

\clearpage

This survey is organized around the proposed unified framework, with each section corresponding to a major component of the overall architecture. This organization is also reflected in Figure~\ref{Fig:big_framework_all}. Section~\ref{S2} introduces the modalities commonly used in recent works, how they are encoded and fused during training, and how encoders can be trained through input data reconstruction. Section~\ref{S3} discusses policy architectures for primary action generation, following mainstream research directions, together with intermediate modality prediction for the next phase and sensory prediction for missing measurements. Section~\ref{S4} discusses techniques for refining actions generated by phase-one policies, as well as modality-selection considerations for fast inference. Section~\ref{S5} introduces the fundamentals of model-based low-level robot control, with a particular focus on force-adaptive and compliance-control techniques used for both data collection and real-world interaction. Section~\ref{S6} discusses common tasks and real-world settings demonstrated in existing studies, including their characteristics and practical implications.

\section{Phase 0: Multimodal Fusion and Representation Learning} \label{S2}

\begin{figure}[t]
  \centering
  \includegraphics[width=\textwidth,height=0.88\textheight,keepaspectratio,trim=1cm 4.8cm 1cm 3.5cm,clip]{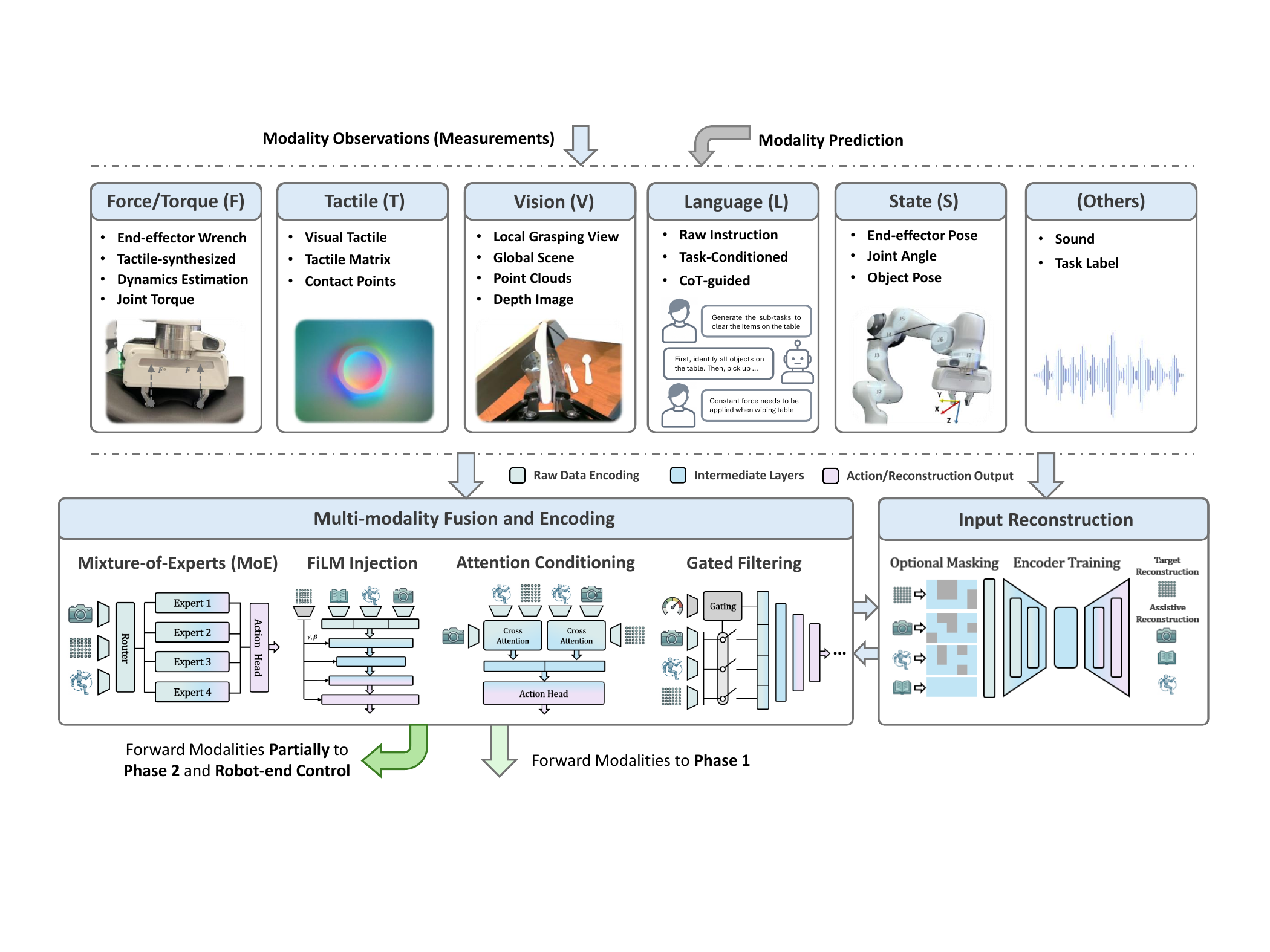}
  \caption{Overview of the multimodal fusion and encoding framework. The upper part visualizes the physical interpretation of different modalities and their sub-classes, while the lower part illustrates the model structures of major representation-learning and modality-fusion methods.}
  \Description{}
  \label{Fig:framework_fusion}
\end{figure}

A multimodal robotic system must address three key questions: which sensing modalities are needed for a given task, how raw multimodal observations can be represented and grounded in robotic control, and how complementary information across modalities can be effectively fused and exploited. Accordingly, this section reviews prior works from five perspectives: tactile (Section~\ref{subsection: Tactile}), force (Section~\ref{subsection: Force}), non-contact modalities (Section~\ref{subsection: diverse modalities}), multimodal representation learning (Section~\ref{subsection: Multimodal Representation Learning}), and multimodal fusion (Section~\ref{subsection: multimodal fusion}). The surveyed papers are summarized in Table~\ref{tab:modality_fusion_summary} according to the metrics introduced in this section. A visual illustration of the modalities, representation learning methods, and input-output data flow is shown in Figure~\ref{Fig:framework_fusion}. We refer to this part as Phase 0 as it defines the sensory and representational basis before action generation, whereas Phases 1-3 correspond to explicit stages of action generation.

\subsection{Tactile (T)}
\label{subsection: Tactile}

\subsubsection{Tactile Sensing Methods}
\label{sec:tactile_sensing}

Tactile perception during manipulation is closely coupled with action. The same object can produce different sensor readings when the contact location, normal force, motion direction, speed, or exploratory action changes. By controlling these variables, the robot can make different geometric and material properties easier to detect~\cite{gibson1962observations,prescott2011active}. Existing tactile sensing methods can be grouped into four categories according to how tactile signals are collected.

\textbf{(a) Static contact and pressure sensing.}
When the robot maintains stable contact with little relative motion, the interaction produces a sustained deformation, pressure, or force distribution. The resulting observation may be represented as a tactile image, a pressure or force map, or a set of distributed contact measurements. Its spatial pattern reflects the contact location, area, and local surface geometry and can support the estimation of contact states and object attributes such as shape and pose. A single contact primarily captures a local surface patch, whereas multiple contacts distributed across the hand additionally capture the spatial relationships among different surface regions, facilitating object recognition and localization. These methods therefore rely primarily on static or quasi-static spatial contact patterns rather than exploratory motion or rapid temporal variations~\cite{gaston1984tactile,luo2017robotic}.

\textbf{(b) Active exploration of shape and material properties.}
Geometric and material properties often become clear only when the sensor moves in a suitable way. Rolling or sliding a fingertip over a surface converts cracks, bumps, and ridges into changes in the contact trajectory. Varying the normal force and speed produces different vibration and reaction-force patterns that help distinguish textures. When object identity or contact location is uncertain, the current estimate can also guide the next contact action~\cite{okamura2001feature, fishel2012bayesian, lepora2013active}. Exploratory motion is therefore part of the sensing process rather than a fixed data-collection procedure.

\textbf{(c) Dynamic contact events.}
Static pressure maps mainly describe sustained contact and may miss short-lived changes. Dynamic tactile sensing instead measures acceleration, vibration, or stress-rate signals to detect slip onset and fine surface features~\cite{howe1989sensing,howe1993dynamic}. Event-based tactile sensors encode rapid signal changes as asynchronous events and, when fused with event-based vision, support low-latency object classification and rotational-slip detection~\cite{taunyazov2020event}. In these methods, the temporal pattern of the signal is more important than the static pressure distribution.

\textbf{(d) Tactile signal transmission through the hand, object, and tool.}
Contact signals can travel through the skin, hand, grasped object, or tool before reaching the sensor. The measured deformation and vibration therefore depend on the mechanical connection between the contact point and the sensor. Whole-hand vibration patterns and vibrations transmitted through tools can be used to infer contacts outside the immediate sensing area~\cite{hayward2011plenhaptic, shao2016spatial, miller2018sensing, taunyazov2021extended}. This extends tactile perception beyond local sensing at the fingertip or skin.

Overall, tactile sensors can produce deformation images, taxel arrays, contact points, pressure or force sequences, vibration signals, and event streams. These measurements are used to estimate surface shape, texture, contact state, and interaction dynamics. Because the measurements change with the robot's actions, tactile perception naturally links sensing, exploration, and control.

\subsubsection{Early Tactile Learning}
Early tactile-learning methods used data-driven models to map sequences of tactile measurements and actions to object identity, material properties, contact states, or control commands. Representative methods can be grouped into three categories.

\textbf{(a) Temporal recognition and continual learning.}
A single contact usually provides only local and ambiguous information. Temporal models reduce this ambiguity by combining measurements across a sequence of contacts. Online models update recognition when new measurements arrive, while incremental discriminative and generative models add new object classes without losing previously learned classes. Robust tactile descriptors further improve recognition across sensor types, exploratory motions, and interaction durations~\cite{soh2012online,soh2014incrementally,kaboli2018robust}. Temporal accumulation and online updating therefore make recognition less dependent on a single contact or a fixed training set.

\textbf{(b) Multimodal learning from vision and touch.}
Vision and tactile sensing provide complementary information at different stages of manipulation. Vision captures global object geometry and scene context before contact, whereas tactile sensing reveals local contact conditions, deformation, and slip after contact. One use of this complementarity is to predict the outcomes of candidate grasp adjustments and select an appropriate regrasp action~\cite{calandra2018more}. A broader formulation learns shared representations from paired visual and tactile observations through self-supervised learning before policy training, allowing the learned features to transfer across contact-rich tasks~\cite{lee2020making}. For interactions requiring faster responses, event-based methods fuse asynchronous visual and tactile streams for low-latency object classification and rotational-slip detection~\cite{taunyazov2020event}. Together, these formulations use global visual context and local contact evidence for action-conditioned outcome prediction, transferable representation learning, or rapid contact-event detection.

\textbf{(c) Tactile feedback for manipulation.}
Tactile feedback can be used at several levels of the control hierarchy. At the lowest level, contact and slip detection adjust grasp force. At the motion level, interaction data are used to learn compliant behaviors or maintain contact while following an object contour. At the skill level, manipulation is divided into primitives, and tactile states are used to monitor execution and trigger corrections or transitions~\cite{romano2011human, kronander2013learning, lepora2017exploratory, hogan2020tactile, welle2023enabling}. This progression extends tactile learning from local grasp regulation to the execution of complete manipulation skills.

These studies established temporal modeling, uncertainty-based exploration, joint vision–touch learning, and closed-loop control as core parts of tactile learning. However, most systems were developed for a specific task, sensor, object set, robot platform, or exploration procedure. The methods reviewed in this survey build on these ideas through shared multimodal representations, broader task conditioning, and generative action models.

\subsubsection{Tactile Measurement Format} \label{sec:tactile_format}
To use different tactile sensors in a common robot-learning pipeline, raw sensor outputs must be converted into a representation that can be processed by the model. This survey considers three main forms that are commonly adopted in the surveyed 53 papers: visual-tactile images, tactile matrices, and contact points. They differ mainly in how contact is organized spatially: as a dense image, a regular grid of sensing elements, or a sparse set of locations.

\textbf{Visual-Tactile Images.}
Camera-based tactile sensors record changes in an elastomeric surface as images. Depending on the sensor design, an image may reflect contact shape, surface texture, marker displacement, and local deformation. After calibration, the image can also be converted into depth, displacement, force, shear, or deformation maps. Because these data follow a regular image grid, they can be processed directly using CNNs or Vision Transformers. This representation preserves fine spatial detail, but it is relatively high-dimensional and often depends on the sensor's optical design and calibration.

\textbf{Tactile Matrices.}
A tactile matrix represents a regular array of tactile sensing elements, or taxels. Each grid cell corresponds to one taxel and may contain pressure, normal force, shear force, displacement, or vibration. CNNs or Transformers can process the grid while preserving the spatial relationships between neighboring taxels. Compared with camera-based tactile images, this representation can be lower-dimensional and often records physical quantities more directly. However, its shape and resolution are tied to the layout of the sensor.

\textbf{Contact Points.}
Contact points are useful when active contacts are sparse or sensing elements are placed on an irregular surface. Each point stores a spatial location and may include attributes such as force, displacement, surface normal, or contact state. Point-based encoders or Transformers can process a variable number of points without requiring a fixed two-dimensional grid. This representation adapts well to curved surfaces and different sensor layouts, although sparse points may not retain fine texture or dense deformation patterns.

All three forms can also be arranged as temporal sequences to capture changes during contact. They are not mutually exclusive: tactile images can be converted into depth or force maps, taxel grids can be reduced to active contact points, and contact points can be combined with visual geometry. Together, they define the tactile input branch of Phase~0 and allow tactile measurements to be encoded and fused with vision, language, force/torque, and robot state.

\subsection{Force/Torque (F)}
\label{subsection: Force}

\subsubsection{Force/Torque Sensing Methods}
\label{sec:force_torque_sensing}

Force/torque sensing provides information about external contact loads exerted on the robot by the environment or a human. These loads are typically represented as a six-dimensional wrench at the end effector or as torques transmitted through individual joints. They can be measured directly using dedicated sensors or estimated indirectly from robot states and actuator signals. This subsection reviews six-axis force/torque sensing, joint torque sensing, and sensorless external-load estimation.

\textbf{(a) Six-axis force/torque sensing.}
A conventional six-axis force/torque sensor contains a slightly deformable elastic structure. When an external force or torque is applied, the structure bends or twists slightly. Strain gauges attached to the structure convert these deformations into changes in electrical quantities, such as resistance or voltage. Forces and torques applied along different directions produce distinct combinations of signals across the strain gauges. Using a mapping obtained through calibration, the sensor converts these signals into numerical measurements of wrench components~\cite{bicchi1992criterion, liu2002novel}.

\textbf{(b) Joint torque sensing.}
A joint torque sensor is installed inside a robot joint, typically between the gearbox and the connected link. It contains an elastic element that twists slightly as torque is transmitted through the joint. Strain gauges attached to this element convert the deformation into changes in electrical quantities, such as resistance or voltage. After calibration, these signals are mapped to a numerical torque measurement for the joint. The sensor is designed to respond primarily to torque about the joint axis while minimizing sensitivity to off-axis forces and moments~\cite{aghili2001design,albu2007dlr}.

\textbf{(c) Sensorless external-load estimation.} 
When direct force/torque sensing is unavailable, contact information can instead be inferred from proprioceptive signals or visual observations. To fully replace dedicated force/torque sensors, proprioception-based approaches commonly train a separate estimator or explicitly model robot dynamics to estimate external contact loads. These approaches typically use indirect force-related signals, such as motor currents or joint-position tracking errors, model robot dynamics using either model-based or learning-based methods, and estimate joint-level external torques or the contact wrench at the end effector~\citep{khalil2004modeling, hu2017contact, 10669209, 10609990, yilmaz2020neural, dou2026neuralactuator}.

\subsubsection{Force/Torque Measurement Format}

This survey distinguishes three input forms according to where the load information is obtained: End-Effector Wrench, Joint Torque, and Tactile-Based Force Estimates.

\textbf{End-Effector Wrench.}
An end-effector wrench combines the net force and moment acting at a specified reference frame. It is commonly represented as a 6-dimensional vector \vspace{-5pt}
\[
\mathbf{w}_t =
\begin{bmatrix}
\mathbf{f}_t^\top &
\mathbf{m}_t^\top
\end{bmatrix}^{\top}
\in \mathbb{R}^{6},
\]
where $\mathbf{f}_t\in\mathbb{R}^{3}$ is the Cartesian force and $\mathbf{m}_t\in\mathbb{R}^{3}$ is the Cartesian moment. The wrench may be measured directly using a wrist-mounted force/torque sensor or estimated from robot dynamics. Before being passed to a learning model, the signal is commonly bias-corrected, transformed into a consistent coordinate frame, and compensated for payload and gravity. A temporal encoder can then process either individual measurements or a short history of measurements to extract force/torque features.

\textbf{Joint Torque.}
Joint torque is represented as a $J$-dimensional vector, where $J$ is the number of actuated joints. Depending on the robot platform, these values may be measured by joint-torque sensors or estimated from motor currents and actuator models. Raw joint torque includes loads caused by both robot motion and external contact. Learning systems may therefore use the torque sequence together with joint position and velocity, or use an external-torque residual obtained after subtracting the robot's predicted internal loads. Joint-torque inputs are useful when interaction along the full robot arm is important or when a wrist-mounted force/torque sensor is unavailable.

\textbf{Tactile-Based Force Estimates.}
Force can also be estimated from visual-tactile observations rather than measured using a dedicated force/torque sensor. A calibrated or learned estimator maps visual-tactile images to quantities such as local contact-force components or a 6D wrench~\cite{yuan2017gelsight,lin20239dtact}. The resulting estimates can be represented as temporal sequences and processed in the same way as directly measured force signals. Unlike direct force/torque measurements, however, their accuracy depends on the tactile sensor design, calibration procedure, deformation model, and data used to train the estimator.

These three forms capture different aspects of physical interaction. An end-effector wrench summarizes the net load at a reference frame, joint torque describes loads at the robot's actuators, and tactile-based force estimation recovers force from local contact deformation. Together, they define the force/torque input branch of Phase~0 and allow load-related information to be encoded and fused with vision, language, tactile observations, and robot state for downstream action generation and control.

\subsection{Non-contact Modalities}
\label{subsection: diverse modalities}
Beyond tactile and force/torque sensing, we further consider three complementary modality groups: language (L), vision (V), and proprioceptive/environmental state (S). We also briefly discuss other less commonly adopted modalities that may offer additional sensing and reasoning capabilities for future robotic systems.

\vspace{-10pt}
\subsubsection{Language (L)}

\textit{Human Instructions} are natural-language phrases or sentences that specify the task to be executed, e.g., ``pick up an apple and place it on the table.'' These instructions are first tokenized into discrete textual units and converted into continuous text embeddings. A text encoder, such as BERT~\cite{devlin2019bert}, then transforms these embeddings into language features that provide task-conditioning information for policy learning~\cite{vaswani2017attention, brown2020language, touvron2023llama}.

\vspace{-6pt}
\subsubsection{Visual Perception (V)}

\textit{RGB Images} are among the most important perception modalities in robot learning, as they provide rich appearance cues, including color, texture, object shape, and scene context. These images are typically represented by their height, width, and three color channels. A visual encoder, such as a Convolutional Neural Network (CNN)~\cite{he2016deep} or Vision Transformer (ViT)~\cite{dosovitskiy2020image}, then processes the raw image observations and converts them into visual features for policy-learning.

\textit{Depth Images} provide explicit spatial information, with each pixel indicating the distance from the camera to the corresponding visible surface. Compared with RGB images, depth observations are less sensitive to appearance variations, making them useful for geometry-aware perception, grasping, collision avoidance, and spatial reasoning. Similar to RGB images, depth observations are typically processed by visual encoders such as CNNs or ViTs and converted into geometric visual features.

\textit{Point Clouds} directly represent the 3D structure of objects and environments as a collection of spatial points, where each point records its position in 3D space. They are useful for spatial reasoning, object-pose estimation, affordance prediction, and manipulation in cluttered scenes. Point clouds are commonly processed by point-cloud encoders, such as PointNet~\cite{qi2017pointnet} or Point Transformer~\cite{zhao2021point}, which convert the raw geometric observations into 3D structural features for policy-learning.

\vspace{-6pt}

\subsubsection{Proprioception and Environmental State (S)}

\textit{EEF State} measurement contains the 3D position of the end-effector, its orientation, and the gripper state in a given coordinate frame. The orientation is commonly described using Euler angles or quaternions, while the gripper state can be represented either as a binary open/closed indicator or as a continuous gripper width. EEF state is typically obtained from the robot's internal state feedback.

\textit{Joint Angle} describes the robot configuration in joint space. A single measurement frame is represented as a vector of $J$ joint-angle positions. Joint angles provide essential proprioceptive information for trajectory optimization and safe manipulation under kinematic constraints. They are usually read from the robot's internal state feedback.

\textit{Object Pose} describes the 6D pose of an object in a camera or world frame. Following the common definition~\cite{wen2024foundationpose}, each object pose consists of a 3D position and an orientation that specify the transformation from the object coordinate frame to a reference frame, such as the camera or world frame. Object pose provides compact task-relevant spatial information, enabling object-centric manipulation. These poses are commonly obtained from motion capture systems or vision-based 6D pose estimation methods such as FoundationPose~\cite{wen2024foundationpose}. 

\vspace{-6pt}
\subsubsection{Potential Modalities}

\textit{Audio} provides acoustic cues about robot-object-environment interactions. The signal can be represented either as a one-dimensional waveform over time or transformed into a spectrogram with temporal and frequency dimensions. Audio captures interaction-related cues, such as collision events, material properties, and task progress during tasks such as liquid pouring, supporting contact-rich manipulation when visual observations are unavailable or insufficient.

\textit{Task Indicator} provides discrete high-level signals that specify which task a policy should execute. It is commonly used in multitask policy learning to condition a unified policy on different task objectives while maintaining a shared multimodal feature space. For example, VT-HMD~\cite{ye2026visual} trains a unified multitask manipulation policy conditioned on a task-specific one-hot identifier, enabling the same policy to switch among different manipulation tasks.

\subsection{Multimodal Representation Learning}
\label{subsection: Multimodal Representation Learning}

\begin{table}[!t]
  \setlength{\tabcolsep}{3.2pt}
  \renewcommand{\arraystretch}{1.18}
  \centering
  \footnotesize
  \caption{Summary of modality usage, fusion strategies, and representation learning in force- and tactile-aware robot learning methods.}
  \label{tab:modality_fusion_summary}
  \resizebox{\textwidth}{!}{%
  \begin{tabular}{l|c|ccc|ccc|ccc|ccc|c|c|ccccc}
    \toprule
    \multicolumn{1}{c|}{\cellcolor{headmethod}\textbf{Method}} &
    \multicolumn{14}{c|}{\cellcolor{headmodal}\textbf{Input Modalities and Use Phase}} &
    \multicolumn{1}{c|}{\cellcolor{headrep}\makecell{\textbf{Representation}\\\textbf{Learning}}} &
    \multicolumn{5}{c}{\cellcolor{headfusion}\textbf{Fusion Method}}\\
    \multicolumn{1}{c|}{\cellcolor{tablesubhead}} &
    \multicolumn{1}{c|}{\cellcolor{tablesubhead}\textbf{Language (L)}} &
    \multicolumn{3}{c|}{\cellcolor{tablesubhead}\textbf{Vision (V)}} &
    \multicolumn{3}{c|}{\cellcolor{tablesubhead}\textbf{Force/Torque (F)}} &
    \multicolumn{3}{c|}{\cellcolor{tablesubhead}\textbf{Tactile (T)}} &
    \multicolumn{3}{c|}{\cellcolor{tablesubhead}\textbf{State (S)}} &
    \multicolumn{1}{c|}{\cellcolor{tablesubhead}\textbf{Other}} &
    \multicolumn{1}{c|}{\cellcolor{tablesubhead}} &
    \multicolumn{5}{c}{\cellcolor{tablesubhead}}\\
    \multicolumn{1}{c|}{\cellcolor{tablecolhead}} &
    \cellcolor{tablecolhead}\makecell{Instruction} &
    \cellcolor{tablecolhead}\makecell{Image} &
    \cellcolor{tablecolhead}\makecell{Depth} &
    \cellcolor{tablecolhead}\makecell{Point\\Cloud} &
    \cellcolor{tablecolhead}\makecell{EEF\\Wrench} &
    \cellcolor{tablecolhead}\makecell{Joint\\Torque} &
    \cellcolor{tablecolhead}\makecell{Tactile-\\synth.} &
    \cellcolor{tablecolhead}\makecell{Visual-\\Tactile} &
    \cellcolor{tablecolhead}\makecell{Tactile\\Matrix} &
    \cellcolor{tablecolhead}\makecell{Contact\\Points} &
    \cellcolor{tablecolhead}\makecell{EEF\\Pose} &
    \cellcolor{tablecolhead}\makecell{Joint\\Angle} &
    \cellcolor{tablecolhead}\makecell{Object\\Pose} &
    \cellcolor{tablecolhead} &
    \cellcolor{tablecolhead} &
    \cellcolor{tablecolhead}\makecell{FiLM} &
    \cellcolor{tablecolhead}\makecell{Attention\\Cond.} &
    \cellcolor{tablecolhead}\makecell{MoE} &
    \cellcolor{tablecolhead}\makecell{Gated\\Filter} &
    \cellcolor{tablecolhead}\makecell{Others}\\
    \hline
    \rowcolor{tablerowA} UPPFC~\cite{zhi2025learningunifiedpolicyposition} &  & P1 &  &  & P1 &  &  &  &  &  &  & P1, P2 &  &  &  & \yes & \yes &  &  & \\
    \rowcolor{tablerowB} FCLM~\cite{portela2024learning} &  &  &  &  &  &  &  &  &  &  &  & P2, P3 &  &  &  &  & \yes &  &  & \\
    \rowcolor{tablerowA} FORGE~\cite{noseworthy2025forge} &  &  &  &  & P1 &  &  &  &  &  & P1, P3 &  &  &  &  &  & \yes &  &  & \\
    \rowcolor{tablerowB} TA-VLA~\cite{zhang2025ta} & P1 & P1 &  &  &  & P1 &  &  &  &  &  & P1 &  &  &  &  & \yes &  &  & \\
    \rowcolor{tablerowA} M3L~\cite{sferrazza2024power} &  & P1 &  &  &  &  &  & P1 &  &  &  &  &  &  & MAE &  & \yes &  &  & \\
    \rowcolor{tablerowB} ACP~\cite{hou2025adaptive} &  & P1 &  &  & P1, P3 &  &  &  &  &  & P1 &  &  &  &  & \yes & \yes &  &  & \\
    \rowcolor{tablerowA} AdmitDiff~\cite{zhou2025admittance} &  & P1 &  &  & P1, P3 &  &  &  &  &  & P1 & P1 &  &  &  & \yes & \yes &  &  & \\
    \rowcolor{tablerowB} HIL-SERL~\cite{luo2025precise} &  & P1 &  &  & P1 &  &  &  &  &  & P1, P3 &  &  &  &  &  & \yes &  &  & \\
    \rowcolor{tablerowA} Bi-ACT~\cite{buamanee2024bi} &  & P1 &  &  &  & P1 &  &  &  &  &  & P1 &  &  &  &  & \yes &  &  & \\
    \rowcolor{tablerowB} DexForce~\cite{chen2025dexforce} &  & P1 &  &  & P1 &  &  &  &  &  & P3 &  &  &  &  & \yes & \yes &  &  & \\
    \rowcolor{tablerowA} FoAR~\cite{he2025foar} &  & P1 & P1 & P1 & P1, P2 &  &  &  &  &  & P2 &  &  &  &  & \yes & \yes &  & \yes & \\
    \rowcolor{tablerowB} ForceMimic~\cite{liu2025forcemimic} &  & P1 & P1 & P1 & P3 &  &  &  &  &  &  &  & P1 &  &  & \yes & \yes &  &  & \\
    \rowcolor{tablerowA} DIPCOM~\cite{aburub2026learning} &  & P1 &  &  & P1 &  &  &  &  &  & P1 &  &  &  &  &  & \yes &  &  & \\
    \rowcolor{tablerowB} Comp-ACT~\cite{kamijo2024learning} &  & P1 &  &  & P1 &  &  &  &  &  & P1 &  &  &  &  &  & \yes &  &  & \\
    \rowcolor{tablerowA} HATO~\cite{lin2025learning} &  & P1 & P1 &  &  &  &  &  &  & P1 & P1 &  &  &  &  & \yes & \yes &  &  & \\
    \rowcolor{tablerowB} RDP~\cite{xue2025reactive} &  & P1 &  &  & P2 & P1 &  & P1, P2 &  &  & P1 &  &  &  &  & \yes & \yes &  &  & \\
    \rowcolor{tablerowA} DWF~\cite{kang2025robotic} &  & P1 &  &  &  & P1 &  &  &  &  & P1 &  &  &  &  & \yes & \yes &  &  & \\
    \rowcolor{tablerowB} TacDiffusion~\cite{wu2025tacdiffusion} &  &  &  &  & P1 & P1 &  &  &  &  & P1, P3 & P3 &  &  &  &  & \yes &  &  & \\
    \rowcolor{tablerowA} FARM~\cite{helmut2025tactile} &  & P1 &  &  &  &  &  & P1 &  &  & P1, P3 &  &  &  &  & \yes & \yes &  &  & \\
    \rowcolor{tablerowB} ViTacFormer~\cite{heng2025vitacformer} &  & P1 &  &  &  &  &  & P1 &  &  &  & P1 &  &  &  &  & \yes &  &  & \\
    \rowcolor{tablerowA} 3D-ViTac~\cite{huang20243d} &  &  &  & P1 &  &  &  & P1 &  &  &  & P1 &  &  &  & \yes & \yes &  &  & \\
    \rowcolor{tablerowB} UniT~\cite{xu2025unit} &  & P1 &  &  &  &  &  & P1 &  &  &  & P1 &  &  & VQR & \yes & \yes &  &  & \\
    \rowcolor{tablerowA} FuSe~\cite{jones2025beyond} & P1 & P1 &  &  &  &  &  & P1 &  &  &  & P1 &  & P1 (Audio) & MCA+AR &  & \yes &  &  & \\
    \rowcolor{tablerowB} ForceVLA~\cite{yu2025forcevla} & P1 & P1 &  &  & P1 & P1 &  &  &  &  & P1 &  &  &  &  &  &  & \yes &  & \\
    \rowcolor{tablerowA} Tactile-VLA~\cite{huang2025tactile} & P1 & P1 &  &  &  &  &  & P1 &  &  & P1 &  &  &  &  &  & \yes &  &  & \\
    \rowcolor{tablerowB} TLA~\cite{hao2025tla} & P1 &  &  &  &  &  &  & P1 &  &  &  &  &  &  &  &  & \yes &  &  & \\
    \rowcolor{tablerowA} FILIC~\cite{ge2025filic} &  & P1 &  &  &  & P1 &  &  &  &  &  & P1, P3 &  &  &  &  & \yes &  &  & \\
    \rowcolor{tablerowB} ManipForce~\cite{lee2025manipforce} &  & P1 &  &  & P1 &  &  &  &  &  &  &  &  &  &  &  & \yes &  &  & \\
    \rowcolor{tablerowA} OGRL-VT~\cite{shirai2025sim} &  & P1 &  &  & P1 &  &  &  &  &  & P1, P3 &  & P1, P3 &  &  &  &  &  &  & SC*\\
    \rowcolor{tablerowB} ViTaL~\cite{zhao2025touch} & P1 & P1, P2 &  &  &  &  &  &  & P1, P2 &  &  &  &  &  &  &  & \yes &  &  & \\
    \rowcolor{tablerowA} TACT~\cite{murooka2025tact} &  & P1 &  &  & P1, P2 &  &  &  &  & P1 & P1, P2 & P1, P2 &  &  &  &  & \yes &  &  & \\
    \rowcolor{tablerowB} VT-HMD~\cite{ye2026visual} & P1 & P1 &  &  &  &  &  &  &  & P1 &  & P1 &  & P1 (TI\textsuperscript{*}) & MAE &  & \yes &  &  & \\
    \rowcolor{tablerowA} TaF-VLA~\cite{huang2026tactile} & P1 & P1 &  &  & P1 & P1 &  & P1 &  &  & P1 &  &  &  & VQR &  & \yes &  &  & \\
    \rowcolor{tablerowB} Force-Policy~\cite{fang2026force} &  & P1, P2 & P1 &  & P1, P2 &  &  &  &  &  & P1, P2 &  &  &  &  & \yes &  &  & \yes & \\
    \rowcolor{tablerowA} FACTR~\cite{liu2025factr} &  & P1 &  &  &  & P1 &  &  &  &  &  & P1 &  &  &  &  & \yes &  &  & \\
    \rowcolor{tablerowB} MFMIL~\cite{ablett2024multimodal} &  & P1 &  &  &  &  &  & P1 &  &  & P1 &  &  &  &  &  &  &  &  & SC*\\
    \rowcolor{tablerowA} ForceSight~\cite{collins2024forcesight} & P1 & P1 & P1, P2 &  & P1, P2 &  &  &  &  &  & P1, P2 &  &  &  &  &  & \yes &  &  & \\
    \rowcolor{tablerowB} CRAFT~\cite{zhang2026craft} & P1 & P1 &  &  &  & P1 &  &  &  &  &  & P3 &  &  &  &  & \yes &  &  & \\
    \rowcolor{tablerowA} EquiContact~\cite{seo2025equicontact} & P1 & P1 &  &  & P1, P3 &  &  &  &  &  & P1, P3 &  &  &  &  & \yes &  &  &  & \\
    \rowcolor{tablerowB} MoDE-VLA~\cite{tang2026towards} & P1 & P1 &  &  & P1 & P1 &  &  &  & P1 &  & P1 &  &  &  &  & \yes & \yes &  & \\
    \rowcolor{tablerowA} ForceVLA2~\cite{li2026forcevla2} & P1 & P1 &  &  & P1, P3 &  &  &  &  &  & P1 & P1 &  &  &  &  & \yes & \yes &  & \\
    \rowcolor{tablerowB} ReTac-ACT~\cite{ruan2026retac} &  & P1 &  &  &  &  &  & P1 &  &  & P1 & P1 &  &  & MCA+AR &  & \yes &  & \yes & \\
    \rowcolor{tablerowA} TacVLA~\cite{zhang2026tacvla} & P1 & P1 &  &  &  &  &  &  & P1 &  & P1 &  &  &  &  &  & \yes &  & \yes & \\
    \rowcolor{tablerowB} KineDex~\cite{zhang2025kinedex} &  & P1 &  &  &  &  & P1 &  & P1 &  & P1 & P1, P2 &  &  &  &  & \yes &  &  & \\
    \rowcolor{tablerowA} OmniVTA~\cite{zheng2026omnivta} &  & P1 &  &  &  &  &  & P1, P2 &  &  & P1, P2 &  &  & & AR & \yes & \yes &  & \yes & \\
    \rowcolor{tablerowB} OmniVTLA~\cite{cheng2025omnivtla} & P1 & P1 &  &  &  &  &  &  & P1 & P1 & P1 &  &  &  &  &  & \yes &  &  & \\
    \rowcolor{tablerowA} ViTaS~\cite{tian2026vitas} & P1 &  &  &  &  &  &  & P1 &  &  &  &  &  &  & MCA+AR &  & \yes &  &  & \\
    \rowcolor{tablerowB} VLA-Touch~\cite{bi2025vla} & P1 & P1, P2 &  &  &  &  & P1, P2 & P1 &  &  & P1,P2,P3 &  &  &  &  &  & \yes &  &  & \\
    \rowcolor{tablerowA} 3DTacDex~\cite{wu2025canonical} &  & P1 &  &  &  &  & P1 &  &  & P1 & P1 & P1 &  &  & MAE &  & \yes &  &  & \\
    \rowcolor{tablerowB} VTAO-BiManip~\cite{sun2025vtao} &  & P1 &  &  &  &  &  &  &  & P1 &  &  & P1 &  & MAE &  & \yes &  &  & \\
    \rowcolor{tablerowA} VTLA~\cite{zhang2025vtla} & P1 & P1 &  &  &  &  &  & P1 &  &  &  &  &  &  &  &  & \yes &  &  & \\
    \rowcolor{tablerowB} ARCH~\cite{sun2024arch} &  & P1 &  &  & P1 &  &  &  &  &  & P1 &  & P1 &  &  & \yes & \yes &  &  & \\
    \rowcolor{tablerowA} TouchGuide~\cite{zhang2026touchguide} & P1 & P1 &  &  & P1 &  &  &  & P1 &  & P1 &  &  &  &  &  &  &  &  & DS*\\
    \bottomrule
  \end{tabular}%
  }
  \vspace{2pt}
  \par\noindent\begingroup\footnotesize\raggedright
  \textbf{Note.} Empty cells denote not reported or not used. P1, P2, and P3 denote different use phases defined in this survey. \textsuperscript{*}TI denotes task indicator. SC denotes simple concatenation, and DS denotes diffusion step.
  \par\endgroup
\end{table}

Multimodal perception can provide rich complementary sensory information at different stages of task execution. However, effectively utilizing and encoding these modalities through multimodal representation learning is still challenging. Desirable representations should satisfy two requirements: (1) semantically related observations from diverse modalities should be close in a shared latent space, and (2) each representation should preserve sufficient fine-grained, modality-specific information for downstream manipulation tasks~\cite{guo2019deep,liang2024foundations,zong2024self}.

To learn multimodal representations that are both informative and semantically aligned, existing methods often introduce reconstruction- and alignment-based auxiliary objectives. Reconstruction encourages encoded representations to preserve task-relevant modality information by recovering raw observations~\cite{zhang2025ta, xu2025unit, ye2026visual, huang2026tactile, ruan2026retac, tian2026vitas, cheng2025omnivtla, wu2025canonical, sun2025vtao, he2022masked, devlin2019bert}. In contrast, alignment projects diverse modalities into a shared latent space and pulls together observations associated with the same semantic concept, physical state, or task event~\cite{jones2025beyond, ruan2026retac, tian2026vitas, oord2018representation, radford2021learning, he2020momentum, chen2020simple, jia2021scaling}. Together, these objectives reduce modality gaps and provide informative representations that allow downstream policies to better exploit complementary multimodal cues.

We group these methods into Auxiliary Reconstruction (AR), Masked Autoencoding (MAE), Vector Quantization (VQ)-based Reconstruction (VQR), and Multimodal Contrastive Alignment (MCA).

\subsubsection{Auxiliary Reconstruction (AR)}
\label{subsec: Auxiliary Reconstruction}

Auxiliary reconstruction requires encoded features to preserve modality-specific and task-relevant information. Instead of optimizing only a task-specific objective, such as trajectory imitation loss, it asks the model to retain information sufficient to recover the original observation~\cite{kingma2013auto, rezende2014stochastic, wu2018multimodal, ruan2026retac}. This is particularly important for fine-grained or task-critical modalities that may be overlooked in end-to-end policy learning. For example, ReTac-ACT~\cite{ruan2026retac} reconstructs raw tactile inputs from tactile latent tokens, encouraging the tactile encoder to capture contact geometry and deformation patterns rather than collapsing into generic visual features.

Given the encoded feature of modality $m$, denoted as $\mathbf{F}^m$, a modality-specific decoder $D_m(\cdot)$ reconstructs the corresponding raw observation $\mathbf{X}^m$. The auxiliary reconstruction objective can be written as
\begin{equation}
\mathcal{L}_{\mathrm{rec}}^m
=
\left\|
\mathbf{X}^m - D_m(\mathbf{F}^m)
\right\|_2^2 ,
\end{equation}
where $\mathbf{X}^m$ is the original observation of modality $m$, and $D_m(\mathbf{F}^m)$ is its reconstruction from the encoded feature.

\vspace{-6pt}
\subsubsection{Masked Autoencoding (MAE)}
\label{subsec: Masked Autoencoding}

Masked Autoencoding (MAE) reconstructs only the missing parts of the input rather than the full observation from encoded features. It first masks a subset of input tokens and then requires the model to infer the masked content from the remaining visible tokens. By solving this incomplete-observation prediction problem, MAE encourages the representation to capture both intra-modal structure and inter-modal complementarity, such as spatial dependencies within each modality and cross-modal correlations between vision and touch~\cite{he2022masked, tong2022videomae, bao2021beit, sferrazza2024power}. For example, M3L~\cite{sferrazza2024power} uses multimodal MAE to reconstruct visual pixels and tactile taxels, learning compact vision-tactile representations for reinforcement learning.

For modality $m$, let $\Omega_m$ denote the set of masked token indices, and let $\mathbf{x}_j$ and $\hat{\mathbf{x}}_j$ denote the original and reconstructed token at index $j$, respectively. The MAE loss is computed only over the masked tokens:
\begin{equation}
    \mathcal{L}_{\mathrm{MAE}}^m
    =
    \frac{1}{|\Omega_m|}
    \sum_{j\in\Omega_m}
    \left\|
    \mathbf{x}_j-\hat{\mathbf{x}}_j
    \right\|_2^2 .
\end{equation}
Here, $|\Omega_m|$ is the number of masked tokens. In multimodal settings, the reconstruction can be conditioned on visible tokens from the same modality as well as complementary visible tokens from other modalities. Intuitively, MAE forces the model to infer missing sensory information from context. This makes the learned representation less dependent on isolated local cues and more sensitive to structural regularities and cross-modal relationships.

\vspace{-6pt}
\subsubsection{VQ-based Reconstruction (VQR)}
\label{subsec: VQ-based reconstruction}

Vector Quantization (VQ)-based reconstruction learns structured multimodal representations by mapping continuous features to a finite set of learnable codebook embeddings. This quantization encourages compact and noise-robust representations, which are valuable for high-dimensional and sensor-dependent modalities such as tactile signals~\cite{van2017neural, yu2021vector, razavi2019generating, xu2025unit}. For example, UniT~\cite{xu2025unit} uses a VQGAN autoencoder to learn tactile representations from GelSight images. By quantizing tactile features into discrete codes, VQ regularization preserves salient contact geometry and force patterns while making the latent space more robust to sensor noise.

Given continuous latent tokens $\mathbf{F}^m = \{\mathbf{f}_j\}_{j=1}^{N_m}$ for modality $m$, VQ maps each token to its nearest codebook entry:
\begin{equation}
    k_j=\arg\min_k\left\|\mathbf{f}_j-\mathbf{c}_k\right\|_2^2,
    \quad
    \mathbf{q}_j=\mathbf{c}_{k_j}.
\end{equation}
where $\mathbf{f}_j$ is the $j$-th continuous latent token, $\mathbf{c}_k$ is the $k$-th learnable codebook embedding, $k_j$ is the selected code index, and $\mathbf{q}_j$ is the corresponding quantized token. The quantized tokens are then decoded to reconstruct the original sensory input. The VQ objective is introduced to stabilize the discrete latent space:
\begin{equation}
    \mathcal{L}_{\mathrm{VQ}}^m
    =
    \left\|
    \mathrm{sg}[\mathbf{F}^m]-\mathbf{Q}^m
    \right\|_2^2
    +
    \beta
    \left\|
    \mathbf{F}^m-\mathrm{sg}[\mathbf{Q}^m]
    \right\|_2^2
\end{equation}
Here, $\mathbf{Q}^m$ denotes the quantized representation formed by all quantized tokens, $\mathrm{sg}[\cdot]$ is the stop-gradient operation, and $\beta$ controls the strength of the commitment term. The first term updates the codebook embeddings toward the encoder outputs, while the second term encourages the encoder outputs to stay close to their selected discrete codes. In practice, this VQ objective is combined with a reconstruction loss so that the discrete codes remain both stable and informative.

\vspace{-6pt}
\subsubsection{Multimodal Contrastive Alignment (MCA)}
\label{subsec: Multimodal Contrastive Alignment}

Multimodal contrastive alignment aims to learn a shared latent space for diverse sensory modalities. Instead of reconstructing inputs, it pulls together cross-modal representations associated with the same semantic concept, physical state, or task event while pushing apart mismatched pairs~\cite{jones2025beyond, ruan2026retac, tian2026vitas, radford2021learning, he2020momentum, chen2020simple, jia2021scaling}. It is essential in robotics because different sensors often observe the same interaction from complementary perspectives. For example, FuSe~\cite{jones2025beyond} applies CLIP-style contrastive learning to align vision, touch, and audio with language instructions, using language as a shared semantic grounding space.

Given two modalities $a$ and $b$, their encoded features are first pooled and projected into a shared latent space, producing $\mathbf{z}^a$ and $\mathbf{z}^b$. For a minibatch of $B$ paired samples, the matched pair $(\mathbf{z}_i^a,\mathbf{z}_i^b)$ is treated as positive, while mismatched pairs $(\mathbf{z}_i^a,\mathbf{z}_j^b)$ with $j\neq i$ are treated as negatives. The contrastive loss from modality $a$ to modality $b$ is
\begin{equation}
    \mathcal{L}_{a\rightarrow b}
    =
    -\frac{1}{B}
    \sum_{i=1}^{B}
    \log
    \frac{
    \exp(\mathrm{sim}(\mathbf{z}_i^a,\mathbf{z}_i^b)/\tau)
    }{
    \sum_{j=1}^{B}
    \exp(\mathrm{sim}(\mathbf{z}_i^a,\mathbf{z}_j^b)/\tau)
    },
\end{equation}
where $\mathrm{sim}(\cdot,\cdot)$ denotes cosine similarity and $\tau$ is the temperature parameter controlling the sharpness of similarity comparison.

A symmetric objective is commonly used so that both modalities are aligned toward each other:
\vspace{-5pt}
\begin{equation}
    \mathcal{L}_{\mathrm{align}}^{a,b}
    =
    \frac{1}{2}
    (\mathcal{L}_{a\rightarrow b}+\mathcal{L}_{b\rightarrow a}).
\end{equation}

\vspace{-5pt}
Intuitively, MCA teaches different modalities to agree on what they are observing. It does not require each modality to reconstruct raw sensory details, but instead encourages representations from vision, touch, audio, force, or language to meet in a common semantic space when they describe the same interaction. MCA can also support low-resource modality adaptation, allowing a newly introduced modality to benefit from semantic structures learned from data-rich modalities during pretraining.

\subsubsection{Task-Relevant Supervision}

Among the surveyed methods that explicitly introduce representation-learning objectives, reconstruction- and alignment-based objectives are the most common. We also acknowledge that task-relevant supervised signals can directly inject manipulation semantics into multimodal representations. Such supervision can arise either from joint encoder training with action-generation objectives or from auxiliary pre-training objectives defined by structured task priors.

The most direct source of such supervision is the action-generation loss, which encourages multimodal representations to encode control-relevant physical information for producing precise motor commands. Representative examples include diffusion-denoising objectives in Diffusion Policy~\cite{chi2025diffusion}, flow-matching objectives in $\pi_0$~\cite{black2024pi}, and next-action-token prediction in OpenVLA~\cite{kimopenvla}. Since these losses are tied to the policy architecture itself, we discuss the corresponding action-generation models in detail in Section~\ref{S3}.

Beyond the primary action loss, recent works further introduce auxiliary objectives that provide structured task priors for multimodal representation learning. These supervision signals include gaze regions~\cite{song2026reconvla}, affordances~\cite{li2025coa,yu2026affordancevla}, keypoints~\cite{yuanrobopoint, lee2025molmoact}, masks and poses~\cite{tu2026sg}, and multitask objectives~\cite{zadeh2018multimodal, mo2024multimed}. Compared with raw sensory reconstruction, these targets are more directly related to manipulation; compared with full action trajectories, they are often easier to annotate, predict, or optimize. As a result, they can serve as interpretable intermediate variables for grounding multimodal representations. For example, ReconVLA predicts gaze regions of manipulated objects to promote target-centric visual grounding~\cite{song2026reconvla}, while MolmoAct predicts depth-aware perception tokens and spatial object traces to enhance 3D action reasoning before low-level control~\cite{lee2025molmoact}.

\subsection{Multimodal Fusion}
\label{subsection: multimodal fusion}

Representation-learning objectives such as reconstruction and contrastive alignment can produce informative modality-specific or shared representations. However, robot policies must further integrate these representations into task-level features for planning and control. This fusion process is non-trivial because different modalities offer distinct and complementary strengths that vary across tasks, execution phases, and environments. For instance, vision often supports global scene understanding, whereas tactile and force signals are crucial during local contact-rich interactions.

Accordingly, effective multimodal fusion should go beyond simple feature aggregation by suppressing redundant information, preserving complementary cues, and regulating modality selection and cross-modal interaction during action generation. Based on these design considerations, we organize existing methods into FiLM injection, attention conditioning, mixture-of-experts, and gated filtering. We then discuss additional potential fusion directions, highlighting possible trends beyond the tactile- and force-centric studies summarized in Table~\ref{tab:modality_fusion_summary}.

\vspace{-6pt}
\subsubsection{FiLM Injection}
\label{subsection: Film Injection}

FiLM injection provides a lightweight mechanism for incorporating multimodal information into action generation. Rather than directly concatenating all modality tokens, it summarizes multimodal representations into compact conditioning vectors and uses them to modulate intermediate action features through feature-wise scaling and shifting~\cite{perez2018film, isola2017image, chi2025diffusion}. This mechanism is particularly useful in diffusion-based robot policies, where noisy action tokens need to be denoised under task- and observation-dependent conditions.

Given noisy action tokens $\mathbf{A}_t$ at diffusion timestep $t$, FiLM generates a scale parameter $\boldsymbol{\gamma}^m$ and a shift parameter $\boldsymbol{\beta}^m$ from modality $m$, and modulates the action tokens as
\begin{equation}
\widetilde{\mathbf{A}}_t
=
\boldsymbol{\gamma}^m \odot \mathbf{A}_t
+
\boldsymbol{\beta}^m .
\end{equation}
Here, $\mathbf{A}_t$ denotes the noisy action tokens, $\widetilde{\mathbf{A}}_t$ denotes the modulated action tokens, and $\odot$ denotes element-wise multiplication. The scale and shift parameters are usually produced by a small network from pooled modality features, allowing the sensory context to directly control how action features are amplified, suppressed, or shifted.

\vspace{-6pt}
\subsubsection{Attention Conditioning}
\label{subsection: Attention Conditioning}

Attention conditioning performs token-level multimodal fusion through interactions between action tokens and modality-specific tokens. Unlike global conditioning mechanisms such as FiLM, attention allows action generation to selectively query task-relevant multimodal evidence. This facilitates fine-grained cross-modal reasoning, such as linking action steps to visual object positions, tactile contact patterns, or force-induced interaction states.

The core operation is scaled dot-product attention:
\vspace{-5pt}
\begin{equation}
\mathrm{Attn}(\mathbf{Q},\mathbf{K},\mathbf{V})
=
\mathrm{softmax}
\left(
\frac{\mathbf{Q}\mathbf{K}^{\top}}{\sqrt{d}}
\right)
\mathbf{V}.
\end{equation}
Here, $\mathbf{Q}$, $\mathbf{K}$, and $\mathbf{V}$ denote query, key, and value tokens, respectively, and $d$ is the feature dimension used for scaling. In self-attention-based conditioning, action tokens and modality tokens are concatenated into a single sequence, from which $\mathbf{Q}$, $\mathbf{K}$, and $\mathbf{V}$ are jointly computed~\cite{black2024pi, intelligence2025pi, driess2023palm}. In cross-attention conditioning, $\mathbf{Q}$ is computed from the action tokens, whereas $\mathbf{K}$ and $\mathbf{V}$ are computed from the modality tokens, allowing the action representation to retrieve relevant sensory information~\cite{chi2025diffusion, brohan2022rt, shridhar2023perceiver}. Different modalities can also be incorporated through distinct conditioning pathways, such as joint self-attention for image-action interaction, adaptive layer normalization for proprioception, and cross-attention for tactile signals~\cite{li2026deco}.

\vspace{-6pt}
\subsubsection{Mixture-of-Experts (MoE)}
\label{subsec: Mixture-of-Experts}

Instead of processing all multimodal features with a single shared fusion module, Mixture-of-Experts (MoE) dynamically routes tokens to a set of specialized expert networks. Through sparse routing, different tokens can be processed by different experts that specialize in particular modalities, task phases, or interaction patterns~\cite{yu2025forcevla, li2026forcevla2, du2022glam, fedus2022switch, li2025uni}. This property is well suited to robotic manipulation, where the relevance of each modality changes throughout execution. For example, in contact-rich insertion, visual and language tokens may be routed to experts that model object geometry and task semantics during the approach phase, whereas force and tactile tokens may be routed to contact-sensitive experts after contact is detected.

For each modality token $\mathbf{f}_j$, a router predicts expert weights and selects a small subset of experts. The updated token is then computed as a weighted combination of the selected expert outputs:
\begin{equation}
    \widetilde{\mathbf{f}}_j
    =
    \sum_{i\in\mathcal{S}_j}
    r_{j,i} E_i(\mathbf{f}_j).
\end{equation}
Here, $\mathcal{S}_j$ denotes the set of selected experts for token $\mathbf{f}_j$, $E_i(\cdot)$ is the $i$-th expert network, and $r_{j,i}$ is the routing weight assigned to that expert. The routing weights are typically produced by a softmax router, while only the top-ranked experts are activated for computational efficiency. MoE does not directly compute pairwise interactions between action tokens and modality tokens. Instead, it adaptively transforms modality features before they are injected into the policy through concatenation, FiLM, or attention.

\vspace{-6pt}
\subsubsection{Gated Filtering}

\label{subsec: Gated Filtering}

Gated filtering dynamically controls each modality's contribution to multimodal fusion according to the current task phase or interaction state. Rather than continuously injecting all multimodal features, it assigns state-dependent gates to modality representations, allowing the policy to emphasize useful information and suppress redundant or noisy information~\cite{he2025foar, fang2026force, ruan2026retac, zhang2026tacvla, zheng2026omnivta, arevalo2017gated}. This mechanism is particularly suitable for phase-dependent modalities. For example, tactile signals are informative after contact is established but may be uninformative or noisy during free-space motion. By filtering modality features before fusion, gating reduces cross-modal interference and encourages task-relevant sensory integration.

Given modality features $\mathbf{F}^m$ and a task-dependent state $\mathbf{s}_t$, gated filtering can be written as
\begin{equation}
\widetilde{\mathbf{F}}^m
=
\sigma(G_m(\mathbf{s}_t))
\odot
\mathbf{F}^m .
\end{equation}
Here, $G_m(\cdot)$ is a gating network for modality $m$, $\sigma(\cdot)$ is the sigmoid function that produces gate values between 0 and 1, and $\mathbf{s}_t$ denotes the current task-dependent state, such as the end-effector pose, contact state, or execution phase. The resulting $\widetilde{\mathbf{F}}^m$ is the filtered modality representation, which can then be fused into action tokens through concatenation, FiLM, or attention-based conditioning.

Intuitively, gated filtering works like a learned sensory switch. It does not assume that every modality is always useful; instead, it lets the policy decide when to attend to each modality and when to suppress it, which is especially important for contact-rich manipulation with phase-dependent sensory relevance.

\subsection{Discussions on Potential Directions}

\textit{Denoising-Stage Modality Composition} performs multimodal fusion at the denoising or sampling level for diffusion- or flow-matching-based robot policies. Instead of fusing different modality features into a single representation before action generation, it composes modality-specific guidance during the iterative denoising process~\cite{sun2024arch, chen2025multi}. This design is motivated by the observation that different modalities may provide complementary constraints at different sampling stages. For example, visual observations can guide coarse trajectory generation in early denoising steps, whereas tactile or force feedback can refine contact-sensitive constraints in later steps.

Let $\mathbf{A}_t^\tau$ denote the noisy action representation at denoising step $\tau$, and $\mathbf{F}^a$, $\mathbf{F}^b$ represent two modality features. During early denoising, the policy follows modality-$a$ guidance:
\begin{equation}
    \hat{\boldsymbol{\epsilon}}^\tau=\boldsymbol{\epsilon}_\theta(\mathbf{A}_t^\tau,\tau,\mathbf{F}^a),
    \quad
    \tau\in\mathcal{T}_{\mathrm{early}},
\end{equation}
\begin{equation}
    \mathbf{A}_t^{\tau-1}
    =
    \mathrm{Denoise}(\mathbf{A}_t^\tau,\hat{\boldsymbol{\epsilon}}^\tau,\tau).
\end{equation}
During later denoising, a modality-$b$ guidance model $S_\phi^b$ further provides a task-relevant guidance score and the denoising direction is then modified by the score gradient:
\begin{equation}
    s_\phi^b=S_\phi^b(\mathbf{A}_t^\tau,\mathbf{F}^a,\mathbf{F}^b).
\end{equation}
\begin{equation}
    \hat{\boldsymbol{\epsilon}}^\tau
    =
    \boldsymbol{\epsilon}_\theta(\mathbf{A}_t^\tau,\tau,\mathbf{F}^a)
    -
    \eta_\tau
    \nabla_{\mathbf{A}_t^\tau}s_\phi^b,
    \quad
    \tau\in\mathcal{T}_{\mathrm{late}},
\end{equation}
where $\eta_\tau$ is the guidance scale. The guided direction is then used to update $\mathbf{A}_t^\tau$ toward the final action sequence $\mathbf{A}_t^0$.

\textit{Balanced Multimodal Fusion} aims to mitigate modality dominance during multimodal policy learning~\cite{peng2022balanced, wei2024fly}. In multimodal optimization, highly predictive modalities such as vision or language can dominate gradient updates, causing weaker but task-critical modalities, such as tactile or force, to be insufficiently explored. This imbalance reduces the policy's ability to exploit complementary multimodal information, especially in contact-rich manipulation, where non-visual modalities may only be informative during specific interaction phases.

A representative strategy is to dynamically adapt the learning losses according to each modality's contribution. Instead of assigning the same optimization strength to all modality branches, balanced learning estimates modality dominance and adjusts gradient updates to suppress over-dominant modalities or enhance under-utilized ones:
\begin{equation}
\nabla_{\theta_m}\mathcal{L}
\leftarrow
\rho^m
\nabla_{\theta_m}\mathcal{L},
\quad
\rho^m
=
\Phi
\left(
s^1,s^2,\ldots,s^M
\right).
\end{equation}
Here, $\theta_m$ denotes the parameters of modality branch $m$, $s^m$ denotes the estimated contribution or dominance score of modality $m$, and $\rho^m$ is the corresponding gradient modulation coefficient produced by $\Phi(\cdot)$. In practice, $\rho^m$ can reduce the update strength of over-dominant modalities or increase the update strength of under-optimized modalities.

\textit{Foresight Modality Imagination} refers to predicting future multimodal observations or latent dynamics from historical multimodal features through a \textbf{world model}~\cite{hou2026world}. Unlike directly fusing only current observations, this strategy imagines how the physical interaction may evolve in the near future, such as whether slippage or collision may occur, or how tactile deformation may change. As a result, it can guide and refine action generation before actual collision or object deformation happens, thereby reducing the risk of unsafe or damaging interactions. For example, OmniVTA employs a visuo-tactile world model~\cite{zheng2026omnivta} to predict future latent tactile representations. The predicted tactile foresight is then passed to a reflexive tactile controller, which refines coarse actions by correcting deviations between predicted and observed tactile signals. In this way, the policy not only reacts to currently observed contact signals but also anticipates future contact evolution before it is fully observed.

Given historical multimodal features, a world model predicts future observations or latent states:
\begin{equation}
    \mathbf{X}^{\mathrm{fs}}_{t+1:t+H}
    =
    W_{\theta}
    \left(
    \{\mathbf{F}^{m}_{t-H:t}\}_{m=1}^{M}
    \right),
\end{equation}
where $H$ denotes the horizon length. The predicted foresight signal is then encoded into a foresight representation $\mathbf{F}^{\mathrm{fs}}_t$, which conditions the downstream policy to guide or refine action generation.

\section{Phase 1: Primary Action Generation} \label{S3}

The primary action policy, referred to here as the Phase-1 policy, serves as the core decision-making module in the proposed framework. Except for methods that explicitly address low-level control, most robot learning approaches rely on such a primary policy to map encoded and fused multimodal observation embeddings, which may be difficult to interpret intuitively, into robot actions or parameters with explicit physical meaning.

On the input side, the primary policy aggregates encoded features from Section~\ref{S2}. Depending on the available sensors and target task, the policy may take raw sensory inputs, fused multimodal representations, or task-relevant features that facilitate action generation. On the output side, it predicts motion trajectories for direct robot execution or generates coarse trajectories that can be further refined by a second-phase policy, as discussed in Section~\ref{S4}. Alternatively, the primary policy may produce outputs that guide downstream controllers, such as contact-force references and interaction-related control stiffness.

\begin{figure}[!t]
  \centering
  \includegraphics[width=\textwidth,height=0.90\textheight,keepaspectratio,trim=1cm 5.8cm 0.5cm 3.5cm,clip]{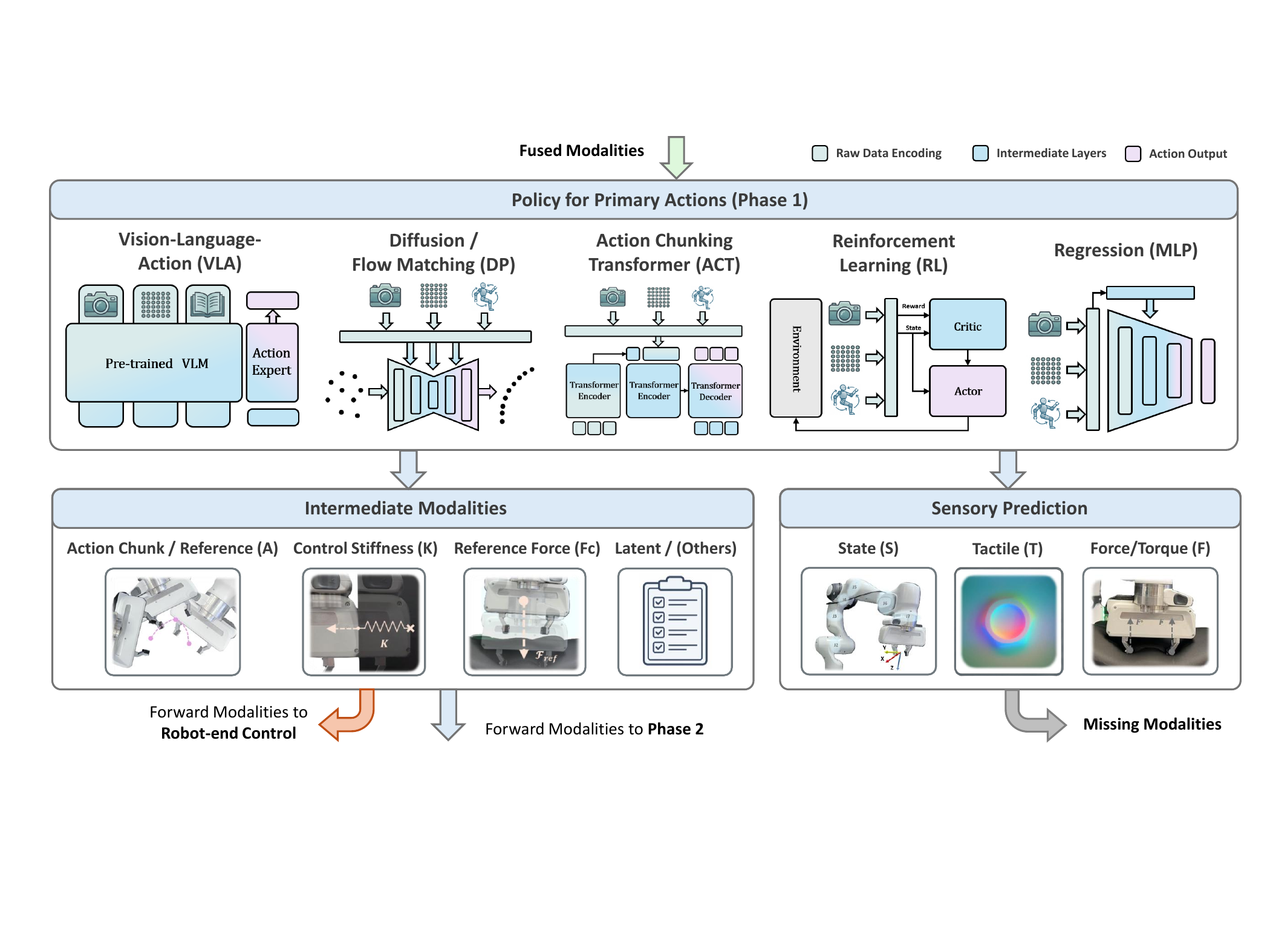}
  \caption{Overview of the Phase-1 primary action-generation policy, explicitly showing the input-output relationships and the common model architectures used in the surveyed papers.}
  \Description{}
  \label{Fig:phase1}
\end{figure}

Figure~\ref{Fig:phase1} provides an overview of the primary action policy, including abstract model architecture visualizations, representative input and output examples, and sensory prediction examples. Table~\ref{tab:model_arch_table} summarizes the surveyed papers according to the following entries: model architecture, action-space classification, intermediate modalities predicted in addition to actions, and missing-observation prediction.

\subsection{Architecture of Primary Policy}

In this survey, we categorize the major action-generation approaches in the surveyed papers into five classes: regression-based policies, reinforcement learning (RL), diffusion/flow-matching policy (DP), action chunking transformer (ACT), and vision-language-action (VLA) models.

\subsubsection{Regression-based Policies}

Regression-based policies formulate robotic action prediction as a supervised learning problem. Rather than modeling complex action distributions, they learn a deterministic mapping from sensory inputs to continuous control commands. This architecture typically uses feature-extraction backbones paired with an action prediction head to generate actions from multimodal observations. Given a multimodal observation feature $o_t$ at time step $t$, the policy $\pi$ predicts an action $\hat{a}_t$. The network is trained by minimizing the discrepancy between the predicted action and the ground-truth expert demonstration $a_t$, commonly using the mean squared error (MSE) loss:
\begin{equation}
    \mathcal{L}_{\text{MSE}} = \frac{1}{N}\sum_{i=1}^{N} \|\pi(o_{t,i}) - a_{t,i}\|_2^2
\end{equation}
Here, $N$ denotes the number of samples in the batch. This objective encourages the policy to reproduce expert actions accurately, providing a simple and effective policy learning baseline for direct position or force control~\citep{ye2026visual, ablett2024multimodal, collins2024forcesight}.

While regression-based policies benefit from architectural simplicity and high inference frequency suitable for real-time control, they struggle with multimodal action distributions. When multiple valid behaviors exist for the same observation, the model tends to average them, producing hesitant or physically infeasible actions.

\subsubsection{Reinforcement Learning (RL)}

RL learns optimal control strategies through trial-and-error interactions rather than relying solely on expert demonstrations. By formulating robot action prediction as a Markov Decision Process (MDP), RL can learn complex and contact-rich behaviors under highly nonlinear dynamics~\citep{kober2013reinforcement, dang_hydroshear_2026, xue_arraybot_2025, hu_dexterous_2023, barreiros_learning_2025, su_sim2real_2024, zhao_visual-tactile_2026}. Many modern RL methods for robotics adopt an actor-critic architecture, where a policy network, or actor, maps multimodal observations to actions, and a value network, or critic, estimates cumulative returns. To maximize the expected long-term return under an environment reward $R_t$, RL methods often use Proximal Policy Optimization (PPO)~\citep{schulman2017proximal}, which updates the policy by maximizing a clipped surrogate objective:
\begin{equation}
    J_{\text{PPO}}(\theta) = \hat{\mathbb{E}}_t \left[ \min\left(\rho_t(\theta)\hat{A}_t, \, \text{clip}(\rho_t(\theta), 1-\epsilon, 1+\epsilon)\hat{A}_t\right) \right]
\end{equation}
Here, $\rho_t(\theta)$ denotes the probability ratio of the action under the current policy to that under the previous policy, $\hat{A}_t$ is the estimated advantage function, and $\epsilon$ is the clipping threshold.

While RL has strong autonomous exploration capabilities, it suffers from severe sample inefficiency and often requires a large number of interaction steps to converge. Furthermore, its performance depends heavily on carefully engineered reward functions to prevent unintended physical behaviors.

\subsubsection{Diffusion/Flow Matching (DP)}

DP \citep{chi2025diffusion} formulates action generation as a conditional progressive denoising process~\citep{yuan_vtam_2026, li_flow_2026, geiger_diffusion-based_2026, wang_phaforce_2026, yue_pocodp3_2026, xue_tube_2026}.\footnote{We use DP to denote both the specific Diffusion Policy and the broader class of generative denoising policies, including flow matching. We acknowledge this overload; despite flow matching relying on vector-field regression instead of standard diffusion, both share the same progressive trajectory generation paradigm.} Unlike deterministic regression, DP transforms Gaussian noise into valid action trajectories through iterative refinement, allowing it to naturally model multimodal action distributions. This helps avoid mode averaging when multiple valid expert behaviors exist for the same sensory context. Given an observation prefix $o_t$, the policy initializes an action sequence as pure Gaussian noise $\epsilon$. A denoising network, typically implemented using a Transformer or time-conditional U-Net, then iteratively refines this noise. The network is trained using either a score-matching objective (for diffusion) or a vector-field matching objective (for flow matching). For instance, the standard diffusion noise-prediction loss is formulated as:
\begin{equation}
    \mathcal{L}_{\text{DP}} = \mathbb{E} \left[ \| \epsilon_\theta(x_\tau, \tau, o_t) - \epsilon \|^2 \right]
\end{equation}
Here, $\epsilon \sim \mathcal{N}(0, \mathbf{I})$ is the ground-truth random noise vector, $x_\tau$ is the noisy action sequence at diffusion timestep $\tau$, $o_t$ denotes the multimodal observation context, and $\epsilon_\theta(\cdot)$ is the noise predicted by the denoising network.
While generative denoising policies can generate trajectories that are robust to noisy demonstrations, the iterative denoising process introduces considerable inference latency. This computational cost makes direct deployment in high-frequency closed-loop control challenging.

\subsubsection{Action Chunking Transformer (ACT)}
 
ACT \citep{zhao2023learning} formulates action generation as a simultaneous sequence-generation problem. By predicting an action chunk in a single forward pass, ACT enforces temporal consistency and reduces the execution drift commonly observed in step-by-step autoregressive methods. The architecture typically pairs a Transformer encoder-decoder with a Conditional Variational Autoencoder (CVAE). During execution, the decoder uses a sampled latent variable $z$ together with the current observation $o_t$ to generate a continuous trajectory over a time horizon $H$. 
The model is trained by minimizing the negative evidence lower bound (ELBO), balancing reconstruction accuracy with latent regularization:\vspace{-3pt}
\begin{equation}
    \mathcal{L}_{\text{ACT}} = \sum_{h=0}^{H} \Vert{} \hat{a}_{t+h} - a_{t+h} \Vert{}_1 + \beta \mathcal{D}_{\text{KL}} \left( \mathcal{Q}(z \vert{} a_{t:t+H}, o_t) \parallel \mathcal{P}(z) \right)
\end{equation}
Here, $a_{t+h}$ represents the ground-truth action, $\mathcal{Q}$ is the CVAE posterior encoding the expert action chunk conditioned on the current observation, and $\mathcal{P}(z)$ is the standard normal prior. During inference, $z$ can be set to zero (the mean of the prior) for deterministic action generation. $\mathcal{D}_{\text{KL}}$ denotes the Kullback--Leibler divergence, and $\beta$ is a scaling hyperparameter~\citep{ye_learning_2026, kang_catch-form-acter_2025, kobayashi_bi-lat_2025, uriguen_eljuri_haptic-informed_2025, kobayashi_alpha-_2025, watanabe_ftact_2025}.

\subsubsection{Vision-Language-Action (VLA)}
 
The VLA model unifies semantic reasoning and low-level execution by grounding representations learned by large vision-language models (VLMs) into downstream action generation. This architecture can process multi-view images, tokenized language instructions, and proprioceptive states through a pre-trained VLM backbone. While some VLA formulations, such as RT-1~\citep{brohan2022rt}, RT-2~\citep{zitkovich2023rt}, and OpenVLA~\citep{kimopenvla}, rely on discretizing physical space to predict autoregressive action tokens, other architectures, such as $\pi_0$~\cite{black2024pi}, ForceVLA, and Tactile-VLA, decode the VLM's rich semantic embeddings directly into continuous robot actions through regression heads, diffusion generators, or action chunking modules. Because the VLA paradigm serves as a multimodal foundation rather than a fixed policy architecture, its training objective depends on the choice of action decoder. Consequently, the optimization objective varies with the generation mechanism: autoregressive cross-entropy (CE) for discrete action-token prediction, standard distance metrics such as $L_1$ or $L_2$ loss for continuous control, flow-matching or score-matching losses for diffusion-based generative modeling, and chunk-level reconstruction objectives for trajectory forecasting~\citep{zhao_fd-vla_2026, li_favla_2026, morissette_tactile_2026, lee_modular_2026, yang_direction_2026}.

VLA models provide strong open-vocabulary understanding and can improve zero-shot generalization, enabling robots to execute unconstrained instructions on unseen objects. However, the large parameter scale of VLM backbones causes substantial training and inference cost. Effectively bridging high-level semantic reasoning with the high-frequency, low-latency continuous control required for precision contact tasks remains a central architectural challenge.

\subsection{Output of Primary Policy}

After establishing the architecture of the primary policy, we next examine the policy outputs and their formats. These outputs, referred to here as \textit{intermediate modalities}, can generally be categorized into four groups: actions, force references, stiffness-control parameters, and auxiliary predictions.

\begin{table}[!t]
  \setlength{\tabcolsep}{3.2pt}
  \renewcommand{\arraystretch}{1.18}
  \centering
  \scriptsize
  \caption{Summary of Phase 1 model architecture choices, action spaces, action horizons, and auxiliary action or prediction outputs.}
  \label{tab:model_arch_table}
  \resizebox{\textwidth}{!}{%
  \begin{tabular}{l|>{\centering\arraybackslash}m{2.05cm}|>{\centering\arraybackslash}m{0.58cm}>{\centering\arraybackslash}m{0.58cm}>{\centering\arraybackslash}m{0.58cm}|>{\centering\arraybackslash}m{0.85cm}>{\centering\arraybackslash}m{0.65cm}|>{\centering\arraybackslash}m{1.35cm}|>{\centering\arraybackslash}m{1.35cm}|>{\scriptsize\centering\arraybackslash}m{3.65cm}|>{\centering\arraybackslash}m{2.00cm}}
    \toprule
    \multicolumn{1}{c|}{\cellcolor{headmethod}\textbf{Method}} &
    \multicolumn{1}{c|}{\cellcolor{headmodal}\textbf{Architecture}} &
    \multicolumn{5}{c|}{\cellcolor{headmethod}\textbf{Action (A)}} &
    \multicolumn{1}{>{\centering\arraybackslash}m{1.35cm}|}{\cellcolor{headmodal}\shortstack[c]{\textbf{Control}\\\textbf{Stiffness (K)}}} &
    \multicolumn{1}{>{\centering\arraybackslash}m{1.35cm}|}{\cellcolor{headmethod}\shortstack[c]{\textbf{Reference}\\\textbf{Force (Fc)}}} &
    \multicolumn{1}{>{\scriptsize\centering\arraybackslash}m{3.65cm}|}{\cellcolor{headmodal}\textbf{Other}} &
    \multicolumn{1}{>{\centering\arraybackslash}m{2.00cm}}{\cellcolor{headmethod}\shortstack[c]{\textbf{Sensory}\\\textbf{Prediction}}}\\
    \multicolumn{1}{c|}{\cellcolor{tablesubhead}} &
    \multicolumn{1}{c|}{\cellcolor{tablesubhead}} &
    \multicolumn{3}{c|}{\cellcolor{tablesubhead}\textbf{Action Space}} &
    \multicolumn{2}{c|}{\cellcolor{tablesubhead}\textbf{Action Horizon}} &
    \multicolumn{1}{c|}{\cellcolor{tablesubhead}} &
    \multicolumn{1}{c|}{\cellcolor{tablesubhead}} &
    \multicolumn{1}{c|}{\cellcolor{tablesubhead}} &
    \multicolumn{1}{c}{\cellcolor{tablesubhead}}\\
    \multicolumn{1}{c|}{\cellcolor{tablecolhead}} &
    \multicolumn{1}{c|}{\cellcolor{tablecolhead}} &
    \cellcolor{tablecolhead}\textbf{Delta} &
    \cellcolor{tablecolhead}\textbf{EEF} &
    \cellcolor{tablecolhead}\textbf{Joint} &
    \cellcolor{tablecolhead}\mbox{\textbf{Per-step}} &
    \cellcolor{tablecolhead}\textbf{Chunk} &
    \cellcolor{tablecolhead} &
    \cellcolor{tablecolhead} &
    \cellcolor{tablecolhead} &
    \cellcolor{tablecolhead}\\
    \midrule
    \rowcolor{tablerowA} TA-VLA~\cite{zhang2025ta} & VLA                           &           &        &\yes         &                  &\yes              &                                                                                      &                                                                                     &                          & Force                                   \\
    \rowcolor{tablerowB} FuSe~\cite{jones2025beyond} & VLA                           &\yes          &\yes       &          &                  &\yes              &                                                                                      &                                                                                     &                          &                                        \\
    \rowcolor{tablerowA} ForceVLA~\cite{yu2025forcevla} & VLA                           &           &\yes       &          &                  &\yes              &                                                                                      &                                                                                     &                          &                                        \\
    \rowcolor{tablerowB} ForceVLA2~\cite{li2026forcevla2} & VLA                           &\yes          &\yes       &          &                  &\yes              &                                                                                      &\yes                                                                                    & transition indicator      &                                        \\
    \rowcolor{tablerowA} Tactile-VLA~\cite{huang2025tactile} & VLA                           &           &\yes       &          &                  &\yes              &                                                                                      &\yes                                                                                    &                          &                                        \\
    \rowcolor{tablerowB} TLA~\cite{hao2025tla} & VLA                           &\yes          &        &          &                  &\yes              &                                                                                      &                                                                                     &                          &                                        \\
    \rowcolor{tablerowA} CRAFT~\cite{zhang2026craft} & VLA                           &\yes          &\yes       &\yes         &                  &\yes              &                                                                                      &                                                                                     &                          &                                        \\
    \rowcolor{tablerowB} TaF-VLA~\cite{huang2026tactile} & VLA                           &           &\yes       &          &                  &\yes              &                                                                                      &                                                                                     &                          &                                        \\
    \rowcolor{tablerowA} OmniVTLA~\cite{cheng2025omnivtla} & VLA                           &           &\yes       &\yes         &                  &\yes              &                                                                                      &                                                                                     &                          &                                        \\
    \rowcolor{tablerowB} VLA-Touch~\cite{bi2025vla} & VLA                           &           &\yes       &          &                  &\yes              &                                                                                      &                                                                                     &                          &                                        \\
    \rowcolor{tablerowA} VTLA~\cite{zhang2025vtla} & VLA                           &           &\yes       &          &\yes                 &               &                                                                                      &                                                                                     &                          &                                        \\
    \rowcolor{tablerowB} TacVLA~\cite{zhang2026tacvla} & VLA                           &\yes          &\yes       &\yes         &                  &\yes              &                                                                                      &                                                                                     &                          &                                        \\
    \rowcolor{tablerowA} MoDE-VLA~\cite{tang2026towards} & VLA, RL                       &\yes          &        &\yes         &                  &\yes              &                                                                                      &                                                                                     &                          & State                                   \\
    \rowcolor{tablerowB} FORGE~\cite{noseworthy2025forge} & RL                            &           &\yes       &          &\yes                 &               &                                                                                      &                                                                                     & early termination action  &                                        \\
    \rowcolor{tablerowA} M3L~\cite{sferrazza2024power} & RL                            &\yes          &\yes       &          &\yes                 &               &                                                                                      &                                                                                     &                          &                                        \\
    \rowcolor{tablerowB} HIL-SERL~\cite{luo2025precise} & RL                            &           &\yes       &          &\yes                 &               &                                                                                      &\yes                                                                                    &                          &                                        \\
    \rowcolor{tablerowA} OGRL-VT~\cite{shirai2025sim} & RL                            &           &\yes       &          &\yes                 &               &                                                                                      &                                                                                     &                          & State                                   \\
    \rowcolor{tablerowB} VTAO-BiManip~\cite{sun2025vtao} & RL                            &           &        &\yes         &\yes                 &               &                                                                                      &                                                                                     &                          &                                        \\
    \rowcolor{tablerowA} ViTaS~\cite{tian2026vitas} & RL, DP                        &\yes          &\yes       &\yes         &\yes                 &\yes              &                                                                                      &                                                                                     &                          &                                        \\
    \rowcolor{tablerowB} ARCH~\cite{sun2024arch} & RL, MP\textsuperscript{1}                        &           &\yes       &          &\yes                 &\yes              &                                                                                      &                                                                                     &                          &                                        \\
    \rowcolor{tablerowA} VT-HMD~\cite{ye2026visual} & RL, Regression                &           &        &\yes         &\yes                 &               &                                                                                      &                                                                                     &                          &                                        \\
    \rowcolor{tablerowB} UPPFC~\cite{zhi2025learningunifiedpolicyposition} & DP                            &           &        &\yes         &\yes                 &               &                                                                                      & \yes                                                                                    &                          & Force\textsuperscript{*}                \\
    \rowcolor{tablerowA} ACP~\cite{hou2025adaptive} & DP                            &           &\yes       &          &                  &\yes              &\yes                                                                                     &                                                                                     &                          &                                        \\
    \rowcolor{tablerowB} AdmitDiff~\cite{zhou2025admittance} & DP                            &           &\yes       &          &                  &\yes              &                                                                                      &\yes                                                                                    &                          &                                        \\
    \rowcolor{tablerowA} DexForce~\cite{chen2025dexforce} & DP                            &           &\yes       &          &                  &\yes              &                                                                                      &                                                                                     &                          &                                        \\
    \rowcolor{tablerowB} FoAR~\cite{he2025foar} & DP                            &           &\yes       &          &                  &\yes              &                                                                                      &                                                                                     &                          &                                        \\
    \rowcolor{tablerowA} ForceMimic~\cite{liu2025forcemimic} & DP                            &           &        &\yes         &                  &\yes              &                                                                                      &\yes                                                                                    &                          &                                        \\
    \rowcolor{tablerowB} DIPCOM~\cite{aburub2026learning} & DP                            &           &\yes       &          &                  &\yes              &\yes                                                                                     &                                                                                     &                          &                                        \\
    \rowcolor{tablerowA} HATO~\cite{lin2025learning} & DP                            &           &        &\yes         &                  &\yes              &                                                                                      &                                                                                     &                          &                                        \\
    \rowcolor{tablerowB} RDP~\cite{xue2025reactive} & DP                            &           &        &          &                  &               &                                                                                      &                                                                                     & latent action chunk       &                                        \\
    \rowcolor{tablerowA} DWF~\cite{kang2025robotic} & DP                            &\yes          &        &          &                  &\yes              &                                                                                      &                                                                                     &                          &                                        \\
    \rowcolor{tablerowB} TacDiffusion~\cite{wu2025tacdiffusion} & DP                            &           &\yes       &          &\yes                 &               &                                                                                      &\yes                                                                                    &                          &                                        \\
    \rowcolor{tablerowA} FARM~\cite{helmut2025tactile} & DP                            &           &\yes       &          &                  &\yes              &                                                                                      &\yes                                                                                    &                          &                                        \\
    \rowcolor{tablerowB} 3D-ViTac~\cite{huang20243d} & DP                            &           &        &\yes         &                  &\yes              &                                                                                      &                                                                                     &                          &                                        \\
    \rowcolor{tablerowA} UniT~\cite{xu2025unit} & DP                            &           &        &\yes         &                  &\yes              &                                                                                      &                                                                                     &                          &                                        \\
    \rowcolor{tablerowB} ManipForce~\cite{lee2025manipforce} & DP                            &\yes          &\yes       &          &                  &\yes              &                                                                                      &                                                                                     &                          &                                        \\
    \rowcolor{tablerowA} Force-Policy~\cite{fang2026force} & DP                            &           &\yes       &          &                  &\yes              &                                                                                      &                                                                                     & global feature            &                                        \\
    \rowcolor{tablerowB} KineDex~\cite{zhang2025kinedex} & DP                            &           &        &\yes         &                  &\yes              &                                                                                      &\yes                                                                                    &                          &                                        \\
    \rowcolor{tablerowA} OmniVTA~\cite{zheng2026omnivta} & DP                            &\yes          &        &          &                  &\yes              &                                                                                      &                                                                                     & future tactile signals    &                                        \\
    \rowcolor{tablerowB} 3DTacDex~\cite{wu2025canonical} & DP                            &           &\yes       &\yes         &                  &\yes              &                                                                                      &                                                                                     &                          &                                        \\
    \rowcolor{tablerowA} TouchGuide~\cite{zhang2026touchguide} & DP                            &           &        &\yes         &                  &\yes              &                                                                                      &                                                                                     &                          &     \\
    \rowcolor{tablerowB} Comp-ACT~\cite{kamijo2024learning} & ACT                           &           &\yes       &          &                  &\yes              &\yes                                                                                     &                                                                                     &                          &                                        \\
    \rowcolor{tablerowA} Bi-ACT~\cite{buamanee2024bi} & ACT                           &           &        &\yes         &                  &\yes              &                                                                                      &\yes                                                                                    &                          &                                        \\
    \rowcolor{tablerowB} ViTacFormer~\cite{heng2025vitacformer} & ACT                           &           &\yes       &\yes         &                  &\yes              &                                                                                      &                                                                                     &                          & Tactile                                 \\
    \rowcolor{tablerowA} FILIC~\cite{ge2025filic} & ACT                           &           &\yes       &          &                  &\yes              &                                                                                      &                                                                                     &                          & Force                                   \\
    \rowcolor{tablerowB} ViTaL~\cite{zhao2025touch} & ACT                           &           &\yes       &          &                  &\yes              &                                                                                      &                                                                                     &                          &                                        \\
    \rowcolor{tablerowA} TACT~\cite{murooka2025tact} & ACT                           &           &        &\yes         &                  &\yes              &                                                                                      &                                                                                     &                          &                                        \\
    \rowcolor{tablerowB} FACTR~\cite{liu2025factr} & ACT                           &           &        &\yes         &                  &\yes              &                                                                                      &                                                                                     &                          &                                        \\
    \rowcolor{tablerowA} EquiContact~\cite{seo2025equicontact} & ACT                           &           &\yes       &          &\yes                 &               &\yes                                                                                     &                                                                                     &                          &                                        \\
    \rowcolor{tablerowB} ReTac-ACT~\cite{ruan2026retac} & ACT                           &           &        &\yes         &                  &\yes              &                                                                                      &                                                                                     &                          &                                        \\
    \rowcolor{tablerowA} MFMIL~\cite{ablett2024multimodal} & Regression                    &\yes          &\yes       &          &\yes                 &               &                                                                                      &                                                                                     & mode switching signal     &                                        \\
    \rowcolor{tablerowB} ForceSight~\cite{collins2024forcesight} & Regression                    &           &\yes       &          &\yes                 &               &                                                                                      &\yes                                                                                    & affordance map, depth map &                                        \\

    \rowcolor{tablerowA} FCLM~\cite{portela2024learning} & -                             &           &        &          &                  &               &                                                                                      &                                                                                     &                          & Force\textsuperscript{*}, State\textsuperscript{*}         \\
    \bottomrule
  \end{tabular}%
  }
  \vspace{2pt}
  \noindent\parbox{\textwidth}{\raggedright\footnotesize\textsuperscript{*}\,The corresponding modality observation occurs in Phase 2 of the architecture; it is repeated here for completeness.}
  \vspace{-5pt}
  \noindent\parbox{\textwidth}{\raggedright\footnotesize\textsuperscript{1}\,Motion Planning.}

\end{table}

\subsubsection{Actions}

Action outputs constitute the most fundamental category of intermediate modalities, as they define the robot's motion behavior in both spatial and temporal dimensions. Depending on task requirements and hardware embodiment, the action space can be formulated in several ways.

End-effector (EEF)-based actions specify the Cartesian pose or displacement of the manipulator in task space. This formulation is suitable when human interpretability is important. Joint-space actions directly specify joint positions, velocities, or torque commands. They are often used in high-degree-of-freedom dexterous systems, such as robotic hands. Direct prediction in joint space can also avoid potential issues introduced by inverse kinematics solvers, such as singularities in redundant manipulators.
Delta actions represent relative changes between the current and target states, such as $\Delta X$, $\Delta Y$, and $\Delta Z$, rather than absolute target states. They are commonly adopted in cross-embodiment learning settings because of their stronger generalization ability and reduced dependence on absolute initial poses. Delta actions are not a parallel category to EEF-space or joint-space actions. Instead, they indicate whether the action is represented in a relative or absolute form.

In addition to spatial representation, action outputs also differ in temporal horizon. Per-step policies generate only the next control action at each control cycle, as commonly seen in early regression-based imitation learning and reinforcement learning architectures. This formulation is suitable for reactive control, as it allows the policy to update its decision at every control step using the latest observations. In contrast, action-chunk-based policies predict a sequence of future actions in a single forward pass, as widely adopted in Transformer- and diffusion-based architectures such as ACT, Diffusion Policy, and VLA. By modeling temporal correlations across multiple future steps, action chunks can produce smoother and more coherent trajectories~\citep{zhang2026action}. They also provide flexibility during execution: later actions in the predicted chunk can be truncated and replaced by newly predicted actions from subsequent forward passes, allowing the policy to combine short-horizon reactivity with longer-horizon temporal consistency~\citep{liu2025bidirectional,black2025realtime}.

\subsubsection{Reference Force}

Beyond motion generation, recent tactile- and force-aware policies sometimes explicitly predict reference force signals to enable safe and stable physical interaction~\citep{zhi2025learningunifiedpolicyposition, zhou2025admittance, luo2025precise, buamanee2024bi, liu2025forcemimic, wu2025tacdiffusion, helmut2025tactile, huang2025tactile, collins2024forcesight, li2026forcevla2, zhang2025kinedex}. These reference signals, ranging from Cartesian wrenches to grasping forces, typically serve as key inputs to robot-end force control, where they help produce compliant and force-reactive behavior. In practice, they can be used through different mechanisms, such as feed-forward compensation to offset interaction loads and reduce response latency, task-driven regulation for dynamically modulating grasp intensity, and internal force guidance for balancing motion accuracy with physical safety. Predicting reference force is most meaningful when paired with explicit low-level force control. Such combinations are commonly used in controllers such as admittance control, impedance control, and hybrid force-position control, as discussed in Section~\ref{S5}.

\subsubsection{Control Stiffness}

Another critical output of the primary policy that naturally works with low-level control is the control stiffness $K$~\citep{hou2025adaptive, aburub2026learning, kamijo2024learning, seo2025equicontact}. The predicted stiffness gains, such as translational and rotational stiffness parameters ($K_p, K_r$), are typically integrated into impedance or admittance control frameworks to shape robot behavior as a virtual spring-damper system. This modulation allows the robot to balance safety and task performance by permitting intentional trajectory deviations when large contact forces arise in precision tasks. The magnitude of the gain $K$ typically determines the compliant behavior of the robot: a larger gain leads to a more rigid response under contact, while a smaller gain makes the robot more compliant.

\subsubsection{Other Auxiliary Modalities}

In addition to the core outputs above, auxiliary modalities can also be generated and integrated for system-level coordination and reasoning~\citep{noseworthy2025forge, xue2025reactive, fang2026force, ablett2024multimodal, collins2024forcesight, li2026forcevla2, zheng2026omnivta}. Examples include skill- or mode-switching triggers that activate specialized expert modules for highly dexterous subtasks~\citep{ablett2024multimodal}, task-progress indicators that estimate completion status and support automatic state transitions~\citep{li2026forcevla2}, affordance maps that highlight target object regions~\citep{collins2024forcesight}, and future latent predictions following the world-model convention~\citep{zheng2026omnivta}. Through these multidimensional outputs, the primary policy connects semantic perception, physical reasoning, and low-level execution, enabling robots to achieve robust interaction in diverse and complex scenarios.

\subsection{Sensory Prediction} \label{sec:obs_predict}

The primary policy in many recent tactile- and force-aware robotic frameworks also performs sensory prediction, where the model estimates or forecasts sensory observations from one or multiple other modalities. The primary motivation behind this mechanism is to enhance the policy's understanding of complex physical interactions through anticipatory perception, while also compensating for missing, delayed, or unavailable sensory inputs during inference.

Existing works typically perform sensory prediction along three major dimensions: force prediction, state prediction, and tactile prediction. Force prediction aims to estimate interaction forces during manipulation, particularly in scenarios where direct force sensing is unavailable or insufficient. Existing approaches generally include predicting force-related observations through cross-modal learning strategies~\citep{portela2024learning, zhang2025ta} and treating force prediction as an auxiliary self-supervised objective during policy learning~\citep{zhi2025learningunifiedpolicyposition}. State prediction focuses on estimating system states, hidden physical properties, or task-progression signals that are difficult to measure directly in the physical setup. Existing approaches generally include estimating privileged physical information from accessible observations~\citep{shirai2025sim, tang2026towards} and predicting internal robot states, including velocity or orientation, for improved motion stability~\citep{portela2024learning}. Tactile prediction aims to provide the robot with future tactile states from historical sensory sequences through temporal prediction models~\citep{heng2025vitacformer}.

Instead of jointly training the observation-prediction module with the action-generation model, another option, especially for force prediction, is to construct a separate observation estimator before policy-model training. This can include analytical estimation of external contact forces based on robot dynamics~\citep{ge2025filic}, following ideas from the contact-estimation literature~\citep{khalil2004modeling, hu2017contact}. It can also include learning-based approaches that train dedicated temporal models, such as LSTMs, for torque estimation~\citep{oh2026factr}, as studied in learning-based dynamics modeling~\citep{10669209, 10609990, yilmaz2020neural}.
NeuralActuator~\citep{dou2026neuralactuator} provides a representative example of a separately trained observation estimator for low-cost, servo-driven manipulators. It maps histories of commanded poses, proprioceptive states, tracking errors, and actuator-side telemetry to a torque surrogate using a temporal Transformer.

\subsection{Challenges of Single-Phase Policies}

A key challenge for the primary policy is to regulate the relative contributions of diverse modalities. Visual and language observations often provide high-level semantic information. When policies target semantic reasoning and long-horizon decision making, vision-language inputs tend to play a dominant role, and a relatively low inference frequency is therefore acceptable. This design is commonly observed in architectures that jointly perform semantic understanding and action generation, such as VLA models.

However, excessive dependence on high-dimensional visual or language encoding can introduce latency and reduce responsiveness, which is particularly problematic when stable contact must be maintained. Accordingly, policies designed for contact-intensive manipulation often benefit from reduced reliance on vision or language and stronger grounding in real-time interaction modalities, such as force feedback. These modalities are typically lower-dimensional and more directly related to contact states, making them better suited for high-frequency motion adjustment and reactive control. This consideration is frequently reflected in more compact and task-specific architectures, such as DP and ACT, where the smaller model scale also makes it more feasible to customize training objectives and introduce intermediate modality prediction, including reference force and control stiffness.

A further strategy is to separate modality usage across multiple models. In such designs, a higher-level model focuses on semantic understanding and low-frequency action generation, while a lower-level model emphasizes force-informed action generation or local action adjustment. This division naturally motivates the multi-phase approaches discussed in the next section, where different stages of the system specialize in distinct levels of perception and decision making.

\section{Phase 2: Modality-Discriminative Action Refinement} \label{S4}

As discussed in the previous section, the need for both semantic reasoning and fast inference poses challenges for a single-phase model. Using one model to incorporate all available modalities can lead to high inference latency in real-world applications. Thus, it is beneficial to distinguish among different modalities according to their characteristics and utilize them selectively. To this end, multi-phase frameworks address these challenges by decomposing the full action-generation pipeline into multiple stages, where each stage focuses on specific modalities or objectives~\cite{tie2025manual2skill, li2024cogact, zhang2024hirt, zhu2024language, posner2020robots, cai2025cookbench, wang2026mistypilot, ahn2022can, driess2023palm, ajay2023compositional}.

In such a multi-phase framework, the first stage, Phase 1, typically learns a primary policy that processes multimodal perception data and generates an initial action plan, as introduced in Section~\ref{S3}. The primary policy mainly focuses on understanding task instructions and generating a coarse action based on measurable sensory information. The second stage, Phase 2, learns a refinement policy or uses a model-based algorithm that takes the output from the primary policy and further refines it using additional or auxiliary modalities, such as force or tactile feedback. The action output of the second stage is then sent to the low-level controller for robot control.

\begin{figure}[!t]
  \centering
  \includegraphics[width=0.90\textwidth,keepaspectratio,trim=2cm 1.1cm 2cm 1.7cm,clip]{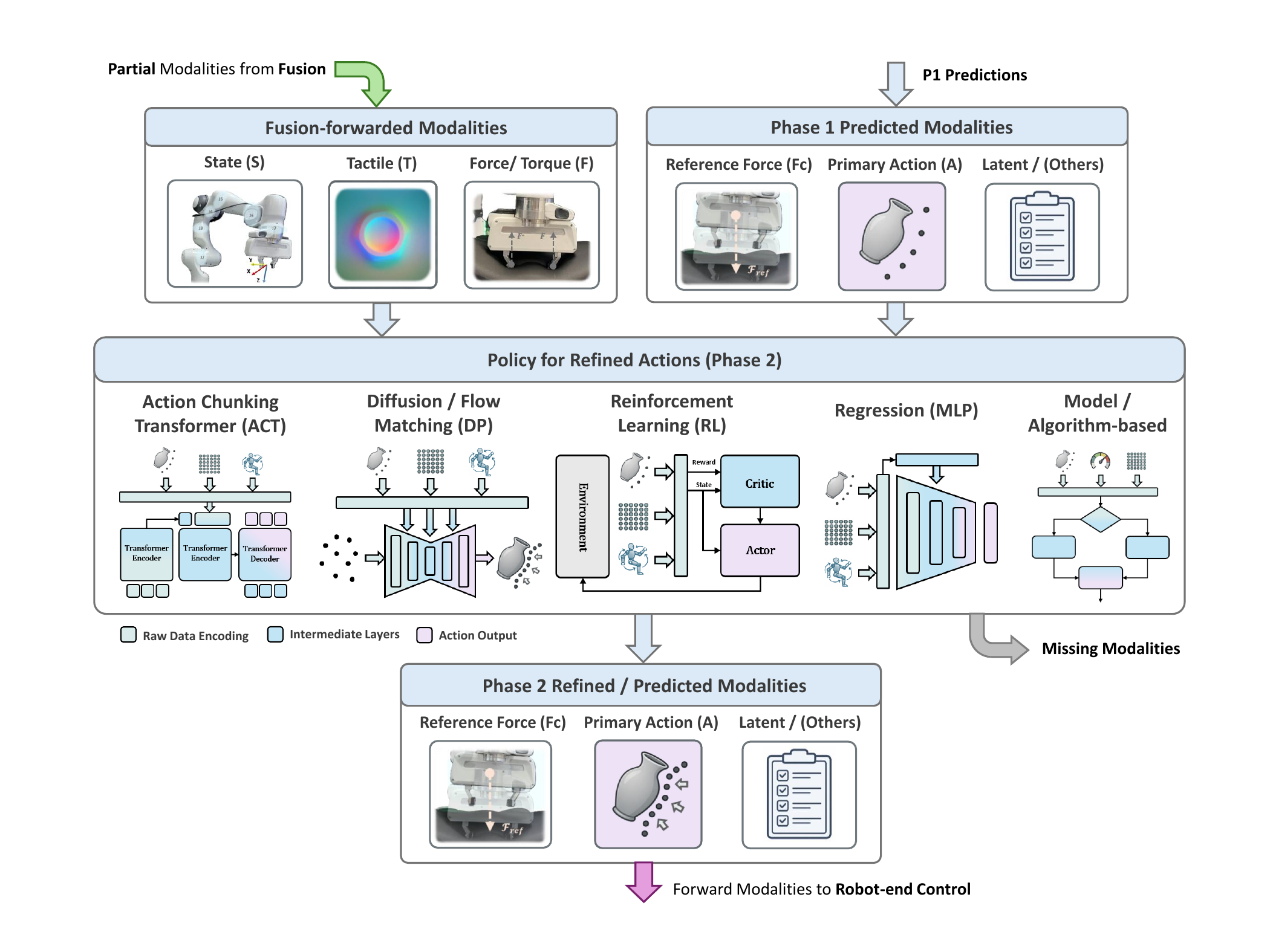}
  \vspace{-5pt}
  \caption{Overview of the Phase-2 action-refinement stage, where forwarded modality fusion and Phase-1 predictions are used for action refinement before low-level control.}
  \Description{}
  \label{Fig:phase2}
  \vspace{-5pt}
\end{figure}

A snapshot of the architecture discussed in this section is shown in Figure~\ref{Fig:phase2}, including the input-output data flow, candidate model structures, and an illustration of the action-refinement concept. The surveyed papers that follow our hierarchical-architecture criteria are further summarized in Table~\ref{tab_Part_three}. A more detailed discussion of the purpose of action refinement is provided in Section~\ref{Sec:4.4}.

\subsection{Discriminative Modality Injection}

The hierarchical framework enables flexible modality allocation across phases. This section discusses this property by first recapping the major input modalities used in Phase 1 for comparison, followed by the two data streams injected into Phase 2: forwarded input fusion and intermediate predictions from Phase 1.

When supplemented by a Phase 2 policy, the Phase 1 policy typically generates coarse actions primarily driven by vision, language, and state information. These inputs are sometimes augmented with auxiliary modalities, such as tactile sensing for fine-grained contact interpretation, as in ViTaL~\cite{zhao2025touch}, or force/torque sensing for direct interaction feedback, as in FoAR~\cite{he2025foar} and RDP~\cite{xue2025reactive}. Notable exceptions arise from task-specific demands. For high-precision assembly, TacDiffusion~\cite{wu2025tacdiffusion} completely omits vision in favor of purely force-proprioceptive feedback. Alternatively, motion-control-centric methods such as FCLM~\cite{portela2024learning} bypass a learned Phase 1 entirely and derive upstream commands directly from human operators.

At Phase 2, the hierarchical framework allows some raw sensory modalities to bypass Phase 1 and feed directly into the refinement stage for real-time action adjustment. This survey refers to this mechanism as \textit{discriminative modality injection}. Four modality types (tactile, force/torque, proprioception, and localized vision) and fusion-phase-inferred features are forwarded in existing works. These inputs appear at different frequencies, as shown in Table~\ref{tab_Part_three}. Proprioception is the most widely forwarded modality, since Phase 2 requires the current joint or end-effector state to represent the robot state. The other physical modalities are forwarded at different prevalence across methods: force/torque for high-frequency contact feedback, as in FoAR~\cite{he2025foar} and ForceSight~\cite{collins2024forcesight}; tactile sensing for contact interpretation, as in TACT~\cite{murooka2025tact} and OmniVTA~\cite{zheng2026omnivta}; and vision in the form of wrist-camera observations or local views, as in ViTaL~\cite{zhao2025touch} and VLA-Touch~\cite{bi2025vla}. Fusion-phase-inferred modalities include the contact probability predicted in FoAR~\cite{he2025foar} and the visuo-tactile latents predicted by the world model in OmniVTA~\cite{zheng2026omnivta}.

Intermediate predictions produced by Phase 1 can be grouped into three broad categories: action chunks, force commands, and other learned intermediates. Passing an action chunk is the most common choice, with most methods taking the Phase-1 generated action sequence as the refinement target. Force commands are key inputs for methods such as TacDiffusion~\cite{wu2025tacdiffusion}, which refines feed-forward force; KineDex~\cite{zhang2025kinedex}, which converts force commands into virtual displacement; and ForceSight~\cite{collins2024forcesight}, which uses a proportional control law with geometric goals. Other intermediate predictions are relatively diverse: RDP~\cite{xue2025reactive} produces a latent chunk for decoding; Force-Policy~\cite{fang2026force} forwards global visual features for fusion with local ones; and ForceSight~\cite{collins2024forcesight} constructs an affordance map and depth map as inputs to the low-level controller. FCLM~\cite{portela2024learning} again presents a special case, where Phase 2 receives teleoperation commands from a human operator rather than learned policy outputs, although these commands still contain both force and action information.

\subsection{Policy Architecture}

Similar to the Phase 1 policy, the model architectures used in Phase 2 can be broadly divided into five categories: RL, ACT, DP, regression, and rule/model-based methods that are specific to the refinement stage. While learnable strategies, including RL, ACT, DP, and regression, commonly use the output from Phase 1, their optimization targets differ according to their architectural characteristics. In the following, we briefly discuss how these architectures are adapted for action refinement, rather than focusing on their fundamental concepts and model structures.

The RL architecture trains a policy to compute residual actions over a Phase 1 reference trajectory. For example, UPPFC~\cite{zhi2025learningunifiedpolicyposition} and FCLM~\cite{portela2024learning} apply PPO to train whole-body force-position controllers for force residuals, whereas ViTaL~\cite{zhao2025touch} adds a capped residual from a DDPG policy to the base action.

The ACT architecture appears in Phase 2 as a lightweight variant adapted for high-frequency processing. Instead of using a full transformer, RDP~\cite{xue2025reactive} employs an asymmetric tokenizer based on CNN/GRU modules to decode latent chunks at 20--30 Hz.

The DP architecture uses stochastic diffusion to refine action plans rather than generating them from scratch. Force-Policy~\cite{fang2026force} denoises local motions by utilizing features inherited from the global policy, while VLA-Touch~\cite{bi2025vla} diffuses a VLA output trajectory toward fine-grained expert distributions.

Focusing on latency, the regression architecture employs shallow networks supervised by MSE or L1 losses to predict high-frequency reflexive corrections. A representative example is the three-layer MLP in OmniVTA~\cite{zheng2026omnivta}, which explicitly targets single-step modifications.

Despite the broader robot learning context, rule/model-based architectures are still frequently used because of their irreplaceable physical priors in contact-rich manipulation. These methods implement explicit, data-free control logic: FoAR~\cite{he2025foar} cascades contact-switching state machines; TacDiffusion~\cite{wu2025tacdiffusion} employs explicit dynamic-system filters to shape force commands; KineDex~\cite{zhang2025kinedex} adapts to external force using a fixed-stiffness admittance-mimicking mechanism; and TACT~\cite{murooka2025tact} and ForceSight~\cite{collins2024forcesight} use admittance tuning and proportional control, respectively. Collectively, these methods illustrate how learning- and model-based approaches complement each other in action refinement, even in data-driven robot learning pipelines.

\subsection{Output of Refinement Policy}

The output of Phase 2 is forwarded to Phase 3 for execution and can be broadly categorized into actions, force commands, and other intermediate representations. In most methods, Phase 2 outputs a refined action that is passed to the low-level controller. TacDiffusion~\cite{wu2025tacdiffusion} is one of the few works whose Phase 2 produces a force command, specifically a refined feed-forward force. Force-Policy~\cite{fang2026force} presents a more unusual case, where Phase 2 simultaneously outputs a force command, an action, and other representations, such as an interaction frame and a selection mask.

Within the action-output channel, methods can be further distinguished by how the action is expressed. A common pattern is that many Phase 2 outputs are incremental rather than absolute trajectories generated from scratch. ViTaL~\cite{zhao2025touch} outputs a residual added to the base action from Phase 1; UPPFC~\cite{zhi2025learningunifiedpolicyposition} and FCLM~\cite{portela2024learning} output joint-level residuals that are composed with a default joint pose to form the execution command; and Force-Policy~\cite{fang2026force}, ForceSight~\cite{collins2024forcesight}, and OmniVTA~\cite{zheng2026omnivta} output incremental or relative actions by utilizing Phase 1 trajectory priors. This pattern is consistent with the role of Phase 2 as a refiner, which makes local corrections based on the draft from Phase 1 or a reference pose. By contrast, a smaller group of methods outputs absolute targets: TACT~\cite{murooka2025tact} and KineDex~\cite{zhang2025kinedex} produce joint-level position targets, while FoAR~\cite{he2025foar} produces an end-effector pose target.

\subsection{Objectives of Action Refinement}\label{Sec:4.4}

\begin{table}[!t]
  \setlength{\tabcolsep}{3.2pt}
  \renewcommand{\arraystretch}{1.18}
  \centering
  \footnotesize
  \caption{Summary of Phase 2 of multi-hierarchical architectures in force-aware robot learning methods.}
  \vspace{-5pt}
  \label{tab_Part_three}
  \resizebox{\textwidth}{!}{%
  \rowcolors{4}{tablerowA}{tablerowB}
  \begin{tabular}{>{\scriptsize}l| >{\scriptsize}c| >{\scriptsize}c >{\scriptsize}c >{\scriptsize}c >{\scriptsize}c >{\scriptsize}c| >{\scriptsize}c| >{\scriptsize}c >{\scriptsize}c >{\scriptsize}c >{\scriptsize}c >{\scriptsize}c |>{\scriptsize}c |>{\scriptsize}c}
    \toprule
    \multicolumn{1}{>{\columncolor{headmethod}}c|}{\scriptsize\textbf{Method}} &
    \multicolumn{1}{>{\columncolor{headmodal}}c|}{\scriptsize\textbf{Architecture}} &
    \multicolumn{5}{>{\columncolor{headmethod}}c|}{\makebox[0pt]{\scriptsize\textbf{Visibility of Modalities of Interest}\textsuperscript{1}}} &
    \multicolumn{1}{>{\columncolor{headmodal}}c|}{\scriptsize\textbf{Phase 1 Output}} &
    \multicolumn{6}{>{\columncolor{headmethod}}c|}{\makebox[0pt]{\scriptsize\textbf{Phase 2 Output Modality}}} &
    \multicolumn{1}{>{\columncolor{headmodal}}c}{\makebox[0pt]{\scriptsize\makecell{\textbf{Sens.} \textbf{Pred.}}}}\\
    \hline

    \multicolumn{1}{c|}{\cellcolor{tablesubhead}} &
    \multicolumn{1}{c|}{\cellcolor{tablesubhead}} &
    \cellcolor{tablesubhead}\scriptsize\textbf{Force} &
    \cellcolor{tablesubhead}\scriptsize\textbf{Tactile} &
    \cellcolor{tablesubhead}\scriptsize\textbf{Vision} &
    \cellcolor{tablesubhead}\scriptsize\textbf{State} &
    \cellcolor{tablesubhead}\scriptsize\textbf{Other} &
    \cellcolor{tablesubhead} &
    \multicolumn{5}{c|}{\cellcolor{tablesubhead}\scriptsize\textbf{Action Type}} &
    \cellcolor{tablesubhead}\scriptsize\textbf{Ref Force} &
    \cellcolor{tablesubhead}\\
    \hline

    \multicolumn{1}{c|}{\cellcolor{tablecolhead}} &
    \multicolumn{1}{c|}{\cellcolor{tablecolhead}} &
    \multicolumn{5}{c|}{\cellcolor{tablecolhead}} &
    \multicolumn{1}{c|}{\cellcolor{tablecolhead}} &
    \cellcolor{tablecolhead}\scriptsize\textbf{Ref.\textsuperscript{2}} &
    \cellcolor{tablecolhead}\scriptsize\textbf{Exe.\textsuperscript{3}} &
    \cellcolor{tablecolhead}\scriptsize\textbf{Delta} &
    \cellcolor{tablecolhead}\scriptsize\textbf{EEF} &
    \cellcolor{tablecolhead}\scriptsize\textbf{Joint} &
    \multicolumn{1}{c|}{\cellcolor{tablecolhead}} &
    \multicolumn{1}{c}{\cellcolor{tablecolhead}}\\
    \hline

UPPFC~\cite{zhi2025learningunifiedpolicyposition} & RL & P1 &   & P1 & P1,P2 &   & Force, Action &   & \ding{51} &   & \ding{51} &   &   & Force\\

FCLM~\cite{portela2024learning} & RL &   &   &   & P2,P3 &   & Force, Action & \ding{51}  &  &   &   & \ding{51} &   & Force, State\\
ViTaL~\cite{zhao2025touch} & RL &   & P1,P2 & P1,P2 &   &   & Action & \ding{51} &   &   & \ding{51} &   &   & \\
Force-Policy~\cite{fang2026force} & DP & P2,P3 &   & P1,P2 & P2 &   & Other & \ding{51} &   &   & \ding{51} &   & \ding{51} & \\
VLA-Touch~\cite{bi2025vla} & DP & P2 & P1 & P1,P2 & P1,P2,P3  &   & Action & \ding{51} &   &   & \ding{51} &   &   & \\
RDP~\cite{xue2025reactive} & ACT-style & P2 & P1,P2 & P1 & P1 &  & Other &   & \ding{51} & \ding{51} & \ding{51} &   &   & \\
TacDiffusion~\cite{wu2025tacdiffusion} & Model-Based & P1 &   &   & P1,P3 &   & Force & \ding{51} &   &   & \ding{51} &   & \ding{51} & \\
TACT~\cite{murooka2025tact} & Model-Based & P2 & P1 & P1 & P1,P2 &   & Action &   & \ding{51} &   & \ding{51}  &  &   & \\
ForceSight~\cite{collins2024forcesight} & Model-Based & P2 &   & P1 & P2 &  &  Force, Action, Other &  \ding{51} &  & \ding{51} &  \ding{51} &  &   & \\
KineDex~\cite{zhang2025kinedex} & Model-Based &    & P1 & P1 & P1,P2 &   &  Force, Action &   & \ding{51} &   &  & \ding{51}  &   & \\
FoAR~\cite{he2025foar} & Algo-Based & P1,P2 &   & P1 & P2 & P2 & Action &   & \ding{51} &   & \ding{51} &   &   & \\   
OmniVTA~\cite{zheng2026omnivta} & Regression &   & P1,P2 & P1 & P1,P2 & P2 & Action &   & \ding{51} & \ding{51} & \ding{51} &   &   & \\

  \bottomrule
\end{tabular}%
}
  
  \vspace{2pt}
  \par\noindent\begingroup\footnotesize\raggedright
  \textbf{Note.} \textsuperscript{1}Similar information is reported in Table~\ref{tab:modality_fusion_summary}. This entry is repeated here to indicate whether and how discriminative modality injection is handled in each paper.
  \par\endgroup
  \par\noindent\begingroup\footnotesize\raggedright\textsuperscript{2} `Ref.' indicates reference action command for low-level controllers.\par\endgroup
  \par\noindent\begingroup\footnotesize\raggedright\textsuperscript{3} `Exe.' indicates robot-executable  actions.\par\endgroup
\end{table}

Serving as the middle layer between the high-level learning policy and the low-level controller, the Phase 2 policy focuses on utilizing auxiliary modalities to generate refined actions~\cite{bjorck2025gr00t, bu2024towards, chen2026fast, huang2024rekep}. Since existing VLA-based methods in Phase 1 rely heavily on visual observations and are often pre-trained on large-scale visual data, directly incorporating additional modalities into these models requires substantial new data and may introduce distribution shifts that degrade the performance of the primary policy. A multi-phase framework can therefore manage the complexity of multimodal perception by allowing the primary policy to focus on visual information, while the refinement policy integrates additional modalities without directly affecting the primary policy.

Meanwhile, the multi-phase framework also enables the robot to refine its actions at a higher control frequency, thereby improving control performance, as discussed in Section~\ref{S5}. Since compliant control methods usually require high control frequencies for stable force tracking, the Phase 1 agent may not be able to output actions at the required rate due to the computational cost of large model scales and multimodal data processing. The refinement policy can therefore benefit from its smaller model scale, taking the output from the primary policy and refining it asynchronously at a higher frequency.

Further, the refinement policy can process exclusive modality information and generate refined actions that are better suited to the current task phase and localized workspace, that is, post-process generated action chunks according to new modality conditions. This can substantially improve the success rate in certain complex tasks. Overall, the multi-phase framework provides a promising approach to addressing mismatches in frequency and information dependency between decision making and motion control, leading to a hybrid system that supports both high-level reasoning and low-level control~\cite{prescott2023understanding}.

\section{Phase 3: Robot-End Control} \label{S5}

A central low-level objective of force-aware robot learning is to achieve compliant manipulation, in which the robot adapts its motion based on force feedback or directly regulates the force applied to the environment. Compliant controllers therefore serve as the essential connection between learning-based action generation and reliable robot execution, providing a foundation for stable and compliant behavior. This capability is particularly important for tasks involving contact-rich manipulation and physical human-robot interaction (pHRI).

\begin{figure}[!t]
  \centering
  \includegraphics[width=0.95\textwidth,height=\textheight,keepaspectratio,trim=1.5cm 0.6cm 1.5cm 0.2cm,clip]{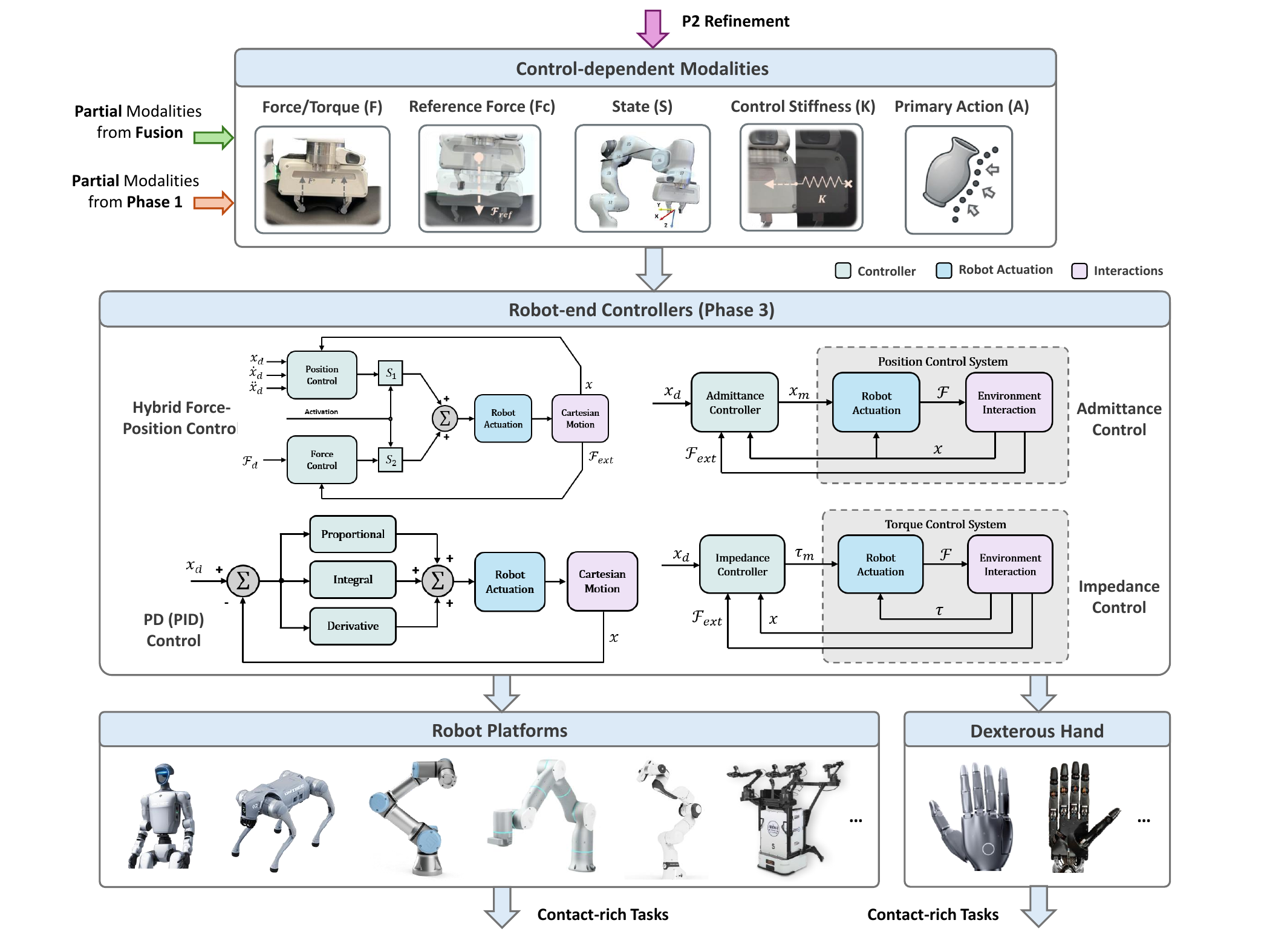}
  \vspace{-5pt}
  \caption{Robot-end control module showing the flow from learned policy outputs and force feedback to compliant low-level control.}
  \Description{}
  \label{Fig:control}
  \vspace{-5pt}
\end{figure}

In this section, we introduce compliant control methods to explain fundamentally how compliant behavior can be achieved, together with their respective advantages and disadvantages. According to the control objectives and input-output relationships, the major methods can be categorized into PD control, hybrid force-position control, admittance control, and impedance control. Figure~\ref{Fig:control} illustrates the control-loop block diagrams and multimodal data flow of the robot-end control module. Table~\ref{tab:control_summary} summarizes the control-related metrics of the surveyed methods, including controller type, control-demanding modality, robot platform, end-effector type, and available infrastructure-level control modes.

\subsection{Preliminaries on Control Theory}

\subsubsection{Robot Dynamics}
 
Compliant control methods are designed to allow robots to adapt their motion based on interaction forces. A prominent distinction between compliant control and position-based approaches is that the controller attempts to implement a dynamics relation between manipulator variables, e.g., position and velocity, and the force/torque applied to the environment, rather than directly controlling the position or velocity alone~\cite{Hogan1985}. In this process, the manipulator dynamics is one of the key equations in compliant control and can be expressed as~\cite{craig2005introduction}:
\begin{equation}
  \label{dynamic model}
  \mathbf{M}(\mathbf{q})\ddot{\mathbf{q}} + \mathbf{C}(\mathbf{q},\dot{\mathbf{q}})\dot{\mathbf{q}} + \mathbf{G}(\mathbf{q}) = \pmb{\tau} + \pmb{\tau}_{ext},
\end{equation}
where $\mathbf{M}(\mathbf{q})$ is the inertia matrix, $\mathbf{C}(\mathbf{q},\dot{\mathbf{q}})$ is the Coriolis and centrifugal matrix, $\mathbf{G}(\mathbf{q})$ is the gravity vector, $\pmb{\tau}$ is the torque control input, $\pmb{\tau}_{ext}$ is the external torque induced by contact, and $\ddot{\mathbf{q}}, \dot{\mathbf{q}}, \mathbf{q}$ are the acceleration, velocity, and position vectors, respectively. This equation, formally the inverse dynamics model, describes the joint torque input required by the robot given the robot states $[\ddot{\mathbf{q}}, \dot{\mathbf{q}}, \mathbf{q}]$ and a set of inherent robot-specific parameters specified by $[\mathbf{M}(\mathbf{q}), \mathbf{C}(\mathbf{q},\dot{\mathbf{q}}), \mathbf{G}(\mathbf{q})]$. Unlike robot kinematics, the dynamics model builds the relationship between the torque and the motion of the robot, allowing the robot to be controlled through torque to achieve the desired motion.

When making contact with the environment or interacting with humans, the robot will be affected by external forces, as described by the torque term $\pmb{\tau}_{ext}$. In this case, the robot is expected to be compliant and adapt its motion based on the external force~\cite{spong2020robot}. In the Cartesian space, the mass-spring-damper model is usually used to describe the compliant behavior of the robot, which can be expressed as:
\begin{equation}
  \label{msd-model}
  \mathbf{M}_d(\ddot{\mathbf{x}}_{\text{des}} -\ddot{\mathbf{x}}) + \mathbf{B}_d(\dot{\mathbf{x}}_{\text{des}} - \dot{\mathbf{x}}) + \mathbf{K}_d(\mathbf{x}_{\text{des}} - \mathbf{x}) = \mathbf{F}_{ext},
\end{equation}
where $\mathbf{M}_d, \mathbf{B}_d, \mathbf{K}_d$ are the inertia, damping, and stiffness matrices, respectively, $\mathbf{x}_{\text{des}}, \dot{\mathbf{x}}_{\text{des}}, \ddot{\mathbf{x}}_{\text{des}}$ are the desired position, velocity, and acceleration vectors of the end-effector, respectively, $\mathbf{x}, \dot{\mathbf{x}}, \ddot{\mathbf{x}}$ are the actual position, velocity, and acceleration vectors of the end-effector, respectively, and $\mathbf{F}_{ext}$ is the external wrench applied to the end-effector. By enforcing this mass-spring-damper relationship, the robot can allow position deviations under external contacts while resisting rapid changes in motion. The corresponding controller will be discussed later in this section.

However, Eqation (\ref{msd-model}) is expressed in Cartesian space, whereas the manipulator dynamics model is expressed in joint space. To combine these two models, the Jacobian matrix $\mathbf{J}(\mathbf{q})$ is used to transform the wrench from Cartesian space to joint space, which can be expressed as:
\begin{equation}
  \label{jacobian}
  \pmb{\tau}_{ext} = \mathbf{J}^T(\mathbf{q})\mathbf{F}_{ext}.
\end{equation}
To this end, the manipulator can be controlled under joint torque command to respond to the external wrench in Cartesian space and adapt its motion accordingly.

\subsubsection{PD Control}
 
Position control is one of the aforementioned ``position-based'' approaches and serves as the most common and fundamental low-level control method. It is mainly based on a proportional-derivative (PD) controller to follow the desired trajectory, which can be expressed as:
\begin{equation}
  \label{pd control}
  \pmb{u} = K_p(\mathbf{q}_{\text{des}} - \mathbf{q}) + K_d(\dot{\mathbf{q}}_{\text{des}} - \dot{\mathbf{q}}),
\end{equation}
where $K_p$ and $K_d$ are the proportional gain and derivative gain, respectively, $\mathbf{q}_{\text{des}}$ and $\dot{\mathbf{q}}_{\text{des}}$ are the desired joint position and velocity, respectively, and $\mathbf{q}$ and $\dot{\mathbf{q}}$ are the actual joint position and velocity, respectively. Here, $\pmb{u}$ denotes the generic control output, which can correspond to a torque command in practice. By adjusting the gains of the PD controller, the robot can follow the desired trajectory through the control output.

However, this method mainly focuses on tracking the desired trajectory and does not provide easily adjustable compliance. Although some existing methods can achieve compliant behavior under position-control mode through force estimation and simulated contact dynamics, traditional compliant control methods are still the most direct and effective way to achieve compliant manipulation.

\subsubsection{Impedance Control}
 
Impedance control enforces a desired dynamic relationship between the robot's motion and the forces it experiences, allowing the robot to exhibit compliant behavior by adjusting its impedance parameters. The inputs to an impedance controller are the desired position, velocity, and acceleration of the end-effector, while the output is the joint torque, which can be expressed as~\cite{Hogan1985,laghi2020unifying}:
\begin{gather}
  \label{impedance control}
  \pmb{\tau} = \mathbf{M}(\mathbf{q})\ddot{\mathbf{q}} + \mathbf{C}(\mathbf{q},\dot{\mathbf{q}})\dot{\mathbf{q}} + \mathbf{G}(\mathbf{q}) - \mathbf{J}^T(\mathbf{q})(\mathbf{M}_d\ddot{\mathbf{e}} + \mathbf{B}_d\dot{\mathbf{e}} + \mathbf{K}_d\mathbf{e}), \\
  \label{tracking error}
  \mathbf{e} = \mathbf{x}_{\text{des}} - \mathbf{x}, \quad
  \dot{\mathbf{e}} = \dot{\mathbf{x}}_{\text{des}} - \dot{\mathbf{x}}, \quad
  \ddot{\mathbf{e}} = \ddot{\mathbf{x}}_{\text{des}} - \ddot{\mathbf{x}}.
\end{gather}
In other words, this formulation specifies a desired trade-off between motion tracking and force tolerance through the impedance parameters $[\mathbf{M}_d, \mathbf{B}_d, \mathbf{K}_d]$. A lower stiffness allows larger positional deviations under external forces, whereas a higher stiffness enforces tighter trajectory tracking.

As illustrated in the impedance-control subfigure of Figure~\ref{Fig:control}, the inner control loop is torque control, while the outer control loop is position control. One advantage of impedance control is that the robot can achieve whole-body compliance without additional force sensors~\cite{hamid2021state}. However, impedance control may be less suitable for tasks that require precise position tracking, because the robot can be affected by external forces or uncompensated dynamics and modeling errors, including joint friction, payload variation, and end-effector weight~\cite{10994502}, causing deviations from the desired position. This limitation can be partly mitigated by adjusting the impedance parameters~\cite{zhang2024robot, chen2024physical, fu2024optimization, ficuciello2015variable, 10517447} or by combining impedance control with other control methods~\cite{ott2010unified, haddadin2024unified}.

\subsubsection{Admittance Control}
 
Admittance control, on the other hand, actively describes the relationship between the forces applied to the robot and its resulting motion~\cite{keemink2018admittance, chen2022human, chen2024compliance}. The input to an admittance controller is the force applied to the end-effector, while the output is the desired motion command, which can be expressed as~\cite{350927}:
\begin{equation}
  \label{admittance control}
  \ddot{\mathbf{x}}_{\text{des}} = \ddot{\mathbf{x}} + \mathbf{M}_d^{-1}(\mathbf{F}_{ext} - \mathbf{B}_d\dot{e} - \mathbf{K}_d e).
\end{equation}
The desired position can then be obtained by integrating the desired acceleration. When the system moves slowly, the acceleration term can be neglected, and the desired position can be obtained by integrating the desired velocity. Intuitively, admittance control follows the principle that a larger measured external force should induce a larger motion response along the force direction, allowing the robot to yield to the applied force.

Unlike impedance control, the inner control loop of admittance control is position control, while the outer control loop is force control. Admittance control is more suitable for tasks that require precise position tracking, as the controller can actively determine the allowable compromise in position accuracy based on force feedback~\cite{kang2019variable, ferraguti2019variable, li2018stable}. However, admittance control may require additional force sensors to accurately measure external forces, and its compliance is limited to the end-effector if no joint-level force/torque information is available.

\subsubsection{Hybrid Force/Position Control}
 
Hybrid force/position control combines position and force control to achieve compliant manipulation in selected directions, allowing the robot to maintain a desired position along certain action dimensions while regulating the force applied to the environment along others~\cite{raibert1981hybrid, yoshikawa2002dynamic, yoshikawa2003dynamic}. The position-control space should be orthogonal to the force-control space, and this decomposition can be defined by a selection matrix $\mathbf{S}$, where $\mathbf{S}$ is a diagonal matrix with binary values indicating the force-control directions~\cite{Khatib1987, fisher1992hybrid, namvar2005adaptive}. The inputs to a hybrid force/position controller are the desired position and the desired force, while the output can be expressed as:
\begin{equation}
\label{hybrid control}
\pmb{u} = \mathcal{F} (\mathbf{S}_2\cdot e_F) + \mathcal{P} (\mathbf{S}_1\cdot e),
\end{equation}
where $\mathbf{S}_1 = \mathbf{I}-\mathbf{S}_2$ is the complementary position-control selection matrix, $\mathcal{F}$ and $\mathcal{P}$ are the force-control and position-control functions, respectively, $e_F$ is the force error between the desired and actual forces, and $e$ is the position error.

This control method enables the robot to achieve both low-stiffness compliance and high-stiffness position control along different directions. It is particularly useful for tasks with clear force-control objectives, such as cutting fruit or wiping a board, where the force-control direction is easy to specify. However, in unstructured or less informative environments, it can be challenging to intuitively disentangle the force-control directions from the position-control directions while maintaining their orthogonality.

\subsection{Robot-End Control Implementations}

\begin{table}[!t]
  \setlength{\tabcolsep}{4.6pt}
  \renewcommand{\arraystretch}{1.30}
  \centering
  \footnotesize
  \caption{Summary of robot-end control configurations in the reviewed tactile- and force-aware robot learning methods.}
  \vspace{-5pt}
  \label{tab:control_summary}
  \makebox[\textwidth][c]{%
  \resizebox{0.95\textwidth}{!}{%
  \begin{tabular}{l|c|ccc|c|c|c}
    \toprule
    \multicolumn{1}{>{\columncolor{headmethod}}c|}{\textbf{Method}} &
    \multicolumn{1}{>{\columncolor{headmodal}}c|}{\textbf{Controller}} &
    \multicolumn{3}{>{\columncolor{headmethod}}c|}{\textbf{Control-Demanding Modality}} &
    \multicolumn{1}{>{\columncolor{headmodal}}c|}{\textbf{Robot}} &
    \multicolumn{1}{>{\columncolor{headmethod}}c|}{\textbf{EEF type}} &
    \multicolumn{1}{>{\columncolor{headmodal}}c}{\textbf{Control Interface\textsuperscript{1}}}\\
    \multicolumn{1}{c|}{\cellcolor{tablesubhead}} &
    \multicolumn{1}{c|}{\cellcolor{tablesubhead}} &
    \cellcolor{tablesubhead}\textbf{Fusion} &
    \cellcolor{tablesubhead}\textbf{Phase1} &
    \cellcolor{tablesubhead}\textbf{Phase2} &
    \multicolumn{1}{c|}{\cellcolor{tablesubhead}} &
    \multicolumn{1}{c|}{\cellcolor{tablesubhead}} &
    \multicolumn{1}{c}{\cellcolor{tablesubhead}}\\
    \hline
    \rowcolor{tablerowA} UPPFC~\cite{zhi2025learningunifiedpolicyposition} & Position &  &  & Action & Unitree Z1 & Gripper & Position\\
    \rowcolor{tablerowB} FCLM~\cite{portela2024learning} & Position & State &  & Action & Unitree Z1 & Gripper & Position\\
    \rowcolor{tablerowB} TA-VLA~\cite{zhang2025ta} &  &  & Action & & ALOHA \& ROKAE SR & Gripper & Position \& Impedance\\
    \rowcolor{tablerowA} M3L~\cite{sferrazza2024power} &  &  & Action & & Franka & Gripper \& Hand & Impedance\\
    \rowcolor{tablerowB} ACP~\cite{hou2025adaptive} & Admittance & Force & Action, $K$ & & UR5e & Customized & Position\\
    \rowcolor{tablerowA} FORGE~\cite{noseworthy2025forge} & Impedance & State & Action & & Franka & Gripper & Impedance\\
    \rowcolor{tablerowA} AdmitDiff~\cite{zhou2025admittance} & Admittance & Force & Action, $F_{ref}$ & & UR5-CB3 & Hand & Position\\
    \rowcolor{tablerowB} HIL-SERL~\cite{luo2025precise} & Impedance & State & Action, $F_{ref}$ & & Franka & Gripper & Impedance\\
    \rowcolor{tablerowA} DexForce~\cite{chen2025dexforce} & Impedance & State & Action & & -- --\textsuperscript{3} & Hand & -- --\\
    \rowcolor{tablerowB} FoAR~\cite{he2025foar} & Position &  &  & Action& Flexiv Rizon & Gripper & Hybrid F/P\\
    \rowcolor{tablerowA} DIPCOM~\cite{aburub2026learning} & FDCC &  & Action, $K$ & & UR5e & Gripper & Position\\
    \rowcolor{tablerowB} Comp-ACT~\cite{kamijo2024learning} & FDCC &  & Action, $K$ & & UR5e & Gripper & Position\\
    \rowcolor{tablerowA} HATO~\cite{lin2025learning} & Position &  & Action & & UR5e & Hand & Position\\
    \rowcolor{tablerowB} RDP~\cite{xue2025reactive} &  &  &  & Action & Flexiv Rizon & Gripper & Hybrid F/P\\
    \rowcolor{tablerowA} TacDiffusion~\cite{wu2025tacdiffusion} & Impedance & State &  & $F_{ref}$ & Franka & Gripper & Impedance\\
    \rowcolor{tablerowB} FARM~\cite{helmut2025tactile} & Impedance & State & Action, $F_{ref}$ & & Franka & Gripper & Impedance\\
    \rowcolor{tablerowA} ViTacFormer~\cite{heng2025vitacformer} &  &  & Action & &  Realman & Hand & Position\\
    \rowcolor{tablerowB} 3D-ViTac~\cite{huang20243d} &  &  & Action & & -- -- & Gripper & -- --\\
    \rowcolor{tablerowA} UniT~\cite{xu2025unit} & Position &  & Action & & ALOHA & Gripper & Position\\
    \rowcolor{tablerowB} FuSe~\cite{jones2025beyond} & Position &  & Action & & WidowX 250 & Gripper & Position\\
    \rowcolor{tablerowA} ForceVLA~\cite{yu2025forcevla} &  &  & Action & &  Flexiv Rizon & Gripper & Hybrid F/P\\
    \rowcolor{tablerowB} Tactile-VLA~\cite{huang2025tactile} & Hybrid F/P & Force & Action, $F_{ref}$ & & Unspecified \textsuperscript{4} & Gripper & Unspecified\\
    \rowcolor{tablerowA} TLA~\cite{hao2025tla} &  &  & Action & & Unspecified & Gripper & Unspecified\\
    \rowcolor{tablerowB} FILIC~\cite{ge2025filic} & Impedance & State & Action & & AIRBOT Play & Gripper & Position\\
    \rowcolor{tablerowA} ManipForce~\cite{lee2025manipforce} &  &  & Action & & Franka & Gripper & Impedance\\
    \rowcolor{tablerowB} OGRL-VT~\cite{shirai2025sim} & Impedance & State & Action & & MELFA & Customized & Position\\
    \rowcolor{tablerowA} ViTaL~\cite{zhao2025touch} &  &  &  & Action& Ufactory xArm 7 & Gripper & Position\\
    \rowcolor{tablerowB} TACT~\cite{murooka2025tact} & Position &  &  & Action& RHP7 Kaleido & Gripper & Position\\
    \rowcolor{tablerowA} VT-HMD~\cite{ye2026visual} & Position &  & Action & & -- -- & Hand & -- --\\
    \rowcolor{tablerowB} TaF-VLA~\cite{huang2026tactile} &  &  & Action & & Franka & Gripper & Impedance\\
    \rowcolor{tablerowA} Force-Policy~\cite{fang2026force} & Hybrid F/P & Force & Action & Action, $F_{ref}$ & Flexiv Rizon & Gripper & Hybrid F/P\\
    \rowcolor{tablerowB} FACTR~\cite{liu2025factr} &  &  & Action & & Franka & Gripper & Impedance\\
    \rowcolor{tablerowA} MFMIL~\cite{ablett2024multimodal} & Impedance & State & Action & & Franka & Gripper & Impedance\\
    \rowcolor{tablerowB} ForceSight~\cite{collins2024forcesight} &  &  &  & Action& Stretch RE1 & Gripper & Position\\
    \rowcolor{tablerowA} CRAFT~\cite{zhang2026craft} & Impedance & State & Action &  & Franka & Gripper & Impedance\\
    \rowcolor{tablerowB} EquiContact~\cite{seo2025equicontact} & Admittance & Force, State & Action, $K$ & & Unspecified & Gripper & Unspecified\\
    \rowcolor{tablerowA} MoDE-VLA~\cite{tang2026towards} & Position &  & Action & & SharpaNorth & Gripper & Position\\
    \rowcolor{tablerowB} ForceVLA2~\cite{li2026forcevla2} & Hybrid F/P & Force & Action, $F_{ref}$ & & Flexiv Rizon & Gripper & Hybrid F/P\\
    \rowcolor{tablerowA} ReTac-ACT~\cite{ruan2026retac} &  &  & Action & & Realman RM75-6FV & Gripper & Position\\
    \rowcolor{tablerowB} TacVLA~\cite{zhang2026tacvla} &  &  & Action & & Franka & Gripper & Impedance\\
    \rowcolor{tablerowA} KineDex~\cite{zhang2025kinedex} &  &  &  & Action& Franka & Hand & Impedance\\
    \rowcolor{tablerowB} OmniVTA~\cite{zheng2026omnivta} &  &  &  & Action& UFactory xArm7 & Gripper & Position\\
    \rowcolor{tablerowA} OmniVTLA~\cite{cheng2025omnivtla} &  &  & Action & & UR5 & Gripper & Position\\
    \rowcolor{tablerowB} ViTaS~\cite{tian2026vitas} &  &  & Action & & Galaxea-R1 & Gripper & Position\\
    \rowcolor{tablerowA} 3DTacDex~\cite{wu2025canonical} & Position &  & Action & & JAKA MiniCobo & Hand & Position\\
    \rowcolor{tablerowB} VTAO-BiManip~\cite{sun2025vtao} & Position &  & Action & & Unitree Z1  & Hand & Impedance\\
    \rowcolor{tablerowA} VTLA~\cite{zhang2025vtla} &  &  & Action & & UR3 & Gripper & Position\\
    \rowcolor{tablerowB} ARCH~\cite{sun2024arch} &  &  &  & Action & UR10e  & Gripper & Position\\
    \rowcolor{tablerowA} TouchGuide~\cite{zhang2026touchguide} &  &  & Action & & BiARX5 \& Flexiv Rizon4 & Gripper & Position \& Hybrid F/P\\
    \rowcolor{tablerowB} VLA-Touch~\cite{bi2025vla} & Position \& Impedance & State &  & Action& Franka & Gripper & Impedance\\
    \rowcolor{tablerowA} ForceMimic~\cite{liu2025forcemimic} & Position \& Hybrid F/P\textsuperscript{2} & Force & Action, $F_{ref}$ &  & Flexiv Rizon & Gripper & Hybrid F/P\\
    \rowcolor{tablerowB} Bi-ACT~\cite{buamanee2024bi} & Bilateral &  & Action, $F_{ref}$ & & OpenMANIPULATOR-X & Gripper & Position\\
    \rowcolor{tablerowA} DWF~\cite{kang2025robotic} &  &  & Action & & KUKA IIWA 14 \&  ABB IRB120  & Customized & Position \& Impedance\\
    \bottomrule
  \end{tabular}%
  }}
  \vspace{2pt}
  \par\noindent\begingroup\footnotesize\raggedright\textsuperscript{1} Control Interface denotes the primary infrastructure-level control strategy provided by each robot.\par\endgroup
  \par\noindent\begingroup\footnotesize\raggedright\textsuperscript{2} Hybrid F/P indicates the hybrid force/position control approach.\par\endgroup
  \par\noindent\begingroup\footnotesize\raggedright\textsuperscript{3} `-- --' indicates that the paper focused on hand manipulation and did not cover robot manipulation.\par\endgroup
  \par\noindent\begingroup\footnotesize\raggedright\textsuperscript{4} `Unspecified' indicates that a robot is used in the paper, but the model is not specified.\par\endgroup
  \vspace{-5pt}
\end{table}

The above introduction provides a basic understanding of compliant control methods and offers a reference for controller selection. Although these controllers serve similar force-compliance purposes, the choice of a specific method still depends on the task configuration, such as whether precise position tracking is required, whether additional force sensors are available, and whether the joint-torque command channel is accessible to users. Thus, the controller should be selected carefully according to the physical system settings. Table~\ref{tab:control_summary} summarizes the implementations of control methods, together with the control-demanding modalities provided by higher-level learning policies.

\textit{Control Mode:} The controllers include position control~\cite{zhi2025learningunifiedpolicyposition, portela2024learning, he2025foar, liu2025forcemimic, lin2025learning, xu2025unit, jones2025beyond, murooka2025tact, ye2026visual, tang2026towards, bi2025vla, wu2025canonical,sun2025vtao}, impedance control~\cite{noseworthy2025forge, luo2025precise, chen2025dexforce, wu2025tacdiffusion, helmut2025tactile, ge2025filic, shirai2025sim, ablett2024multimodal, zhang2026craft, bi2025vla}, admittance control~\cite{hou2025adaptive, zhou2025admittance, seo2025equicontact}, hybrid force/position control~\cite{liu2025forcemimic, huang2025tactile, fang2026force, li2026forcevla2}, and Forward Dynamics Compliance Control (FDCC)~\cite{aburub2026learning, kamijo2024learning}, which are used to control robots using commands from either teleoperation systems or learned policies~\cite{8206325}. Since some methods do not explicitly report the control method they use, we mark them as unknown to avoid ambiguity. In such cases, however, joint or end-effector position commands are commonly sent directly to the robot.

\textit{Control-demanding Modalities:} The control-demanding modality refers to the input required by the low-level controller and is categorized into three types according to where the modality is forwarded from: (1) Fusion: This indicates that raw measurements or fused features are used not only by the high-level learning policy but also directly forwarded to the low-level control loop as feedback signals; (2) Phase 1: This indicates that the controller input is forwarded from the primary policy output; and (3) Phase 2: This indicates that the controller input is directly obtained from the refinement policy output. The control-demanding modality reflects how the learning policy supports low-level robot control and how control behavior is coupled with the learning-model output, thereby providing a clearer understanding of modality importance. The corresponding data flow is also illustrated in Figure~\ref{Fig:big_framework_all}.

\textit{Robot Model Selection:} To further elaborate on the selection criteria for different platforms, we summarize the robot types and their supported compliant control modes. This information clarifies whether each method leverages the native control interface provided by the robot or employs a custom-designed controller to achieve compliant behavior. The most commonly used robot platform is Franka, which natively provides a stable impedance-control interface. The Flexiv Rizon offers a hybrid force/position-control interface, making it particularly suitable for tasks requiring precise force regulation. The UR5e is equipped with a force/torque (F/T) sensor, enabling admittance control through direct force feedback and facilitating controller design. Other robot platforms primarily provide position-control interfaces, where compliant behavior must be achieved through either custom controller design or action-space manipulation.

\textit{End-effector Selection:} The end-effector (EEF) type describes how each method interacts with the environment. The most common EEF is the gripper, which is often used to grasp different tools for various tasks. Although a gripper typically has only a single degree of freedom (DoF), it is easy to control and can maintain stable contact. In contrast, a dexterous hand has multiple DoFs and can grasp objects with irregular shapes, but it is also more difficult to control. Therefore, some works focus more on hand control than on the manipulator itself~\cite{chen2025dexforce, huang20243d, ye2026visual}. Apart from grippers and hands, some works use customized tools for specific tasks to achieve better interaction performance~\cite{hou2025adaptive, kang2025robotic, shirai2025sim}.

\subsection{Practical Considerations for Compliant Control}

Robot hardware largely determines the compliant control strategy employed in practice. Many force-aware robot learning methods still use position control as their low-level controller~\cite{zhi2025learningunifiedpolicyposition, portela2024learning, he2025foar, liu2025forcemimic, lin2025learning, xu2025unit, jones2025beyond, murooka2025tact, ye2026visual, tang2026towards, bi2025vla, wu2025canonical,sun2025vtao}, because position control is the most natural user interface and is generally supported by most robot platforms. Under position control, the robot remains rigid and can achieve high positional accuracy, but this rigidity may also lead to unintended contact forces. Such contact forces or collisions may be acceptable for manipulators with low-power motors, but can be dangerous for industrial-level robot manipulators.

Among compliant control methods, impedance control is the most widely used method in the surveyed papers~\cite{noseworthy2025forge, luo2025precise, chen2025dexforce, wu2025tacdiffusion, helmut2025tactile, ge2025filic, shirai2025sim, ablett2024multimodal, zhang2026craft, bi2025vla}. It can achieve whole-body compliance through feed-forward torque commands without active force sensing. Methods using impedance control are usually based on Franka, KUKA, or ROKAE robot platforms, which provide impedance control as their low-level controller. Meanwhile, methods using hybrid force/position control~\cite{liu2025forcemimic, huang2025tactile, fang2026force, li2026forcevla2} are usually based on the Flexiv Rizon, which provides a hybrid force/position control interface. These robot platforms provide convenient ways to implement compliant control, allowing learning algorithms to focus mainly on action output rather than learning complex compliant interactions from scratch.

Compliant control provides a solid foundation for force-aware manipulation. However, it also introduces another layer of complexity. If the control gains and parameters are not properly designed and selected, the robot may still experience large contact forces and undesired phenomena such as overshooting. Therefore, the low-level compliant controller should be carefully tested with properly tuned parameters, rather than treated merely as a callable interface.

With reliable robot-end force-control methods, the robot can generally behave more safely by avoiding excessive contact forces while maintaining stable contact for contact-demanding tasks such as peeling and wiping. The controller and the learning algorithm should complement each other to achieve better compliant interaction performance through appropriate modality exchange, interface connection, and frequency handling.

\section{Contact-Rich Manipulation Tasks} \label{S6}

\emph{Contact-rich manipulation} refers to tasks in which the robot perceives and adapts to interaction states, such as contact force, contact location, friction, and object deformation, rather than merely generating contact-free trajectories. Unlike traditional motion-centric manipulation, tactile- and force-aware manipulation requires finer-grained motion generation and explicit contact and collision handling to prevent damage to the robot and its environment. The 53 papers analyzed in this section constitute the TF-ART corpus mapped in Figure~\ref{Fig:big_framework_all}. Table~\ref{tab:tasks} summarizes the evaluation tasks across these papers, covering 21 manipulation tasks and the tactile, force, and torque sensors used. Figure~\ref{fig:task-histogram} presents a histogram showing the frequency with which each task appears in the corpus.

\subsection{Task Classification}

Based on their dominant interaction characteristics, we categorize the surveyed tasks into four groups: precision-contact manipulation, deformable and surface-contact manipulation, dynamic and multi-contact manipulation, and household and long-horizon manipulation. These categories are not strictly disjoint, since many real-world tasks combine geometric constraints, deformable contact, temporal dynamics, and unstructured environments.

These task categories also stress different phases of TF-ART. Precision-contact manipulation mainly emphasizes Phase 2 refinement and Phase 3 compliant control, since small misalignment requires force/tactile-guided correction during contact. Deformable and surface-contact manipulation emphasizes Phase 0 representation learning and Phase 3 force regulation, as material-dependent contact states must be encoded and controlled over extended interactions. Dynamic and multi-contact manipulation stresses the coupling between Phase 2 and Phase 3, where fast refinement and low-latency reactive control are required. Household and long-horizon manipulation places stronger demand on Phase 1 primary policies, since semantic reasoning, object affordance understanding, and multi-stage action generation are needed before local contact adaptation.

\vspace{-4pt}
\subsubsection{Precision Contact Manipulation}

This class includes fine-grained tasks, such as peg-in-hole insertion and cap or nut screwing, which require contact-misalignment detection and force-guided motion control. During insertion, the contact region is often occluded and positional errors may be too small to detect visually, so even slight deviations can cause jamming. Cap and handle rotation similarly require coordinated rotational torque and axial force to prevent slipping. These tasks share a common demand for contact localization, fine-grained alignment correction, and compliance control under constrained contact.

\begin{table}[!t]
  \setlength{\tabcolsep}{1.0pt}
  \renewcommand{\arraystretch}{1.04}
  \centering
  \scriptsize
  \renewcommand{\rot}[1]{\makebox[\linewidth][c]{\rotatebox[origin=lB]{90}{\scriptsize\strut #1\hspace{5pt}}}}
  \caption{Summary of tactile/force-aware manipulation tasks and sensing modalities across the surveyed methods.}
  \label{tab:tasks}
  \makebox[\textwidth][c]{%
  \begin{tabular}{@{}>{\fontsize{6.4}{7.1}\selectfont}l|*{5}{>{\centering\arraybackslash}p{0.52cm}}|*{5}{>{\centering\arraybackslash}p{0.52cm}}|*{6}{>{\centering\arraybackslash}p{0.52cm}}|*{5}{>{\centering\arraybackslash}p{0.52cm}}|*{3}{>{\centering\arraybackslash}p{0.58cm}}@{}}
  \toprule
  
  \multicolumn{1}{c|}{\cellcolor{headmethod}\tiny\textbf{\strut Method}}
  & \multicolumn{5}{c|}{\cellcolor{cA}\makebox[0pt][c]{\tiny\textbf{\strut A Precision Contact}}}
  & \multicolumn{5}{c|}{\cellcolor{cB}\makebox[0pt][c]{\tiny\textbf{\strut B Deformable \& Surface-contact}}}
  & \multicolumn{6}{c|}{\cellcolor{cC}\makebox[0pt][c]{\tiny\textbf{\strut C Dynamic \& Multi-contact}}}
  & \multicolumn{5}{c|}{\cellcolor{cD}\makebox[0pt][c]{\tiny\textbf{\strut D Household and Long-Horizon}}}
  & \multicolumn{3}{c}{\cellcolor{cS}\makebox[0pt][c]{\tiny\textbf{\strut Sensor}}} \\
  
  \cmidrule(lr){2-6}
  \cmidrule(lr){7-11}
  \cmidrule(lr){12-17}
  \cmidrule(lr){18-22}
  \cmidrule(lr){23-25}
  
  \multicolumn{1}{c|}{\cellcolor{tablecolhead}}
  & \rot{\textbf{A1~Insertion}}
  & \rot{\textbf{A2~Cap/Nut Screwing}}
  & \rot{\textbf{A3~Assembly Plug}}
  & \rot{\textbf{A4~Bi.\ Insertion}}
  & \rot{\textbf{A5~Rotate Handle/Box}}
  & \rot{\textbf{B1~Peeling}}
  & \rot{\textbf{B2~Wiping}}
  & \rot{\textbf{B3~Cutting}}
  & \rot{\textbf{B4~Slide Object}}
  & \rot{\textbf{B5~Soft/Fragile Grasp}}
  & \rot{\textbf{C1~Mobile Catch}}
  & \rot{\textbf{C2~Sliding}}
  & \rot{\textbf{C3~Reorientation}}
  & \rot{\textbf{C4~In-hand Manip.}}
  & \rot{\textbf{C5~Bi.\ Lifting}}
  & \rot{\textbf{C6~Bi.\ Wiping}}
  & \rot{\textbf{D1~Open/Close Door}}
  & \rot{\textbf{D2~Pick-and-place}}
  & \rot{\textbf{D3~Pouring}}
  & \rot{\textbf{D4~Weight Pulling}}
  & \rot{\textbf{D5~Push Object}}
  & \rot{\textbf{EEF Wrench}}
  & \rot{\textbf{Joint Torque}}
  & \rot{\textbf{Tactile Sensor}} \\
  
  \midrule
  
  \rowcolor{tablerowA}  UPPFC~\cite{zhi2025learningunifiedpolicyposition}        &\n &\n &\n &\n &\n &\n &\y &\n &\n &\n &\n &\n &\n &\n &\n &\n &\y &\n &\n &\n &\n &\n &\n &\n \\
  \rowcolor{tablerowB}    FCLM~\cite{portela2024learning}         &\n &\n &\n &\n &\n &\n &\n &\n &\n &\n &\n &\n &\n &\n &\n &\n &\y &\n &\y &\y &\n &\n &\n &\n \\
  \rowcolor{tablerowA}  FORGE~\cite{noseworthy2025forge}        &\y &\y &\n &\n &\n &\n &\n &\n &\n &\n &\n &\n &\n &\n &\n &\n &\n &\n &\n &\n &\n &\y &\n &\n \\
  \rowcolor{tablerowB}    TA-VLA~\cite{zhang2025ta}       &\n &\n &\y &\n &\y &\n &\n &\n &\n &\n &\n &\n &\n &\n &\n &\n &\y &\y &\y &\n &\y &\n &\y &\n \\
  \rowcolor{tablerowA}  M3L~\cite{sferrazza2024power}          &\y &\n &\n &\n &\n &\n &\n &\n &\n &\n &\n &\n &\n &\y &\n &\n &\y &\n &\n &\n &\n &\n &\n &\y \\
  \rowcolor{tablerowB}    ACP~\cite{hou2025adaptive}          &\n &\n &\n &\n &\n &\n &\y &\n &\n &\n &\n &\n &\n &\n &\n &\n &\n &\n &\n &\n &\n &\y &\n &\n \\
  \rowcolor{tablerowA}  AdmitDiff~\cite{zhou2025admittance}    &\y &\n &\n &\n &\y &\n &\y &\n &\n &\n &\n &\n &\n &\n &\n &\n &\y &\y &\n &\n &\n &\y &\n &\n \\
  \rowcolor{tablerowB}    HIL-SERL~\cite{luo2025precise}     &\n &\n &\y &\n &\n &\n &\n &\n &\n &\n &\n &\n &\n &\n &\n &\n &\n &\n &\n &\n &\n &\n &\n &\n \\
  \rowcolor{tablerowA}  Bi-ACT~\cite{buamanee2024bi}       &\n &\n &\n &\n &\n &\n &\n &\n &\n &\n &\n &\n &\n &\n &\n &\n &\y &\y &\n &\n &\n &\n &\y &\n \\
  \rowcolor{tablerowB}    DexForce~\cite{chen2025dexforce}     &\n &\y &\n &\n &\n &\n &\n &\n &\y &\n &\n &\n &\y &\n &\n &\n &\y &\y &\n &\n &\n &\y &\n &\n \\
  \rowcolor{tablerowA}  FoAR~\cite{he2025foar}         &\n &\n &\n &\n &\n &\y &\y &\n &\n &\n &\n &\n &\n &\n &\n &\n &\n &\n &\n &\n &\n &\y &\n &\n \\
  \rowcolor{tablerowB}    ForceMimic~\cite{liu2025forcemimic}   &\n &\n &\n &\n &\n &\y &\n &\n &\n &\n &\n &\n &\n &\n &\n &\n &\n &\n &\n &\n &\n &\y &\n &\n \\
  \rowcolor{tablerowA}  DIPCOM~\cite{aburub2026learning}       &\n &\n &\n &\y &\n &\n &\n &\n &\n &\n &\n &\n &\n &\n &\n &\n &\n &\n &\n &\n &\n &\y &\n &\n \\
  \rowcolor{tablerowB}    Comp-ACT~\cite{kamijo2024learning}     &\y &\n &\n &\n &\n &\n &\y &\n &\n &\n &\n &\n &\n &\n &\n &\y &\n &\n &\n &\n &\n &\y &\n &\n \\
  \rowcolor{tablerowA}  HATO~\cite{lin2025learning}         &\n &\n &\n &\n &\n &\n &\n &\n &\n &\n &\n &\n &\n &\n &\n &\n &\n &\n &\y &\n &\n &\n &\n &\y \\
  \rowcolor{tablerowB}    RDP~\cite{xue2025reactive}          &\n &\n &\n &\n &\n &\y &\y &\n &\n &\n &\n &\n &\n &\n &\y &\n &\n &\n &\n &\n &\n &\n &\y &\y \\
  \rowcolor{tablerowA}  DWF~\cite{kang2025robotic}          &\n &\n &\n &\n &\n &\n &\n &\n &\n &\n &\n &\n &\n &\n &\n &\n &\n &\n &\n &\n &\n &\n &\y &\n \\
  \rowcolor{tablerowB}    TacDiffusion~\cite{wu2025tacdiffusion} &\y &\n &\n &\n &\n &\n &\n &\n &\n &\n &\n &\n &\n &\n &\n &\n &\n &\n &\n &\n &\n &\y &\y &\n \\
  \rowcolor{tablerowA}  FARM~\cite{helmut2025tactile}         &\y &\n &\n &\n &\n &\n &\n &\n &\n &\y &\n &\n &\n &\n &\n &\n &\n &\n &\n &\n &\n &\n &\n &\y \\
  \rowcolor{tablerowB}    ViTacFormer~\cite{heng2025vitacformer}  &\y &\y &\n &\n &\n &\n &\y &\n &\n &\n &\n &\n &\n &\n &\n &\n &\n &\n &\n &\n &\n &\n &\n &\y \\
  \rowcolor{tablerowA}  3D-ViTac~\cite{huang20243d}     &\y &\n &\n &\n &\n &\n &\n &\n &\n &\y &\n &\n &\n &\n &\n &\n &\n &\n &\n &\n &\n &\n &\n &\y \\
  \rowcolor{tablerowB}    UniT~\cite{xu2025unit}         &\y &\n &\n &\n &\n &\n &\n &\n &\n &\y &\n &\n &\n &\n &\n &\n &\n &\n &\n &\n &\n &\n &\n &\y \\
  \rowcolor{tablerowA}  FuSe~\cite{jones2025beyond}         &\n &\n &\n &\n &\n &\n &\n &\n &\n &\y &\n &\n &\n &\n &\n &\n &\n &\n &\n &\n &\n &\n &\n &\y \\
  \rowcolor{tablerowB}    ForceVLA~\cite{yu2025forcevla}     &\y &\n &\n &\n &\n &\y &\y &\n &\n &\n &\n &\n &\n &\n &\n &\n &\n &\n &\n &\n &\n &\y &\y &\n \\
  \rowcolor{tablerowA}  Tactile-VLA~\cite{huang2025tactile}  &\y &\n &\n &\n &\n &\n &\y &\n &\n &\y &\n &\n &\n &\n &\n &\n &\n &\n &\n &\n &\n &\n &\n &\y \\
  \rowcolor{tablerowB}    TLA~\cite{hao2025tla}          &\y &\n &\n &\n &\n &\n &\n &\n &\n &\n &\n &\n &\n &\n &\n &\n &\n &\n &\n &\n &\n &\n &\n &\y \\
  \rowcolor{tablerowA}  FILIC~\cite{ge2025filic}        &\y &\y &\n &\n &\n &\n &\n &\n &\n &\n &\n &\n &\n &\n &\n &\n &\n &\n &\n &\n &\n &\n &\n &\n \\
  \rowcolor{tablerowB}    ManipForce~\cite{lee2025manipforce}   &\y &\n &\y &\n &\n &\n &\n &\n &\n &\n &\n &\n &\n &\n &\n &\n &\y &\n &\n &\n &\n &\y &\n &\n \\
  \rowcolor{tablerowA}  OGRL-VT~\cite{shirai2025sim}      &\n &\n &\n &\n &\n &\n &\n &\n &\n &\n &\n &\n &\y &\n &\n &\n &\n &\n &\n &\n &\n &\n &\n &\n \\
  \rowcolor{tablerowB}    ViTaL~\cite{zhao2025touch}        &\y &\n &\n &\n &\n &\n &\n &\n &\n &\n &\n &\n &\n &\n &\n &\n &\n &\n &\n &\n &\n &\n &\n &\y \\
  \rowcolor{tablerowA}  TACT~\cite{murooka2025tact}         &\n &\n &\n &\n &\n &\n &\n &\n &\n &\n &\n &\n &\y &\n &\n &\n &\n &\n &\n &\n &\n &\y &\n &\y \\
  \rowcolor{tablerowB}    VT-HMD~\cite{ye2026visual}       &\n &\y &\n &\n &\n &\n &\n &\n &\n &\n &\n &\y &\y &\n &\n &\n &\n &\n &\n &\n &\n &\n &\n &\y \\
  \rowcolor{tablerowA}  TaF-VLA~\cite{huang2026tactile}      &\y &\n &\n &\n &\n &\n &\y &\n &\n &\y &\n &\n &\n &\n &\n &\n &\n &\y &\n &\n &\n &\y &\n &\y \\
  \rowcolor{tablerowB}    Force-Policy~\cite{fang2026force} &\y &\n &\n &\n &\n &\n &\y &\n &\n &\n &\n &\n &\n &\n &\n &\n &\n &\n &\n &\n &\n &\y &\n &\n \\
  \rowcolor{tablerowA}  FACTR~\cite{liu2025factr}        &\n &\n &\n &\n &\n &\n &\n &\n &\n &\n &\n &\n &\n &\n &\y &\n &\n &\y &\n &\n &\n &\n &\y &\n \\
  \rowcolor{tablerowB}    MFMIL~\cite{ablett2024multimodal}        &\n &\n &\n &\n &\n &\n &\n &\n &\n &\n &\n &\n &\n &\n &\n &\n &\n &\n &\n &\n &\n &\n &\n &\y \\
  \rowcolor{tablerowA}  ForceSight~\cite{collins2024forcesight}   &\n &\n &\n &\n &\n &\n &\n &\n &\n &\n &\n &\y &\n &\n &\n &\n &\n &\y &\n &\n &\n &\y &\n &\n \\
  \rowcolor{tablerowB}    CRAFT~\cite{zhang2026craft}        &\y &\n &\y &\n &\n &\n &\y &\n &\n &\n &\n &\n &\n &\n &\n &\n &\n &\n &\n &\n &\n &\n &\y &\n \\
  \rowcolor{tablerowA}  EquiContact~\cite{seo2025equicontact}  &\y &\n &\n &\n &\n &\n &\y &\n &\n &\n &\n &\n &\n &\n &\n &\n &\n &\y &\n &\n &\n &\y &\n &\n \\
  \rowcolor{tablerowB}    MoDE-VLA~\cite{tang2026towards}     &\n &\n &\y &\n &\n &\y &\n &\n &\n &\y &\n &\n &\n &\y &\n &\n &\n &\n &\n &\n &\n &\n &\y &\y \\
  \rowcolor{tablerowA}  ForceVLA2~\cite{li2026forcevla2}    &\n &\n &\y &\n &\n &\n &\y &\n &\n &\n &\n &\n &\n &\n &\n &\n &\n &\n &\n &\n &\n &\y &\n &\n \\
  \rowcolor{tablerowB}    ReTac-ACT~\cite{ruan2026retac}    &\y &\n &\n &\n &\n &\n &\n &\n &\n &\n &\n &\n &\n &\n &\n &\n &\n &\n &\n &\n &\n &\n &\n &\y \\
  \rowcolor{tablerowA}  TacVLA~\cite{zhang2026tacvla}       &\y &\y &\y &\n &\n &\n &\n &\n &\n &\n &\n &\n &\n &\n &\n &\n &\n &\y &\n &\n &\n &\n &\n &\y \\
  \rowcolor{tablerowB}    KineDex~\cite{zhang2025kinedex}      &\y &\y &\y &\n &\n &\n &\n &\n &\n &\n &\n &\n &\n &\n &\n &\n &\n &\y &\n &\n &\n &\n &\n &\y \\
  \rowcolor{tablerowA}  OmniVTA~\cite{zheng2026omnivta}      &\n &\n &\y &\n &\n &\y &\y &\y &\n &\y &\n &\n &\y &\n &\n &\n &\n &\n &\n &\n &\n &\n &\n &\y \\
  \rowcolor{tablerowB}    OmniVTLA~\cite{cheng2025omnivtla}     &\n &\n &\n &\n &\n &\n &\n &\n &\n &\n &\n &\n &\n &\n &\n &\n &\n &\y &\n &\n &\n &\n &\n &\y \\
  \rowcolor{tablerowA}  ViTaS~\cite{tian2026vitas}        &\y &\n &\y &\n &\n &\n &\y &\n &\n &\n &\y &\n &\n &\y &\y &\n &\n &\y &\n &\n &\n &\n &\n &\y \\
  \rowcolor{tablerowB}    VLA-Touch~\cite{bi2025vla}    &\n &\n &\n &\n &\n &\y &\y &\n &\n &\n &\n &\n &\n &\n &\n &\n &\n &\y &\n &\n &\n &\n &\n &\y \\
  \rowcolor{tablerowA}  3DTacDex~\cite{wu2025canonical}     &\n &\n &\y &\n &\n &\n &\n &\n &\n &\n &\n &\n &\y &\n &\n &\n &\y &\n &\n &\n &\n &\n &\n &\y \\
  \rowcolor{tablerowB}    VTAO-BiManip~\cite{sun2025vtao} &\n &\n &\n &\n &\n &\n &\n &\n &\n &\n &\n &\n &\n &\n &\n &\n &\n &\n &\n &\n &\n &\n &\n &\y \\
  \rowcolor{tablerowA}  VTLA~\cite{zhang2025vtla}         &\y &\n &\n &\n &\n &\n &\n &\n &\n &\n &\n &\n &\n &\n &\n &\n &\n &\n &\n &\n &\n &\n &\n &\y \\
  \rowcolor{tablerowB}    ARCH~\cite{sun2024arch}         &\y &\n &\n &\n &\n &\n &\n &\n &\n &\n &\n &\n &\n &\n &\n &\n &\n &\n &\n &\n &\n &\y &\n &\n \\
  \rowcolor{tablerowA}  TouchGuide~\cite{zhang2026touchguide}   &\y &\n &\n &\n &\n &\y &\y &\n &\n &\y &\n &\n &\n &\n &\n &\n &\n &\n &\n &\n &\n &\n &\n &\y \\
  
  \bottomrule
  \end{tabular}%
  }

\end{table}

\vspace{-4pt}
\subsubsection{Deformable and Surface-contact Manipulation}
 
This class involves deformable objects or irregular surfaces, such as soft or fragile grasping, peeling, and wiping. For soft objects, geometry and dynamics can evolve continuously under contact; thus, the robot must adjust the applied force in real time, balancing the risk of damaging the object against the risk of task failure. Wiping and polishing additionally require stable contact forces over extended trajectories on potentially varying surfaces, which is qualitatively different from the localized contact events in insertion. These tasks share a common demand for compliance- and material-aware manipulation under adaptive force control over extended trajectories.

\begin{figure}[!t]
    \centering
    \IfFileExists{figs/task_histogram.pdf}{%
        \includegraphics[width=0.90\linewidth]{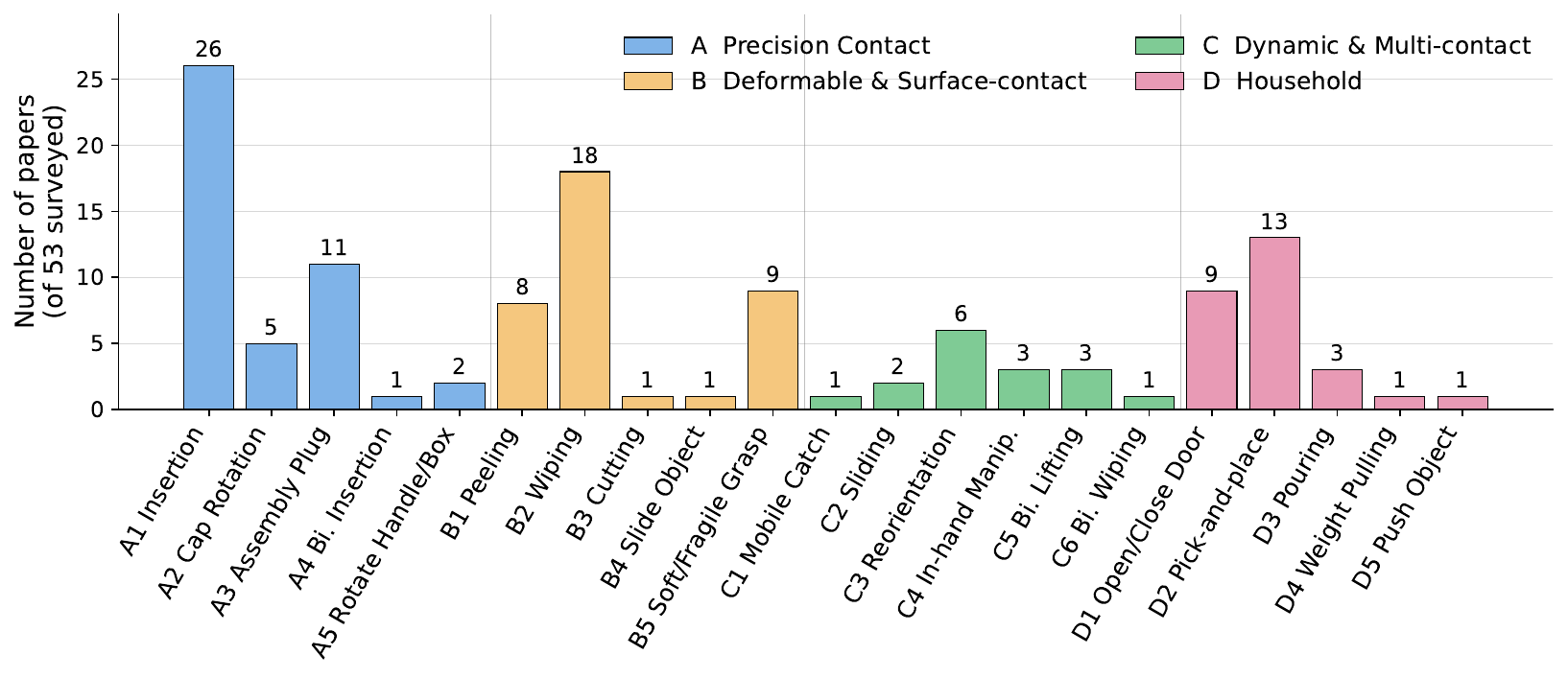}%
    }{%
        \fbox{\parbox[c][0.28\textheight][c]{0.92\linewidth}{\centering Missing figure: \texttt{figs/task\_histogram.pdf}}}%
    }
    \vspace{-15pt}
    \caption{Number of papers (out of 53) evaluating each of the 21 benchmark tasks, grouped by category. A~Precision Contact (blue), B~Deformable \& Surface-contact (orange), C~Dynamic \& Multi-contact (green), D~Household and Long-Horizon Manipulation (pink). The numeric label above each bar is the paper count.}
    \vspace{-5pt}
    \Description{}
    \label{fig:task-histogram}
\end{figure}

\vspace{-4pt}
\subsubsection{Dynamic and Multi-contact Manipulation}
 
This class includes time-critical or coordinated tasks, such as reorientation, in-hand manipulation, and bimanual manipulation, which require rapid responses to transient contact. Tasks such as dynamic catching require trajectory estimation and contact response within a narrow time window, placing stringent demands on real-time feedback integration. In-hand manipulation requires simultaneous estimation of multiple contact states and fine-grained force control as the object shifts within the gripper. Bimanual tasks require balanced force regulation and coordinated trajectory generation across two manipulators. These tasks share a common demand for rapid, multi-point coordination of force and tactile feedback under partially observable contact.

\vspace{-4pt}
\subsubsection{Household and Long-horizon Manipulation}
 
This class involves long-horizon, multi-stage tasks, such as pick-and-place and door opening, in unstructured but common daily environments. While pick-and-place is traditionally treated as vision-centric, the tactile- and force-aware variants surveyed here involve fragile objects, uncertain grasp stability, deformable materials, or cluttered contact conditions that require contact feedback during execution. Opening a door or drawer requires constrained contact, rotational motion, and friction-aware control; cooking-related tasks typically integrate grasping, cutting, and container manipulation within a single scene. In this context, material texture and friction can vary widely across daily objects, and visual occlusion is more common in complex household environments, making contact perception a primary requirement. These tasks share a common demand for long-horizon, generalizable force-aware manipulation across diverse and unstructured settings.

\vspace{-4pt}
\subsubsection{Summary of Task Distribution}
 
These four categories capture distinct perspectives on physical interaction intelligence: precision contact emphasizes force-guided interaction under tight geometric constraints; deformable and surface-contact manipulation emphasizes compliance and material awareness over extended trajectories; dynamic and multi-contact manipulation emphasizes rapid feedback and multi-point coordination; and household and long-horizon manipulation emphasizes long-horizon generalization in unstructured environments. Together, they trace a progression from localized force-guided contact alignment to long-horizon, multi-stage interaction reasoning.
The actual task distribution, however, is strongly skewed, as illustrated in Figure~\ref{fig:task-histogram}. Peg-in-hole insertion (A1, 26 papers), surface wiping (B2, 18), and pick-and-place (D2, 13) together account for nearly half of all task-method pairings. A second tier, including A3, D1, B5, B1, and C3, covers 6--11 papers each and covers most of the remaining studies. This distribution indicates that the field is concentrated on narrowly scoped precision-contact tasks, whereas dynamic interaction, long-horizon manipulation, and multi-contact coordination remain comparatively under-explored. 
Overall, the task distribution shows that current studies mainly evaluate contact-refinement problems, while long-horizon reasoning and highly dynamic interaction remain less explored.

\subsection{Force and Tactile Sensing Devices}

Contact-rich interaction does not always require dedicated force or tactile hardware. However, when contact conditions are uncertain, partially observable, or dynamically changing, vision alone cannot reliably infer the relevant interaction states or support timely adaptation, making tactile or force feedback essential. The experimental setups of the 53 surveyed methods therefore place strong emphasis on contact sensing, as shown in Table~\ref{tab:tasks}: 25 employ wrist-mounted force/torque (F/T) or joint torque sensing (18 use wrist F/T sensing and 9 use joint torque sensing, with 2 using both), 27 use tactile sensors, 4 combine a force-domain sensor with a tactile sensor, and 5 either estimate force from proprioceptive signals or rely on vision alone.

Each sensing modality provides distinct information about physical interaction and differs in its mounting location, output signals, and ability to capture different contact properties. Although their sensing mechanisms have been introduced in Sections~\ref{subsection: Tactile} and~\ref{subsection: Force}, here we examine the practical functionality, advantages, and limitations of each sensor type, grounded in its use across the surveyed methods. Table~\ref{tab:sensor-compare} summarizes these characteristics, while Table~\ref{tab:tasks} records the sensing modalities adopted by each paper.

\begin{table}[!t]
\centering
\caption{Sensing modalities used by the surveyed 53 methods.}
\vspace{-6pt}
\label{tab:sensor-compare}
\small
\setlength{\tabcolsep}{4pt}
\renewcommand{\arraystretch}{1.25}
\hspace*{-0.45cm}\begin{tabular}{>{\raggedright\arraybackslash}m{2.00cm}>{\raggedright\arraybackslash}m{2.65cm}>{\raggedright\arraybackslash}m{3.15cm}>{\raggedright\arraybackslash}m{3.75cm}}
\toprule
\rowcolor{cS}
 & \multicolumn{1}{c}{\textbf{Wrist F/T}} & \multicolumn{1}{c}{\textbf{Joint Torque}} & \multicolumn{1}{c}{\textbf{Static Tactile}} \\
\midrule
\textbf{Mounting}     & wrist \emph{(extrinsic/intrinsic sensing)} & each joint \emph{(intrinsic sensing)} & gripper, fingertip, palm, or skin \emph{(extrinsic/cutaneous)} \\
\rowcolor{tablerowA}
\textbf{Raw signal}   & 6-axis F/T vector & joint-torque vector & image, taxel array, or contact points \\
\textbf{Derived info} & end-effector wrench & Jacobian-mapped wrench; external joint torque & contact patch, pressure, shear \\
\rowcolor{tablerowA}
\textbf{Limitation}   & no localization on the tool & needs an accurate dynamic model & limited to the sensing surface \\
\textbf{Typical use}  & insertion, assembly, compliance & joint compliance & in-hand manipulation, grasp and texture sensing \\
\rowcolor{cS!60}
\textbf{Adoption}     & 18/53 (34\%) & 9/53 (17\%) & 27/53 (51\%) \\
\bottomrule
\end{tabular}
\vspace{-8pt}
\end{table}

\subsubsection{End-Effector Wrench Sensors}
 
End-effector wrench sensors measure the resultant wrench acting at the end-effector. Typically mounted between the robot flange and the end-effector, the sensor provides a single six-axis wrench measurement. This sensing modality offers two main advantages. First, because the sensor is located near the end of the arm, it directly measures external interaction wrenches without relying on robot-dynamics models or joint-level torque estimation. Second, it can be retrofitted to a wide range of standard robotic arms, enabling wrench sensing even when the robot is not originally equipped with force/torque sensing capability. The limitations are also clear. Because the sensor provides only a single wrench measurement at the wrist, the precise contact location on the tool generally cannot be determined from the measured wrench alone. Moreover, contacts occurring elsewhere along the robot arm may remain undetected because only loads transmitted through the wrist interface are measured. 

\vspace{-4pt}
\subsubsection{Joint Torque Sensors}
 
Joint torque sensing captures interaction effects across the entire robot arm, rather than only at the end-effector. It measures the torque exerted at each actuated joint, from which two types of information can be inferred. First, the joint-torque vector can be mapped through the manipulator Jacobian to estimate the same six-axis end-effector wrench that a wrist F/T sensor provides, making end-effector contact observable even without a wrist-mounted sensor. Second, contact applied anywhere along the robot body appears in the joint-torque signal, supporting whole-arm compliance control and safe behavior under accidental link contact. The limitations are twofold. Isolating interaction torque from gravity and inertial effects requires an accurate dynamic model of the arm. In addition, joint torque sensors must be integrated into the actuators rather than added afterward, making this modality unavailable on standard position-controlled manipulators without additional hardware cost. 

\vspace{-4pt}

\subsubsection{Tactile Sensors}

Tactile sensors provide information about contact between the robot and its environment. Section~\ref{sec:tactile_format} introduced the data formats used to represent tactile signals. Here, we discuss the practical specifications and selection considerations for tactile sensors. Depending on their design, these sensors can capture contact location and area, local surface geometry, pressure or force distribution, deformation, slip, and vibration. In the surveyed learning systems, the two dominant types are vision-based tactile sensors and taxel arrays. They differ in spatial resolution, sensing area, sampling rate, sensitivity, and integration requirements.

\vspace{-4pt}
\paragraph{Vision-Based Tactile Sensors.}
A vision-based tactile sensor places one or more cameras behind a deformable contact surface. When contact occurs, the camera records changes in the shape or appearance of the surface or the movement of visual markers. The resulting tactile images can be used to estimate contact geometry, surface texture, slip, and force distribution. The main advantage of these sensors is their dense contact information and high spatial resolution, often at the submillimeter scale. However, their sensing area is generally limited to the deformable surface observed by the camera. Their sampling rate and latency also depend on the camera, image transmission, and subsequent image processing. Among the 15 surveyed methods using vision-based tactile sensing, 7 employ sensors from the GelSight family, including GelSight, GelSight Mini, and GelStereo, with one setup additionally using MC-Tac. The remaining 8 use DIGIT, STS, vision-based tactile sensors integrated into the SharpaWave hand, Xense optical tactile sensors, configurations combining multiple optical tactile sensors, or custom-built sensors.

\vspace{-4pt}
\paragraph{Taxel-Based Tactile Sensors.}
Taxel-based sensors consist of individual sensing units arranged over a contact surface. Depending on the sensor design, each taxel records local pressure, normal or shear force, or deformation. Together, these measurements form a spatial contact map whose resolution depends on the size, spacing, and arrangement of the taxels. Compared with vision-based tactile sensors, taxel arrays can provide higher sampling rates and may be easier to distribute over curved or larger surfaces, although these properties depend on the specific sensor design.

Taxel arrays can use resistive or piezoresistive, capacitive, barometric, or magnetic sensing. In a magnetic tactile sensor, contact deforms a soft material containing a magnet or magnetic particles, which changes the magnetic field measured by nearby sensors. These changes can be used to estimate local deformation or contact forces. Magnetic tactile sensors can be compact and durable. Resistive and capacitive arrays are also widely used because they can be relatively inexpensive to fabricate and extended to larger sensing areas. Such arrays may be deployed as localized patches on gripper pads or robot links~\cite{cheng2019comprehensive}, or integrated into the fingertips and phalanges of dexterous robotic hands.

\paragraph{Vibration- and Acoustic-Based Sensing.} Although not yet common among the 53 methods included in the TF-ART corpus, dynamic vibrotactile and acoustic sensing provides an additional ways of capturing contact information. Rather than measuring quasi-static deformation or pressure distributions, these sensors use accelerometers, piezoelectric elements, or high-bandwidth pressure transducers to capture short-lived vibrations and acoustic signals generated by sliding, impact, or slip. For example, BioTac encloses a conductive fluid beneath an elastomeric skin, allowing contact-induced microvibrations to propagate through the fluid as pressure waves and be measured by an internal hydro-acoustic pressure sensor~\cite{fishel2012bayesian}. The resulting temporal and spectral patterns can support slip, texture, and impact detection. While direct tactile transduction remains confined to the instrumented local surface, vibrations generated by remote contacts can propagate through the hand and a grasped object or tool, enabling learned inference of contact locations beyond that surface~\cite{taunyazov2021extended}.

\subsection{Challenges in Contact Sensing}
Wrist-mounted wrench, joint torque, and tactile sensing capture complementary aspects of physical interaction: wrench sensing provides a direct global wrench at the end-effector, joint torque sensing extends contact visibility along the robot body, and tactile sensing focuses on the spatial location and distribution of contact at the end-effector. Together, these modalities support different levels of force, compliance, and contact-geometry awareness.

However, these modalities are rarely combined in practice. Only 4 of the 53 methods (RDP~\cite{xue2025reactive}, TACT~\cite{murooka2025tact}, TaF-VLA~\cite{huang2026tactile}, MoDE-VLA~\cite{tang2026towards}) pair a force-domain sensor with a tactile sensor. Most evaluations are confined to single-stage tasks on fixed robot platforms in controlled environments, while dynamic multi-contact interaction, long-horizon manipulation, and cross-platform generalization remain under-explored. Standardized evaluation protocols for tactile- and force-aware foundation models are also largely absent, although taxonomy-grounded dexterity benchmarks for anthropomorphic hands such as POMDAR~\cite{liconti2026pomdar}, which score contact-rich manipulation as task throughput in both real-world and simulated settings, offer a reproducible, performance-based template that such protocols could build on. Closing these gaps is a prerequisite for developing foundation models that acquire transferable physical interaction intelligence rather than task-specific motion patterns.

\section{Concluding Discussion and Future Directions} \label{S7}

Although recent tactile- and force-aware robot learning has achieved substantial progress, the field remains far from developing broadly transferable physical interaction intelligence. Existing studies have shown that force and tactile feedback can improve contact-rich manipulation, stabilize task execution, and complement vision-centric policies. However, the reviewed literature also exposes unresolved challenges in sensing, representation learning, policy architecture, control integration, and evaluation. We therefore divide the discussion into two parts. First, we organize the discussion around the individual components of the TF-ART pipeline and identify future research directions for each component. We then discuss broader research directions that extend beyond individual pipeline components. 

\subsection{Concluding Discussion on TF-ART-Centered Challenges}

\subsubsection{Force and Tactile Modalities}

Force and tactile sensing provide complementary information about physical interaction. Force/torque measurements capture the overall interaction loads between the robot and its environment, whereas tactile sensing, as currently used in robot-learning systems, primarily provides local information about contact geometry, pressure distribution, and deformation at the contact surface. Together with vision, language, and proprioception, these signals constitute key components of the TF-ART observation space.

Despite their complementary roles, systematically integrating force and tactile sensing across different robot-learning frameworks remains challenging because of substantial cross-sensor and cross-embodiment variation. Wrist wrench and joint torque measurements are embodiment-dependent: the same physical contact can produce different readings across robot platforms because of sensor bias, sensor mounting frames, tool loads, and differences in robot dynamics and geometry. Tactile signals are similarly heterogeneous across hardware. Tactile sensors differ in sensing mechanism, spatial resolution, morphology, contact response, and mounting location, while their outputs take different forms, including tactile images, taxel maps, and contact-point measurements. Consequently, policies and encoders are often tied to particular sensors and robot embodiments, making cross-sensor and cross-embodiment generalization difficult. The field therefore lacks shared pretrained backbones comparable to those available for vision and language, leaving the complementary information provided by force and tactile sensing underutilized. The surveyed methods~\cite{xue2025reactive, murooka2025tact, huang2026tactile, tang2026towards} demonstrate the potential of combining these two modalities to obtain a more complete representation of physical interaction, but their systematic integration remains underexplored.

Future work should develop systematic methods for jointly encoding and fusing force and tactile observations across sensors and embodiments. It should also verify whether combining their complementary load-level and contact-level information consistently improves performance across diverse contact-rich tasks. Generalist tactile policies pretrained across different tactile sensors~\cite{yuan2026ftp1, zhangcross} indicate that a shared foundation is attainable, while human-to-robot tactile alignment further demonstrates the potential for cross-embodiment policy transfer~\cite{wi2025tactalign}. Extending such a foundation to incorporate force-domain sensing remains an open problem.

\subsubsection{Representation Learning and Fusion}

Representation learning and fusion together determine how multimodal observations become usable for action generation. The former aligns semantically related observations in a shared latent space while preserving fine-grained, modality-specific information, whereas the latter regulates how and when the encoded features contribute to the policy. Across the surveyed methods, these techniques help bridge cross-modal gaps and enable policies to incorporate contact information alongside vision and language~\cite{jones2025beyond,ruan2026retac,sferrazza2024power,tian2026vitas}.

Representation learning and fusion nevertheless remain challenging in practice because different modalities vary substantially in dimensionality, temporal frequency, noise characteristics, and task-dependent relevance. Vision and language provide effective semantic context in contact-free settings, whereas tactile and force signals become most informative during physical contact. Several surveyed methods address this phase-dependent variation in modality relevance by regulating modality contributions through expert routing~\cite{yu2025forcevla,tang2026towards,li2026forcevla2} or gated filtering~\cite{he2025foar,fang2026force,ruan2026retac,zhang2026tacvla,zheng2026omnivta}. However, the signals that implicitly encode task phases, such as contact probabilities and learned routing weights, remain system-specific and are typically learned from task-specific demonstrations.

A promising direction is therefore to develop a generalizable phase-encoding framework. Such a framework could provide auxiliary signals throughout task execution to indicate interaction phases, such as non-contact, pre-contact, and in-contact, as well as broader task stages in general long-horizon manipulation. The resulting phase representation could guide the policy in determining which modalities are most informative, which motion primitives are most suitable, and which expert skill should be activated from an available skill library.

\subsubsection{Hierarchical Policy Design}

There is a fundamental tension between large-scale semantic reasoning and low-latency contact response. Robot foundation models show strong cross-task generalization, but contact-rich manipulation often requires millisecond-level feedback, millimeter-level precision, and accurate force regulation. Large policies alone may be too slow to meet the latency requirement and too coarse to achieve the required motion and force-control precision. Some surveyed systems therefore employ multi-phase architectures to address these competing requirements.

Multi-phase architectures provide a natural way to reconcile these requirements. A high-level model generates semantic or coarse motion plans, while a lower-level refinement policy or controller uses force and tactile feedback to make rapid adjustments~\cite{fang2026force,he2025foar, murooka2025tact, xue2025reactive, zhang2025kinedex, zheng2026omnivta, zhi2025learningunifiedpolicyposition}. Future systems should further formalize this division of labor, with slow reasoning modules handling task planning and fast reactive modules handling contact adaptation. Another promising direction is to develop dedicated low-level skill experts, such as modular in-hand manipulation skills or elementary manipulation primitives, and integrate them with high-level reasoning through skill routing. These skill-level experts should not be tied to specific object-task pairs. Instead, they should generalize across objects while specializing in particular motion patterns, such as approaching, pushing, or avoiding obstacles.

\subsubsection{Integrating Learned Policies with Robot Control}

Learning-based methods can model complex action distributions and learn from demonstrations, but contact-rich execution also depends on model-based controllers that provide stability and safety. Current systems often connect learned policies to low-level controllers mainly through action commands, leaving other control-relevant parameters, such as reference force and stiffness, to be tuned manually through experiments. Several works have explored more flexible integration by allowing learning models to predict these control parameters or controller-specific targets, which are referred to as intermediate modalities in~\cite{collins2024forcesight, fang2026force, portela2024learning, wu2025tacdiffusion, zhang2025kinedex, zhi2025learningunifiedpolicyposition, hou2025adaptive, aburub2026learning, xu2026contact}. Since predicted parameters are meaningful only when the robot exposes a matching control mode, the trained models are inherently tied to the corresponding platforms.

To reduce this platform dependence, a promising direction is to develop action-generation policies that mimic compliant-control behavior for low-cost robots without robust force-sensing or force-control capabilities. Such policies could leverage robot-dynamics-based force prediction, operate under position-control interfaces, and generalize across a wider range of platforms and contact-rich tasks.

\subsubsection{Tasks, Benchmarks, and Evaluation}

Evaluation protocols are still insufficient for assessing generalizable physical interaction capability. Many existing studies focus on short-horizon, single-stage tasks such as insertion, wiping, and pick-and-place, while dynamic multi-contact manipulation, deformable-object handling, and long-horizon household tasks are underexplored. Moreover, experiments are often conducted on fixed platforms with specific sensor configurations, making it difficult to compare methods systematically beyond task success rate or to evaluate cross-embodiment transfer.

The field therefore needs standardized tactile- and force-aware benchmarks that cover diverse objects, robot platforms, and sensor configurations. Recent taxonomy-grounded dexterity benchmarks discussed in Section~\ref{S6} illustrate one route toward such reproducible, performance-based evaluation. Future benchmarks should measure not only task success, but also force safety, contact stability, recovery from perturbations, generalization to unseen materials, and robustness under sensor failure. Nevertheless, assembling data collected from multiple platforms and sensors into a unified benchmark remains highly challenging, requiring substantial effort from both methodological and engineering perspectives.

\subsection{Broader Future Directions}

Beyond the component-wise directions summarized under the proposed TF-ART, we further discuss broader directions that are not yet fully developed in the tactile- and force-aware manipulation context, but are necessary and promising for future research.

\textbf{Simulation and Data Scaling.} Simulation and data scaling remain major bottlenecks. Large-scale robot foundation models benefit from broad and diverse datasets, but tactile and force data are expensive to collect and difficult to simulate accurately. Current simulators still struggle to reproduce high-resolution tactile deformation, frictional contact, material compliance, and multi-contact dynamics. Future research should combine real-world data collection, physics-based simulation, and tactile and force prediction to reduce data-collection costs and enable data scaling in simulation environments~\cite{hou2026world, wu2025canonical, zheng2026omnivta}. In particular, real-world data can be used to calibrate and guide more accurate simulation of visual-tactile images, taxel-array deformation, external wrench measurements, and internal force or torque perception. This would support the generation of large-scale synthetic data that better approximates real-world contact phenomena, thereby reducing the sim-to-real gap and enabling more direct real-world deployment of tactile- and force-aware models.

\textbf{Contact-informed World Action Models.} Another promising direction is to connect tactile- and force-aware robot learning with recent advances in World Action Models (WAMs)~\cite{hou2026world, tian2026vtwamvisualtactileworldaction, yuan_vtam_2026, lou2026dreamtacunifiedtactileworld, he2026fawamforceawareworldaction, zang2026tacforesightforceguidedtactileworld, zhang2026contactworldmattersvisiontactileworld, wu2026tactilewamtouchawareworldaction}. Most existing WAMs use video-based latent prediction as their main objective, but they rarely formulate future-state prediction from an interaction-centered perspective. In contact-rich manipulation, a world model should predict not only the visual evolution of the environment, but also object deformation, dynamic behavior, and the spatial relationship between the robot and the environment under contact. In this way, WAMs could move toward more general-purpose world simulators that equip robots with physically grounded reasoning capabilities. Such models may allow robots to anticipate interaction outcomes and respond to predicted, or “imagined,” forces even when direct force sensing is unavailable. A more practical near-term direction is to jointly predict visual, tactile, and force evolution instead of modeling the full environment dynamics. This could help robots anticipate contact events, avoid force overshoot, and establish contact more reliably.

\textbf{Touch and Force as Intrinsic Critics.} Building on these predictive capabilities, a further promising direction is to use tactile and force sensing as a primary substrate for robot self-supervision and self-improvement, rather than merely as auxiliary feedback. This direction reflects a broader shift from merely generating robot behaviors to continuously evaluating, correcting, and improving them, as exemplified by recent work on test-time policy steering~\cite{wu2026inference}, verifier-guided action selection~\cite{kwok2026scalingverification, kwok2026llmverifier}, self-correcting and contact-aware world models~\cite{liu2026worldactionverifier, higuera_visuo-tactile_2026, he2026fawamforceawareworldaction}, force-based anomaly detection~\cite{lin2025anofdiff}, and robotic reward modeling~\cite{liang2026robometer,lee2026roboreward}. In contact-rich manipulation, tactile and force signals are particularly well suited to this role because they reveal whether an interaction is physically correct, rather than merely visually plausible. Future systems could verify candidate actions against predicted contact trajectories, use discrepancies between expected and observed signals to detect failures and refine policies or world models online, and train contact-aware reward models that assess execution quality, stability, and recovery from both successful and failed experiences. In this view, touch and force are not only perception channels or control inputs, but intrinsic critics by which robots can assess and improve the physical quality of their own execution.

\textbf{Mechanical Compliance as a Design Variable.} A further underexplored direction is to treat mechanical compliance as a design variable rather than only a control objective. The control methods surveyed in Section~\ref{S5} regulate compliance through impedance, admittance, and force control on largely rigid manipulators. Given the tight coupling between learned policies and platform control interfaces discussed above, an orthogonal route is to embed compliance directly in the hardware, through soft or tendon-driven actuation, antagonistic variable-stiffness joints, and inherently compliant anthropomorphic hands~\cite{sonoda2026sensorless, neumann2026palm, kazemipour2026decoupling}. Such embodiments absorb impact, distribute contact, and stabilize interaction passively, reducing the bandwidth and accuracy demanded of force sensing and feedback control, and in some cases enabling safe contact-rich behavior with little or no explicit force sensing. This points toward joint co-design of body and policy, in which morphology, actuation compliance, and learned control are optimized together rather than layering a policy onto a fixed platform. For tactile- and force-aware learning specifically, mechanically compliant embodiments also reshape the sensing problem, since intrinsic compliance changes how contact forces are transmitted to and observed by the available sensors.

In the long term, tactile- and force-grounded robot intelligence should move beyond learning task-specific motion patterns and toward acquiring transferable physical concepts, such as contact, pressure, friction, and stability. Achieving this goal will require unified multimodal representations, phase-aware policy architectures, systematic policy-control integration, compliant embodiment and body-policy co-design, stronger awareness of world evolution, and standardized benchmarks for real-world physical interaction.

\bibliographystyle{unsrtnat}
\bibliography{reference}

\end{document}